\documentclass[10pt,twocolumn,letterpaper]{article}

\usepackage{cvpr}              %
\graphicspath{ {./images/} }
\usepackage{multirow}
\usepackage{pifont}
\usepackage{algorithm}
\usepackage{algpseudocode}

\definecolor{cvprblue}{rgb}{0.21,0.49,0.74}
\usepackage[pagebackref,breaklinks,colorlinks,allcolors=cvprblue]{hyperref}
\usepackage{tikz}
\usepackage[accsupp]{axessibility}  %

\def\paperID{} %
\def\confName{CVPR}
\def\confYear{2026}

\title{RAD: Retrieval-Augmented Monocular Metric Depth Estimation \\ for Underrepresented Classes}

\author{
\vspace{-0.07cm}
Michael Baltaxe$^{1, 2}$ \quad Dan Levi$^{1}$ \quad Sagie Benaim$^{2}$\\
\\ $^{1}$General Motors, Herzliya, Israel \quad $^{2}$The Hebrew University of Jerusalem, Jerusalem, Israel
}

\begin{document}

\maketitle

\begin{abstract}

Monocular Metric Depth Estimation (MMDE) is essential for physically intelligent systems, yet accurate depth estimation for underrepresented classes in complex scenes remains a persistent challenge. To address this, we propose RAD, a retrieval-augmented framework 
that approximates the benefits of multi-view stereo by utilizing retrieved neighbors as structural geometric proxies. 
Our method first employs an uncertainty-aware retrieval mechanism to identify low-confidence regions in the input and retrieve RGB-D context samples containing semantically similar content. We then process both the input and retrieved context via a dual-stream network and fuse them using a matched cross-attention module, which transfers geometric information only at reliable point correspondences. Evaluations on NYU Depth v2, KITTI, and Cityscapes demonstrate that RAD significantly outperforms state-of-the-art baselines on underrepresented classes, reducing relative absolute error by 29.2\% on NYU Depth v2, 13.3\% on KITTI, and 7.2\% on Cityscapes, while maintaining competitive performance on standard in-domain benchmarks. Project page: \href{https://michaelbaltaxe.github.io/rad}{https://michaelbaltaxe.github.io/rad}

\end{abstract}

\begin{figure}
    \centering
    \includegraphics[width=\linewidth]{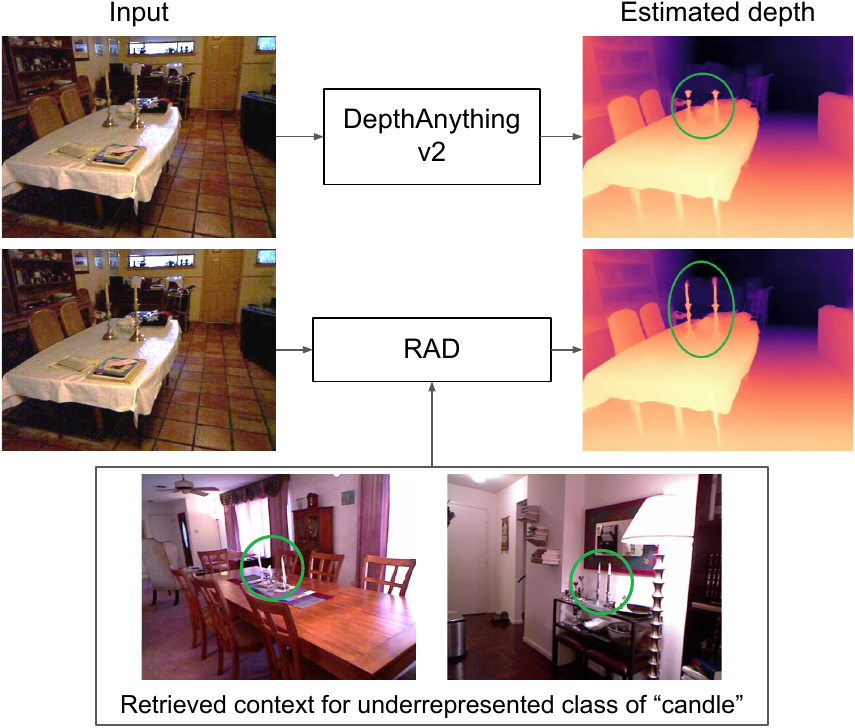}
    \caption{\textbf{Illustration.} Given an input image, \textit{RAD} (using DepthAnything v2 backbone~\cite{DepthAnything_V2}) retrieves context views for highly uncertain objects of underrepresented classes (e.g., candles) to serve as structural geometric proxies. 
    These are used as part of a dual-stream network to output an accurate monocular metric depth estimation, in comparison to the direct baseline of DepthAnything v2, fixing uncertain regions.}
    \label{fig:teaser}
    \vspace{-0.5cm}
\end{figure}

\section{Introduction}
\label{sec:introduction}
Monocular metric depth estimation (MMDE) is crucial to building intelligent systems and immersive digital experiences that interact with the physical world. Applications span autonomous driving \cite{pseudo-lidar}, human-computer interaction \cite{learning-to-be-a-depth-camera, du2020depthlab}, and novel view synthesis \cite{casual-3d-photography}. Given the long-tailed nature of real-world data, accurately estimating depth for underrepresented classes is essential for their robust deployment. In this work, we address MMDE in complex scenes that contain long-tailed, underrepresented classes.

Recent years have seen substantial progress in MMDE. Two prevailing directions have been investigated, depending on the type of data available for the task. One set of approaches relies on using domain-specific data to fine-tune depth estimation networks that incorporate strong priors \cite{dorn, newcrf, adabins, iebins}. These methods aim for specialization, resulting in networks that accurately estimate depth when trained with sufficient representative data. The second set of approaches learns from several large-scale datasets to perform zero-shot MMDE  \cite{UniDepth_V2, DepthPro, ZoeDepth, zerodepth}, which aims for strong generalization, enabling the model to perform well on unseen target domains without test-time adaptation. 
Although promising, key gaps persist. On the one hand, fine-tuning methods yield highly accurate estimates for in-distribution classes, but when presented with underrepresented classes, the priors are inaccurate, leading to large errors. On the other hand, zero-shot methods ignore domain-specific information, resulting in a performance degradation. 

It is well established that multi-view methods outperform monocular depth estimation due to their incorporation of geometric constraints from multiple viewpoints. Such additional information helps mitigate the limitations inherent to monocular setups. Our approach, called RAD (standing for Retrieval Augmented Depth), remains within the monocular estimation framework but seeks to approximate the benefits of multi-view systems by retrieving content-relevant images that serve as proxy auxiliary supporting views for the input (see Fig.~\ref{fig:teaser}). To this end, we draw inspiration from the retrieval-augmented generation (RAG) paradigm \cite{lewis-rag}, used in 
vision and language tasks.

Our training procedure for RAD consists of three main stages. First, we source a context sample, which consists of a context image (RGB) and its corresponding ground-truth depth map (D). This sample is generated in one of two ways: (a) via our uncertainty-aware retrieval pipeline, which uses the baseline's uncertainty to find a similar context image, or (b) via 3D augmentation, which generates a new image and depth map from a random viewpoint. Second, we establish spatial correspondences between the input image and the sourced context image (for retrieved images, this is done via point matching; for 3D augmentations, these correspondences are known via the geometry). Finally, the complete context sample (RGB+D) is processed by our dual-stream network, which uses a matched cross-attention mechanism to fuse the context information only at these corresponding locations. At inference, we follow the same procedure as in training, with the exception that context images are only obtained using our uncertainty-aware retrieval pipeline.

We demonstrate that RAD delivers substantial gains over state-of-the-art MMDE methods in regions containing underrepresented classes, while maintaining or even enhancing performance in in-domain regions. To validate this, we evaluate RAD across both underrepresented and in-domain categories in complex scenes, comparing against fine-tuning and zero-shot baselines on NYU Depth v2 \cite{dataset-nyud_v2}, KITTI \cite{dataset-kitti}, and Cityscapes \cite{dataset-cityscapes}. To enable a robust evaluation on underrepresented classes, we introduce a new benchmark depicting underreprested classes from each of these datasets. 
On underrepresented classes, our method achieves relative improvements in absolute relative error of 29.2\%, 13.3\%, and 7.2\% on NYU, KITTI, and Cityscapes, respectively. For in-domain classes, RAD matches or even improves upon existing methods without degradation.

\section{Related Work}
\label{sec:related work}

\noindent \textbf{Supervised Moncular Depth Estimation.} \quad
Most monocular depth estimation approaches rely on fully annotated datasets. Early work \cite{eigen2014depth, deeper-depth-prediction} showed that depth can be directly regressed from ground truth and introduced loss functions to address scale ambiguity. This paradigm has evolved through improved regression techniques \cite{dorn, dpt}, depth distribution modeling \cite{adabins, localbins, iebins}, and global scene consistency enforcement \cite{newcrf}, resulting in highly accurate methods. While supervised models learn strong priors from large datasets, they struggle to generalize to long-tail regions, precisely the focus of our work.

\noindent \textbf{Self-supervised Moncular Depth Estimation.} \quad
Due to the data demands of fully supervised methods, self-supervised approaches have emerged as an alternative. These typically use video sequences, multi-camera setups, or auxiliary signals, paired with loss functions enforcing color constancy to enable monocular depth estimation \cite{zhou2017unsupervised, monodepth_v2, packnet, self-supervision-planedepth, self-supervision-superdepth}. Like supervised methods, they rely on large datasets to learn strong priors and struggle with underrepresented data. Our retrieval mechanism addresses this by explicitly sourcing relevant context.

\noindent \textbf{Domain Adaptation for Monocular Depth Estimation.} \quad
Domain adaptation for monocular depth estimation addresses the shift between labeled source and unlabeled target domains \cite{domain-adaptation-real-time, domain-adaptation-geometry, domain-adaptation-desc}. While it bridges cross-domain gaps, our work focuses on intra-domain imbalance caused by long-tailed data distributions.

\noindent \textbf{Zero-shot and Few-shot Monocular Depth Estimation.} \quad
Recent zero-shot methods use massive datasets to enable generalization. These approaches have progressed from estimating relative depth \cite{midas, omnidata, leres, DepthAnything_V2, marigold}, to recovering metric scale via camera intrinsics \cite{zerodepth, metric3d, metric3d_v2, UniDepth, ZoeDepth}, and more recently, to jointly estimating depth and camera parameters \cite{DepthPro, UniDepth_V2}. \cite{clip-few-shot-mde} proposed a few-shot method to adapt to new distributions using minimal labeled data. Unlike our approach, both paradigms trade domain-specific precision for generality, often degrading performance on complex, underrepresented classes where local priors are crucial.

\noindent \textbf{Retrieval for Monocular Depth Estimation.} \quad
While \cite{non-parametric-depth} introduced non-parametric retrieval with costly test-time optimization, we present the first deep, parametric, feed-forward retrieval-augmented framework. Unlike multi-view methods \cite{fusing-flow-to-depth, Single-View-and-Multiview-Depth-Fusion, Learning-to-Fuse-Monocular-and-Multi-view, Stereo-and-Mono-Depth} that fuse views of the \textit{same} scene, our approach retrieves \textit{semantically similar} in-the-wild images to enhance single-image depth estimation.

\begin{figure}[tbh]
    \centering
    \includegraphics[width=\linewidth]{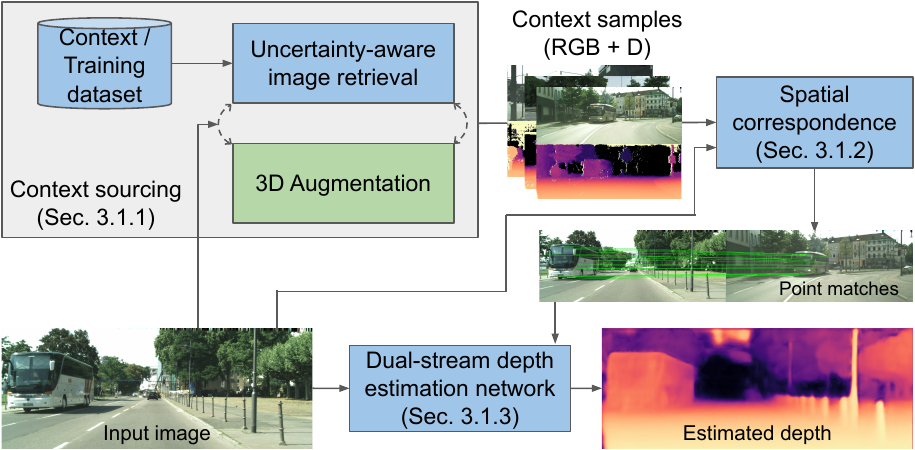}
    
    \caption{\textbf{RAD Pipeline.} Given an input image, a set of context samples is sourced (Sec.~\ref{sec:point matching}) using either \textit{uncertainty aware image retrieval} (both at training and inference)  or \textit{3D augmentation} (only during training). Subsequently, spatial correspondences are established (Sec.~\ref{sec:point matching}). These are used to infer depth via a dual-stream depth estimation network employing \textit{matched cross-attention}  (Sec.~\ref{sec:depth estimation network}). Blue blocks indicate components used for  training and inference, while the green block is only for training.
    }
    \label{fig:system architecture}
    \vspace{-0.4cm}
\end{figure}

\noindent \textbf{OOD and Open-Set 3D Vision.} \quad
Most out-of-distribution (OOD) approaches in 3D vision focus on \textit{anomaly detection} \cite{ood-detection-huang, ood-detection-veeramacheneni, ood-detection-hornauer, ood-detection-kosel, ood-detection-soum, ood-detection-zhang}. Recent open-set methods use deep visual \cite{open-set-detection-cen, open-set-detection-he} and textual encodings \cite{open-set-detection-yang}, but remain limited to detection and classification. In contrast, we tackle \textit{metric depth estimation}, using retrieved priors to reconstruct geometry for underrepresented data, not just detect it.

\section{Method}
\label{sec:method}

We now describe our method. First, we detail the \textit{training procedure} in Sec.~\ref{sec:training}. During training, context images and corresponding depth maps are sourced either by retrieving samples containing similar underrepresented classes or via 3D augmentation of the input image (Sec.~\ref{sec:data preparation}). We then establish spatial correspondences between the input and context images (Sec.~\ref{sec:point matching}) to guide matched cross-attention within a dual-stream network that performs depth estimation (Sec.~\ref{sec:depth estimation network}). Second, we describe the \textit{inference procedure} in Sec.~\ref{sec:inference}. This follows the same pipeline as training, except that context is sourced exclusively via retrieval. Fig.~\ref{fig:system architecture} presents an overview of RAD.

\subsection{Training}
\label{sec:training}

\subsubsection{Context Sourcing}
\label{sec:data preparation}

Let \( T = \{(x_i, d_i)\}_{i=1}^N \) be the training dataset, where each \( x_i \) is an input image and \( d_i \) its corresponding ground-truth depth map. For each sample \( (x_t, d_t) \in T \), we construct a context \( C = \{(x_c^i, d_c^i)\}_{i=1}^M \), consisting of \( M \) context images \( x_c^i \) and their associated depth maps \( d_c^i \).
For a given train sample  \( (x_t, d_t) \in T \), the context \( C \) is generated using one of two randomly selected procedures: (1) uncertainty-aware image retrieval or (2) 3D augmentation, as detailed below.

\noindent \textbf{Uncertainty-aware Image Retrieval.} \quad
Let \(D_{pool} = \{(x_i, d_i)\}_{i=1}^L\) denote the context dataset from which context samples are retrieved, where \(x_i\) represents an image and \(d_i\) its corresponding ground-truth depth map (in our setting, we use the training set for \(D_{pool}\), i.e., $D_{pool}=T$). 
Given a training sample \((x_t, d_t) \in T\), our objective is to retrieve a subset of $M$ context samples, \(C \subseteq D_{pool}\) 
(from different scenes than that of \((x_t, d_t)\)), 
such that images in \(C\) contain objects similar to those underrepresented objects in \(x_t\).

\begin{figure}[tbh]
    \centering
    \includegraphics[width=\linewidth]{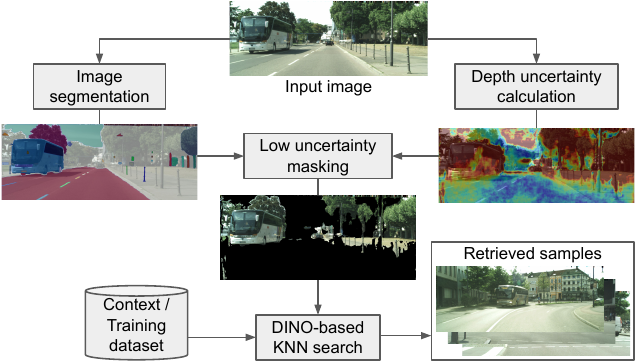}
    \caption{\textbf{Uncertainty-aware retrieval flow.} Pixel-wise depth uncertainty is calculated in parallel to image segmentation. We use these to keep only highly uncertain segments, masking the rest of the image. Given the masked image we retrieve relevant examples from the context/training set using DINO descriptors. 
    }
    \label{fig:retrieval flow}
    \vspace{-0.5cm}
\end{figure}

Conventional retrieval methods based on image descriptors tend to favor images with similar scene layouts. To address this limitation, we propose a custom retrieval strategy. Our central insight is that a depth estimation network trained on a dataset with representative distributions exhibits elevated uncertainty when encountering underrepresented object classes. This phenomenon arises from the weak priors encoded in the network’s parameters due to insufficient training data. We leverage this uncertainty signal to identify underrepresented objects within an image and guide the retrieval of relevant samples. Fig.~\ref{fig:retrieval flow} illustrates the proposed uncertainty-aware retrieval pipeline.

Following the methodology outlined in \cite{depth-uncertainty}, pixel-wise uncertainty is estimated by training a depth estimation network \(D\) on a representative dataset, applying it multiple times to perturbed versions of the input image, and analyzing the resulting standard deviation. Let \(x_t\) denote the training image, and let \(N(x_t)\) represent a noisy variant of \(x_t\), generated by injecting random Gaussian noise, with standard deviation $\sigma$. Let \(D(y)\) denote the estimated depth for some image \(y\), calculated by a frozen version of a depth estimation model that uses our backbone. Define \(\text{std}(X)\) as the pixel-wise standard deviation and \(\mathbb{E}(X)\) as the pixel-wise mean of a set of images \(X\). The pixel-wise uncertainty for \(x_t\), denoted \(\text{Uncertainty}(x_t)\), is computed as follows:
\begin{align}
    \text{Noisy}(x_t) &= \{N(x_t)\}_{i=1}^n \\
    \text{Uncertainty}(x_t) &= \frac{\text{std}(\{D(y) \mid y \in \text{Noisy}(x_t)\})}{\mathbb{E}(\{D(y) \mid y \in \text{Noisy}(x_t)\})}.
\end{align}

Given a training sample \((x_t, d_t) \in T\), we first compute \(U = \text{Uncertainty}(x_t)\) using DepthAnything v2 \cite{DepthAnything_V2} for \(D(\cdot)\), and generate a segmentation map \(S = \text{Segmentation}(x_t)\) using SAM2 \cite{sam2}. For each segment \(s \in S\), we compute the number of pixels exceeding a predefined uncertainty threshold \(h\): $p(s, U, h) = \left| \{x \mid x \in s \wedge U(x) > h\} \right|$.

To isolate regions of high uncertainty, we retain only segments \(s \in S\) for which at least \(q\%\) of pixels exceed threshold \(h\). Thus, we generate a masked image \(\tilde{x_t}\), as follows:
\begin{align}
    &Keep(U, S, q, h) = \{s \in S \mid p(s, U, h) > \frac{|s| \cdot q}{100}\} \\
    &\tilde{x_t}(x) = 
        \begin{cases}
            x_t(x), & \exists s \in Keep(U, S, q, h), \text{s.t. } x \in s \\
            0, & otherwise.
        \end{cases}
\end{align}

Finally, let \(\text{DINO}(x)\) represent the DINO v2 descriptor \cite{dinov2} of image \(x\), and define \(\text{KNN}(x, Y, M)\) as the set of \(M\) nearest neighbors to descriptor \(x\) within the descriptor set \(Y\). Then, the uncertainty-aware retrieved context \(C\) is:
\begin{align}
    \text{DINO}_{D_{pool}} &= \{\text{DINO}(x) \mid (x, d) \in D_{pool}\} \\
    W &= \text{KNN}(\text{DINO}(\tilde{x_t}), \text{DINO}_{D_{pool}}, M) \\
    C &= \{c \in D_{pool} \mid \text{DINO}(c) \in W\}.
    \label{eq:uncertainty-aware retrieval}
\end{align}

In our implementation, cosine similarity is used to compute nearest neighbors. In effect, we use a masked image as query to retrieve from a pool of non-masked images. Empirically, this showed to be viable as the DINO v2 embedding aggregates a thorough description of the image contents.

\noindent \textbf{3D Augmentation.} \quad
Training using only retrieved samples from a fixed context dataset leads to repetitive context for each training image across epochs. To promote robustness to varying contexts, we use a 3D augmentation method.

To promote robustness to varying contexts, we employ a 3D augmentation strategy. Given a training sample $(x_t, d_t) \in T$ and camera intrinsics $K$, we generate a synthetic context set $C = \{(x_c^i, d_c^i)\}_{i=1}^M$. We first back-project the input image to 3D points $X = \text{Backproj}(x_t, d_t, K)$. Then, we sample $M$ random camera poses $\{[R_i|t_i]\}_{i=1}^M$ and render the scene from these new viewpoints to obtain $x_c^i = \text{Proj}(X, K, [R_i|t_i])$ and $d_c^i = Z(X, K, [R_i|t_i])$, where $\text{Proj}$ and $Z$ denote the image and depth projection functions respectively. To avoid holes when projecting from a new point of view, we create a mesh (see supplementary).

\begin{figure}[tbh]
  \centering
  \begin{subfigure}[b]{\linewidth}
    \centering
    \includegraphics[width=\textwidth]{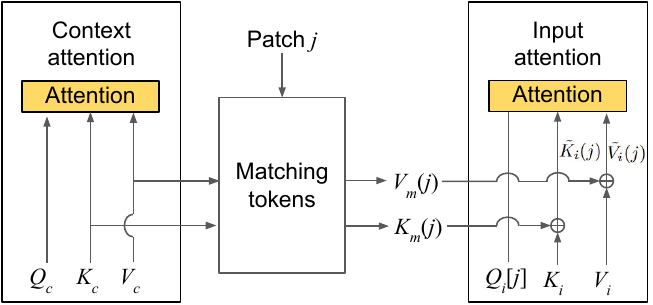}
    \caption{}
  \end{subfigure}
  
  \vspace{0em} %

  \begin{subfigure}[b]{\linewidth}
    \centering
    \includegraphics[width=\textwidth]{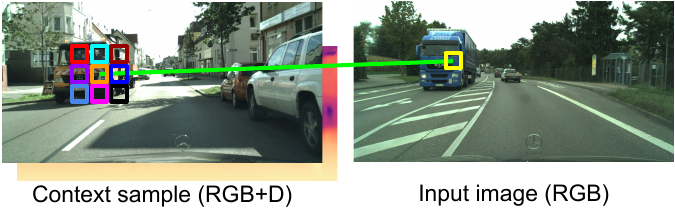}
    \caption{}
  \end{subfigure}
  \vspace{-0.6cm}

  \caption{\textbf{Matched Cross-Attention.} (a) illustrates the modified attention architecture designed to enable effective information transfer from the context stream to the input stream. For each token $j$ in the input image, with query vector $Q_i[j]$, attention is computed using key and value matrices formed by concatenating the input’s keys $(K_i$) and values ($V_i$) with the matched context keys ($K_m(j)$) and values ($V_m(j)$). These matched matrices are constructed by selecting $j$'s matching context tokens from the full context matrices $K_c$ and $V_c$, respectively. (b) shows that matching tokens are defined as those located within a spatial neighborhood surrounding the matched point of $j$ in the context image. %
  }
  \label{fig:network architecture}
  \vspace{-0.4cm}
\end{figure}

\subsubsection{Spatial Correspondence}
\label{sec:point matching}
Once training sample $(x_t, d_t) \in T$ is complemented with context $C$, we search for precise spatial correspondences.

When uncertainty-aware retrieval is used, we apply an off-the-shelf point matching method to identify semantic correspondences between training image $x_t$ and context images $\{ x_c^i\}_{i=1}^M$. Specifically, we use LightGlue \cite{lightglue}, which efficiently detects matches. When 3D augmentation is used, correspondences are calculated analytically through projective geometry. These correspondences then guide cross-attention at matched locations, as detailed next in Sec.~\ref{sec:depth estimation network}.

\subsubsection{Dual-stream Depth Estimation Network}
\label{sec:depth estimation network}

Our network builds upon the DepthAnything v2 architecture \cite{DepthAnything_V2}, utilizing the same Vision Transformer (ViT) encoder and DPT decoder \cite{dpt} for dense depth estimation.

The network is formed by two parallel ViT encoders, one for the RGB input and one for the RGB+D context sample. To incorporate depth into the context stream, a fourth channel is added to the projection operation at the beginning of its ViT encoder. Let $(x_c,d_c)\in C$ be a context sample, with $x_c \in \mathbb{R}^{H \times W \times 3}$  and $d_c \in \mathbb{R}^{H \times W \times 1}$, where $W$ and $H$ are the image's width and height, respectively. The projection operation in the context stream is defined as a function $P:\mathbb{R}^{H \times W \times 4} \rightarrow \mathbb{R}^{H_P \times W_P \times c}$, where $H_P\times W_P$ are the number of patches extracted, and $c$ is the dimensionality of the projected patch embeddings. The input to $P$ is constructed by concatenating $x_c$ and $d_c$ along the channel axis, yielding the projected representation $P(\mathrm{concat}(x_c,d_c))$.

The input and context streams are linked via cross-attention at matching points. The context stream processes tokens using standard self-attention, while the input stream integrates information from the context stream through matched cross-attention detailed below. The architecture for both streams is the same as DepthAnything v2, except for the context encoder adaptation described above and the attention mechanism in each block of the input stream. Both streams run in parallel. 
Our training objectives are identical to DepthAnything v2, aiming to predict the depth of the input stream.  Note that these design choices are general, making our method easily adaptable to different backbones and objectives. 
The overall network architecture, training and implementation details, as well as memory and inference-time analysis, are provided in the supplementary. 

\noindent \textbf{Matched Cross-Attention.} \quad 
To enable effective information exchange between the context and input streams, we propose a modified attention mechanism called  \textit{matched cross-attention}. As depicted in Fig.~\ref{fig:network architecture}(a), this mechanism allows the input stream to integrate contextual information by augmenting its attention computation as follows.

Let $Q_i$, $K_i$, and $V_i$ be the queries, keys, and values matrices for the input feature map, where $j$ represents the position of a given token in this feature map. 
Likewise, let $Q_c$, $K_c$ and $V_c$ be the queries, keys, and values matrices for the corresponding context feature map. 
For any of the matrices $X$  defined above, by $X[a]$ we refer to the vector in $X$ corresponding to token $a$. Additionally, let \( \text{matching\_tokens}(j) \) represent the set of context tokens matched to token corresponding to \( j \) in the input image. Then, the attention for token $j$ in the input feature map, denoted \( A_j \), is:
\begin{align}
    K_m(j) &= \left\{ K_c[a] \mid a \in \text{matching\_tokens}(j) \right\} \\
    V_m(j) &= \left\{ V_c[a] \mid a \in \text{matching\_tokens}(j) \right\} \\
    \tilde{K_i}(j) &= \operatorname{concat}(K_i, K_m(j)) \\
    \tilde{V_i}(j) &= \operatorname{concat}(V_i, V_m(j)) \\
    A_j &= \operatorname{attention}(Q_i[j], \tilde{K}_i(j), \tilde{V}_i(j))
    \label{eq:matched cross-attention}
\end{align}
where attention is the standard attention mechanism. %

As illustrated in Fig.~\ref{fig:network architecture}(b), the matching tokens associated with token \( j \) are defined as those located within a spatial neighborhood surrounding its matched point in the context sample. This approach preserves spatial locality while facilitating cross-image correspondence. If more than one image is retrieved, we aggregate context tokens through all of them. Importantly, the set $\text{matching\_tokens}(j)$ varies for each input token $j$, resulting in distinct key and value matrices for each attention computation.

\begin{table*}[tbh]
    \centering
        \caption{\textbf{Underrepresented classes evaluation.} Comparison of depth estimation performance on \textbf{underrepresented classes} across NYU, KITTI, and Cityscapes datasets. Top: fine-tuning methods, middle: zero-shot methods, bottom: our method. \dag: ZoeDepth is zero-shot for Cityscapes, but fine-tuned for NYU and KITTI. Best result in bold, second best underlined. Missing values correspond to methods without available code or trained model and that have not reported the relevant values in the literature.
    }
    \footnotesize
    \setlength{\tabcolsep}{2.7pt}
    \begin{tabular}{l|cccccc|cccccc|cccccc}
        \toprule
        & \multicolumn{6}{c|}{NYU} & \multicolumn{6}{c|}{KITTI} & \multicolumn{6}{c}{Cityscapes} \\
        Method & \multicolumn{3}{c}{$\uparrow$} & \multicolumn{3}{c|}{$\downarrow$}
        & \multicolumn{3}{c}{$\uparrow$} & \multicolumn{3}{c|}{$\downarrow$}
        & \multicolumn{3}{c}{$\uparrow$} & \multicolumn{3}{c}{$\downarrow$} \\
        \cmidrule(lr){2-4} \cmidrule(lr){5-7} \cmidrule(lr){8-10} \cmidrule(lr){11-13} \cmidrule(lr){14-16} \cmidrule(lr){17-19}
        & $\delta_1$ & $\delta_2$ & $\delta_3$ & AbsRel & RMS & $\mathrm{Log_{10}}$ 
        & $\delta_1$ & $\delta_2$ & $\delta_3$ & AbsRel & RMS & $\mathrm{RMS_{log}}$ 
        & $\delta_1$ & $\delta_2$ & $\delta_3$ & AbsRel & RMS & $\mathrm{RMS_{log}}$ \\
        \midrule
        AdaBins \cite{adabins} 
        & 81.0 & 88.3 & 93.7 & 0.173 & 0.522 & 0.053 
        & 87.2 & 95.3 & 98.7 & 0.143 & 4.567 & 0.193 
        & — & — & — & — & — & — \\
        
        NewCRF \cite{newcrf} 
        & 80.3 & 89.7 & 95.4 & 0.169 & 0.497 & 0.049 
        & 89.1 & 96.7 & \underline{99.1} & 0.121 & 4.247 & 0.166 
        & — & — & — & — & — & — \\
        
        DepthAnyV2 \cite{DepthAnything_V2} 
        & 91.4 & 98.2 & 99.8 & 0.089 & 0.327 & 0.049 
        & 92.3 & \underline{97.8} & \underline{99.1} & \underline{0.083} & 2.893 & \underline{0.086} 
        & 89.2 & 94.7 & \textbf{99.1} & 0.110 & 7.804 & 0.166 \\
        
        \midrule
        ZoeDepth$^\dag$ \cite{ZoeDepth} 
        & 90.9 & 98.1 & 99.6 & 0.098 & 0.334 & 0.043 
        & 88.6 & 96.9 & 98.4 & 0.116 & 3.908 & 0.140 
        & 40.9 & 66.7 & 79.3 & 0.314 & 17.051 & 0.496 \\
        
        DepthPro \cite{DepthPro} 
        & 89.6 & 97.6 & 99.1 & 0.106 & 0.375 & 0.048 
        & 77.9 & 97.7 & 99.3 & 0.160 & 2.698 & 0.170 
        & 47.5 & 87.8 & 98.0 & 0.224 & 11.973 & 0.285 \\
        
        Metric3Dv2 \cite{metric3d} 
        & 92.4 & 97.6 & 99.4 & 0.095 & 0.331 & \textbf{0.039} 
        & 89.8 & 97.7 & \textbf{99.4} & 0.121 & \underline{2.577} & 0.141 
        & \underline{92.1} & \textbf{97.1} & 98.5 & \underline{0.097} & 7.063 & \underline{0.152} \\
        
        UniDepthV2 \cite{UniDepth_V2} 
        & 92.8 & 98.1 & 99.3 & 0.092 & 0.329 & \textbf{0.039} 
        & 90.5 & 98.2 & \textbf{99.4} & 0.111 & 4.059 & 0.130 
        & 88.3 & 96.4 & 98.2 & 0.136 & 7.777 & 0.171 \\
        
        \midrule
        RAD-Small 
        & 95.3 & 98.5 & 99.1 & 0.084 & 0.321 & 0.057 
        & 92.2 & 95.9 & 98.0 & 0.143 & 3.943 & 0.097 
        & 86.7 & 96.6 & 98.5 & 0.133 & 5.573 & 0.168 \\
        
        RAD-Base 
        & \underline{96.7} & \underline{99.4} & \underline{99.7} & \underline{0.072} & \underline{0.299} & 0.053 
        & \underline{95.1} & 97.6 & 98.8 & 0.084 & 3.107 & 0.088 
        & 88.5 & 96.9 & \underline{99.0} & 0.127 & \underline{5.191} & 0.158 \\
        
        RAD-Large 
        & \textbf{97.5} & \textbf{99.5} & \textbf{99.9} & \textbf{0.063} & \textbf{0.288} & \underline{0.040} 
        & \textbf{96.6} & \textbf{98.6} & \textbf{99.4} & \textbf{0.072} & \textbf{2.498} & \textbf{0.052} 
        & \textbf{93.5} & \underline{97.0} & \textbf{99.1} & \textbf{0.090} & \textbf{5.083} & \textbf{0.150} \\
        \bottomrule
    \end{tabular}
    \label{tab:underrepresented classes - all datasets}
\end{table*}

\subsection{Inference}
\label{sec:inference}
At inference time, given an input image \( I \), RAD employs uncertainty-aware retrieval to assemble context \( C = \{(x_c^i, d_c^i)\}_{i=1}^M \), as defined in Eq. \ref{eq:uncertainty-aware retrieval} (Sec. \ref{sec:data preparation}). Then, point correspondences between \( I \) and the retrieved context images \( \{x_c^i\}_{i=1}^M \) are computed following the procedure outlined in Sec. \ref{sec:point matching}. Finally, depth estimation is performed using the network described in Sec. \ref{sec:depth estimation network}, incorporating matched cross-attention as detailed in Eq. \ref{eq:matched cross-attention}. 

\begin{table*}[tbh]
    \centering
        \caption{\textbf{All classes evaluation.} Comparison of depth estimation performance on \textbf{all classes} across NYU, KITTI, and Cityscapes datasets. Top: fine-tuning methods, middle: zero-shot methods, bottom: our method. Bold indicates best (for $\delta$ columns: largest; for lower-is-better columns: smallest). Underline indicates second best. \dag: ZoeDepth is zero-shot for Cityscapes, but fine-tuned for NYU and KITTI. Best result in bold, second best underlined. Missing values correspond to methods without available code or trained model and that have not reported the relevant values in the literature.}
    \footnotesize
    \setlength{\tabcolsep}{2.7pt}
    \begin{tabular}{l|cccccc|cccccc|cccccc}
        \toprule
        & \multicolumn{6}{c|}{NYU} & \multicolumn{6}{c|}{KITTI} & \multicolumn{6}{c}{Cityscapes} \\
        Method & \multicolumn{3}{c}{$\uparrow$} & \multicolumn{3}{c|}{$\downarrow$}
        & \multicolumn{3}{c}{$\uparrow$} & \multicolumn{3}{c|}{$\downarrow$}
        & \multicolumn{3}{c}{$\uparrow$} & \multicolumn{3}{c}{$\downarrow$} \\
        \cmidrule(lr){2-4} \cmidrule(lr){5-7} \cmidrule(lr){8-10} \cmidrule(lr){11-13} \cmidrule(lr){14-16} \cmidrule(lr){17-19}
        & $\delta_1$ & $\delta_2$ & $\delta_3$ & AbsRel & RMS & $\mathrm{Log_{10}}$ 
        & $\delta_1$ & $\delta_2$ & $\delta_3$ & AbsRel & RMS & $\mathrm{RMS_{log}}$ 
        & $\delta_1$ & $\delta_2$ & $\delta_3$ & AbsRel & RMS & $\mathrm{RMS_{log}}$ \\
        \midrule
        BTS \cite{bts}
        & 88.5 & 97.8 & 99.4 & 0.109 & 0.391 & 0.046
        & 96.2 & 99.4 & 99.8 & 0.056 & 2.430 & 0.089
        & — & — & — & — & — & — \\

        AdaBins \cite{adabins}
        & 90.3 & 98.4 & 99.7 & 0.103 & 0.364 & 0.044
        & 96.4 & 99.5 & \underline{99.9} & 0.058 & 2.360 & 0.088
        & — & — & — & — & — & — \\

        NewCRF \cite{newcrf}
        & 92.1 & 99.1 & 99.8 & 0.095 & 0.333 & 0.040
        & 97.4 & \underline{99.7} & \underline{99.9} & 0.052 & 2.129 & 0.079
        & — & — & — & — & — & — \\

        IEBins \cite{iebins}
        & 93.6 & 99.2 & 99.8 & 0.087 & 0.314 & 0.038
        & 97.1 & 99.6 & \underline{99.9} & 0.050 & 2.011 & 0.075
        & — & — & — & — & — & — \\

        iDisc \cite{idisc}
        & 93.8 & 99.2 & 99.8 & 0.086 & 0.313 & 0.037
        & 97.5 & \underline{99.7} & \underline{99.9} & 0.050 & 2.070 & 0.077
        & — & — & — & — & — & — \\

        DepthAnyV2 \cite{DepthAnything_V2}
        & 98.4 & \textbf{99.8} & \textbf{100} & 0.056 & 0.206 & 0.024
        & 98.3 & \textbf{99.8} & \textbf{100} & 0.045 & 1.861 & 0.067
        & 93.5 & \underline{98.4} & \underline{99.4} & \underline{0.080} & \underline{4.901} & \underline{0.143} \\

        \midrule
        ZoeDepth$^{\dag}$ \cite{ZoeDepth}
        & 95.1 & 99.4 & \underline{99.9} & 0.077 & 0.282 & 0.033
        & 97.1 & 99.6 & \underline{99.9} & 0.054 & 2.281 & 0.082
        & 51.1 & 73.3 & 82.2 & 0.292 & 12.182 & 0.474 \\
        
        DepthPro \cite{DepthPro}
        & 91.8 & 98.0 & 99.2 & 0.099 & 0.387 & 0.045
        & 83.5 & 97.8 & 99.3 & 0.141 & 3.375 & 0.157
        & 70.8 & 93.3 & 97.7 & 0.169 & 8.020 & 0.255 \\

        Metric3Dv2 \cite{metric3d}
        & \textbf{98.9} & \textbf{99.8} & \textbf{100} & \underline{0.047} & \underline{0.183} & \textbf{0.020}
        & \underline{98.5} & \textbf{99.8} & \textbf{100} & 0.044 & 1.990 & \underline{0.064}
        & \textbf{94.9} & 98.0 & 98.8 & 0.087 & 5.404 & 0.161 \\

        UniDepthV2 \cite{UniDepth_V2}
        & \underline{98.8} & \textbf{99.8} & \textbf{100} & \textbf{0.046} & \textbf{0.180} & \textbf{0.020}
        & \textbf{98.9} & \textbf{99.8} & \underline{99.9} & \textbf{0.037} & \textbf{1.710} & \textbf{0.061}
        & 87.0 & 97.7 & 98.8 & 0.175 & 5.948 & 0.206 \\

        \midrule
        RAD-Small
        & 96.8 & \underline{99.7} & \underline{99.9} & 0.066 & 0.233 & 0.028
        & 97.7 & \underline{99.7} & \underline{99.9} & 0.051 & 2.311 & 0.077
        & 90.6 & 97.8 & 99.3 & 0.092 & 5.375 & 0.154 \\

        RAD-Base
        & 98.0 & \underline{99.7} & \textbf{100} & 0.056 & 0.211 & \underline{0.023}
        & 98.1 & \textbf{99.8} & \textbf{100} & 0.046 & 1.913 & 0.070
        & 92.1 & 98.3 & \underline{99.4} & 0.081 & 4.980 & \underline{0.143} \\

        RAD-Large
        & \underline{98.8} & \textbf{99.8} & \textbf{100} & 0.048 & 0.192 & \textbf{0.020}
        & \textbf{98.9} & \textbf{99.8} & \textbf{100} & \underline{0.043} & \underline{1.801} & 0.065
        & \underline{93.8} & \textbf{98.7} & \textbf{99.7} & \textbf{0.078} & \textbf{4.891} & \textbf{0.140} \\
        \bottomrule
    \end{tabular}
    \label{tab:all classes - all datasets}
    \vspace{-0.4cm}
\end{table*}

\usetikzlibrary{positioning,spy}

\begin{figure*}[htbp]
  \centering
  \renewcommand{\arraystretch}{0}  %
  \begin{tabular}{@{}c@{}c@{}c@{}c@{}c@{}c@{}}
    \includegraphics[width=0.166\textwidth]{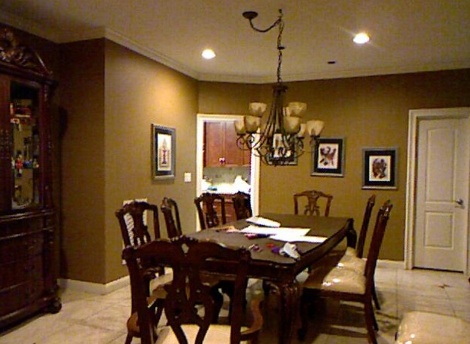} &
    \includegraphics[width=0.166\textwidth]{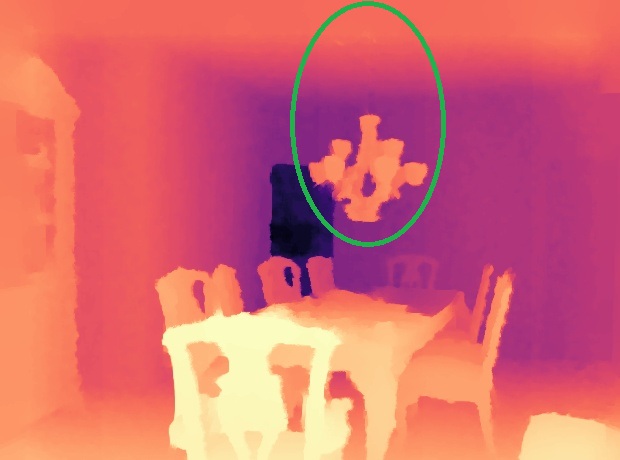} &
    \includegraphics[width=0.166\textwidth]{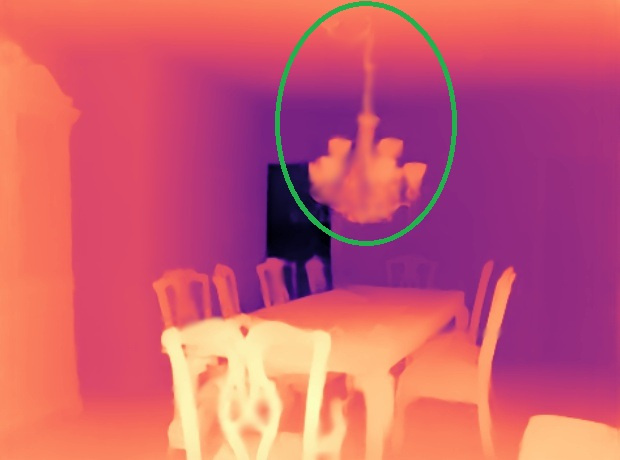} &
    \includegraphics[width=0.166\textwidth]{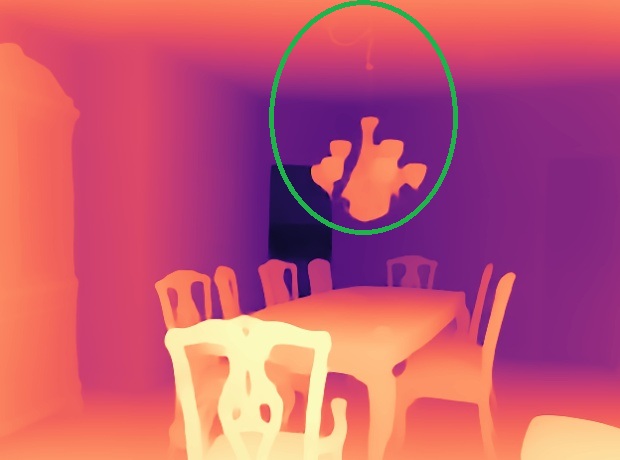} &
    \includegraphics[width=0.166\textwidth]{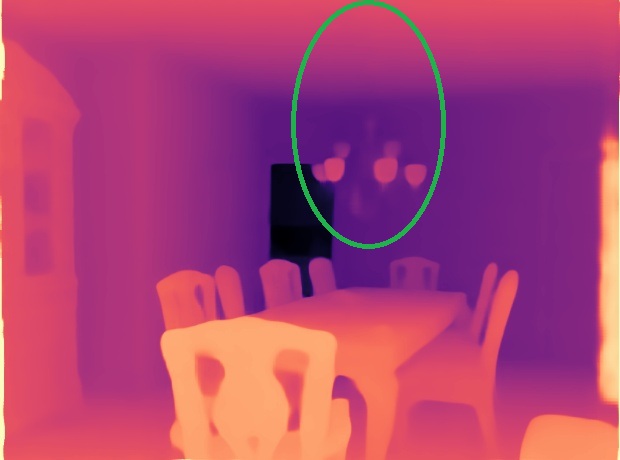} &
    \includegraphics[width=0.166\textwidth]{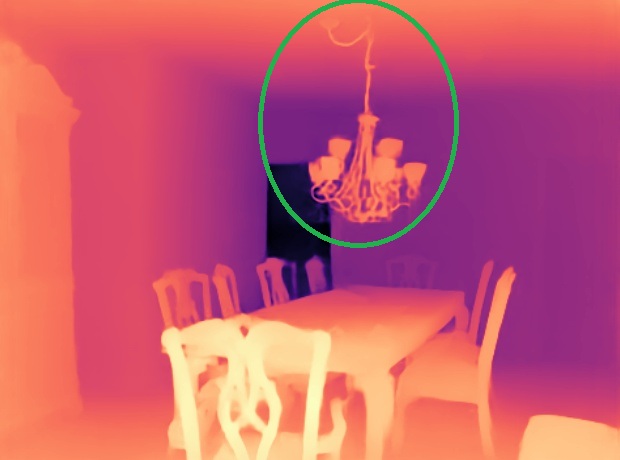} \\
    \includegraphics[width=0.166\textwidth]{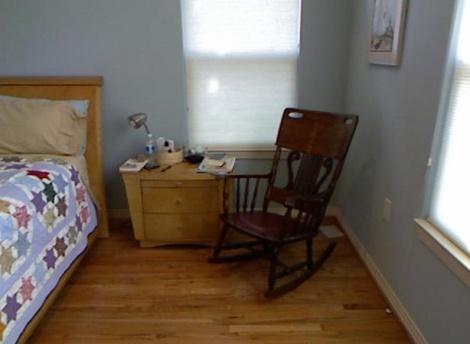} &
    \includegraphics[width=0.166\textwidth]{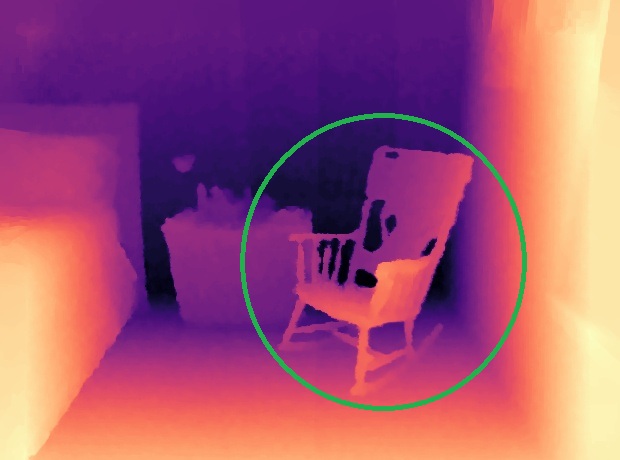} &
    \includegraphics[width=0.166\textwidth]{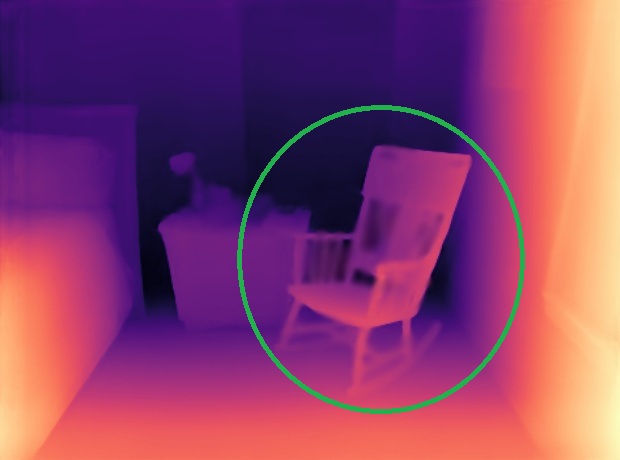} &
    \includegraphics[width=0.166\textwidth]{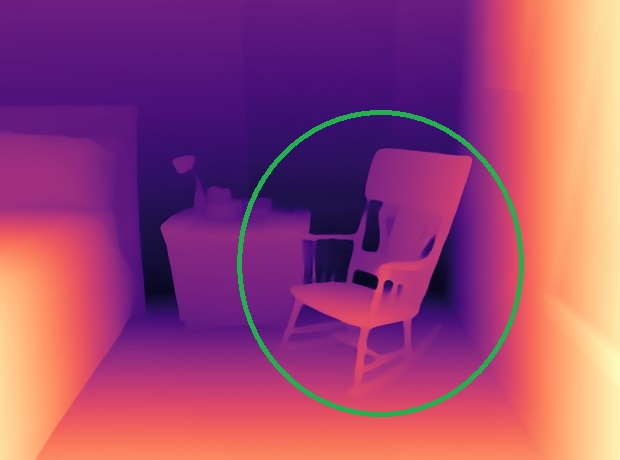} &
    \includegraphics[width=0.166\textwidth]{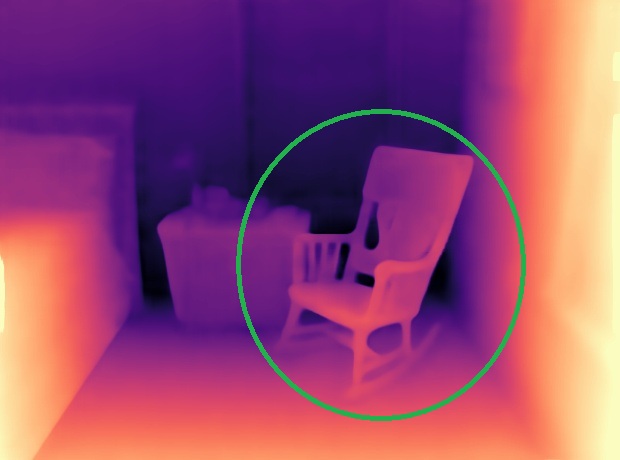} &
    \includegraphics[width=0.166\textwidth]{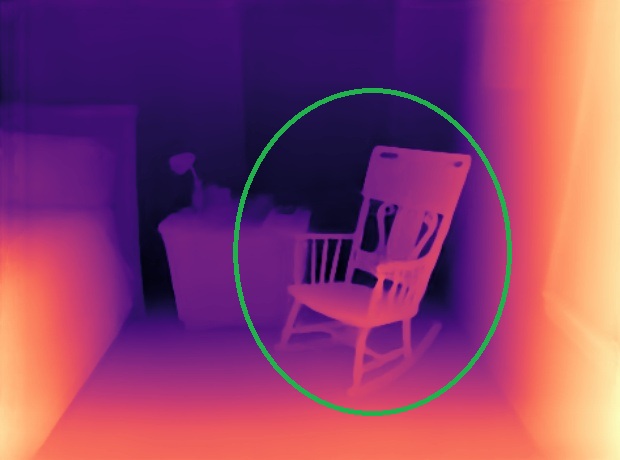} \\
    \includegraphics[width=0.166\textwidth]{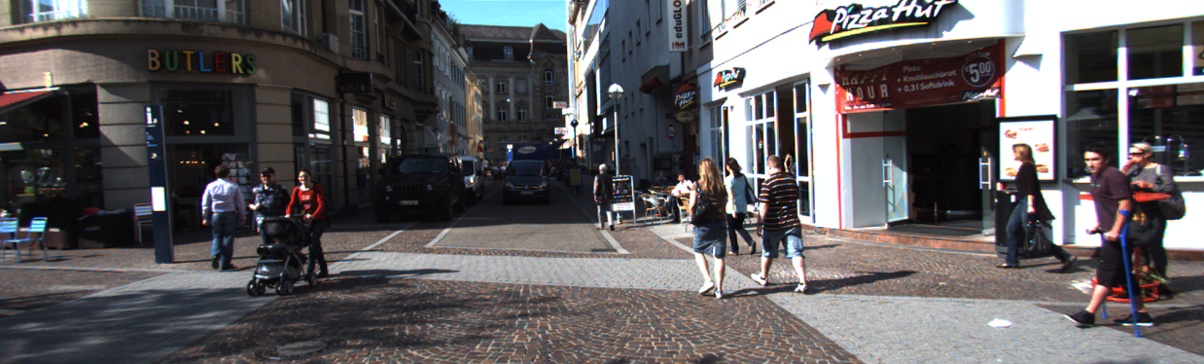} &
    \includegraphics[width=0.166\textwidth]{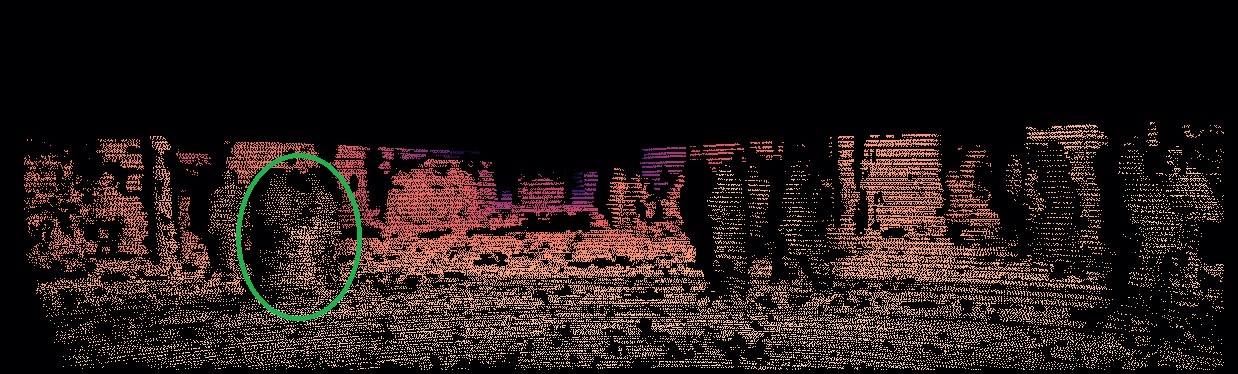} &
    \includegraphics[width=0.166\textwidth]{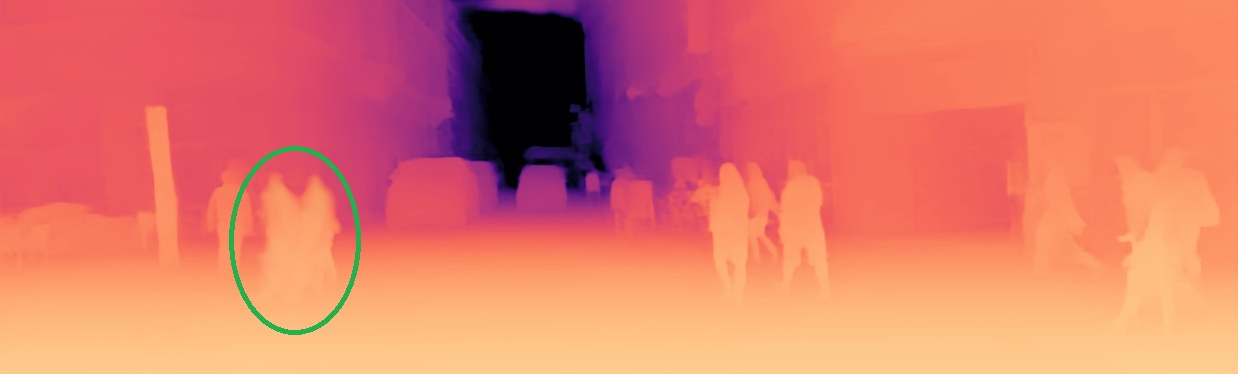} &
    \includegraphics[width=0.166\textwidth]{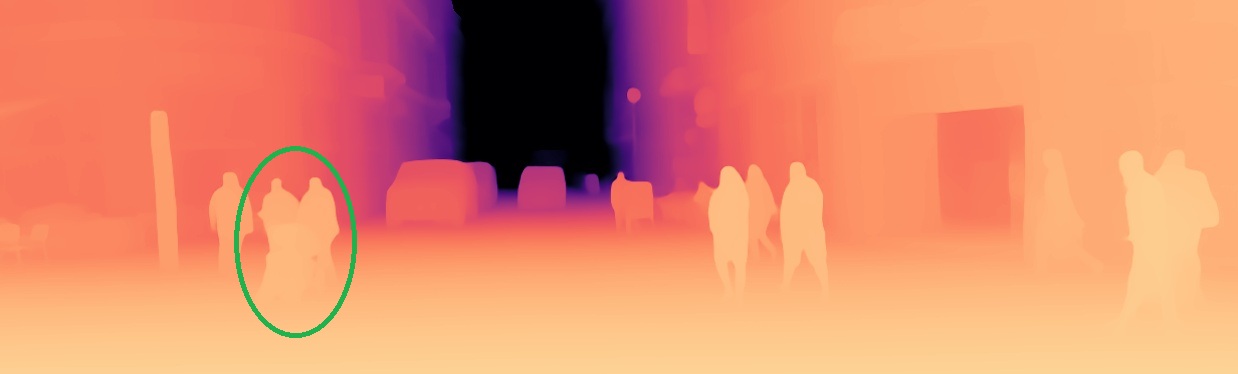} &
    \includegraphics[width=0.166\textwidth]{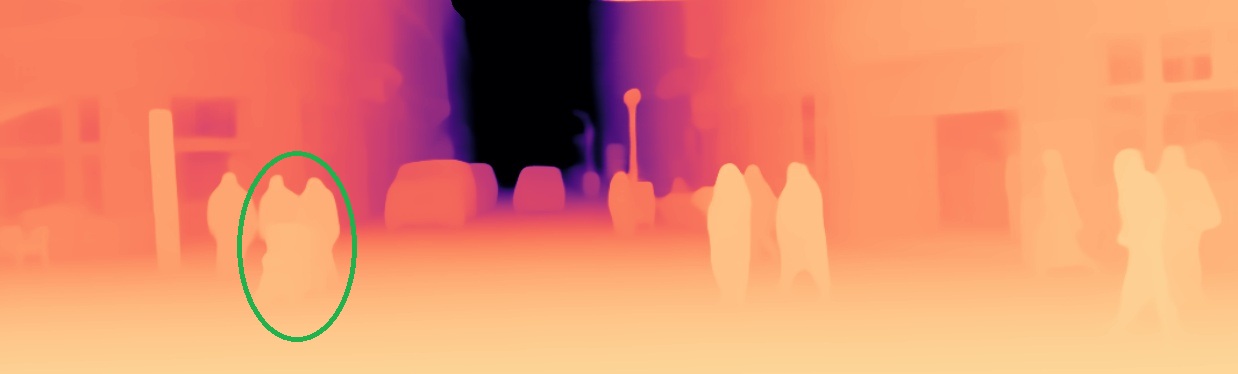} &
    \includegraphics[width=0.166\textwidth]{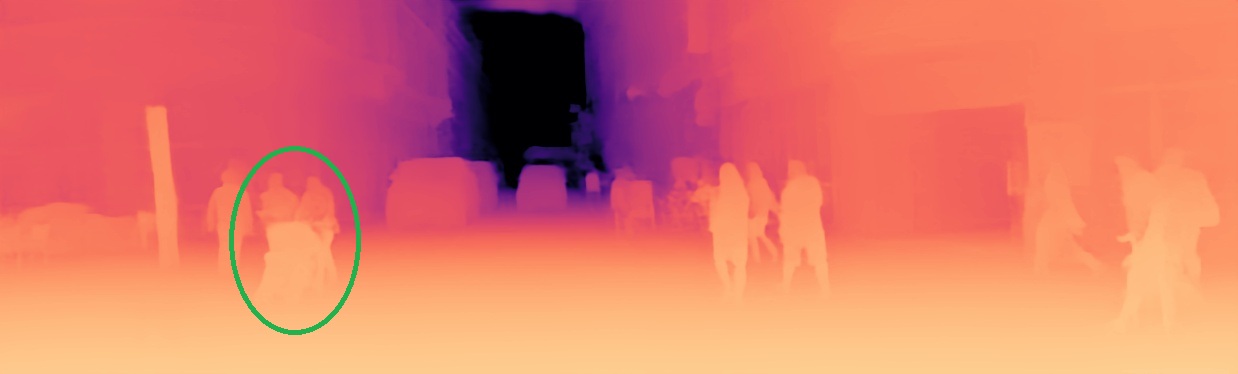} \\
    \includegraphics[width=0.166\textwidth]{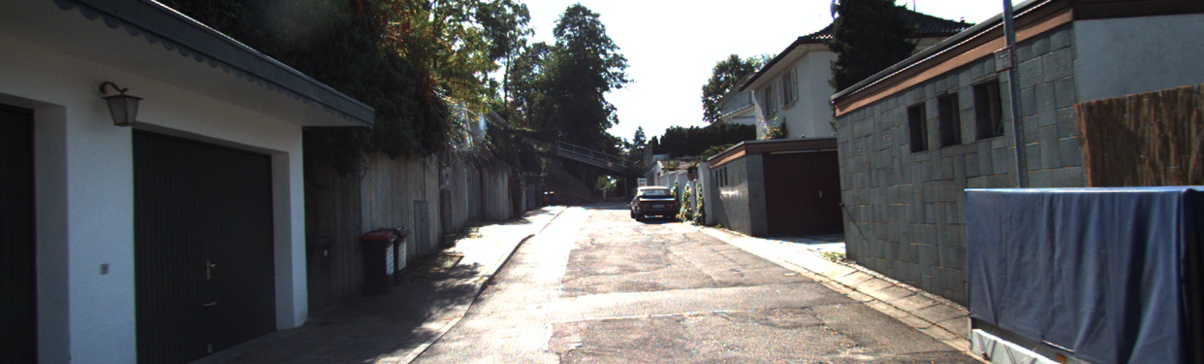} &
    \includegraphics[width=0.166\textwidth]{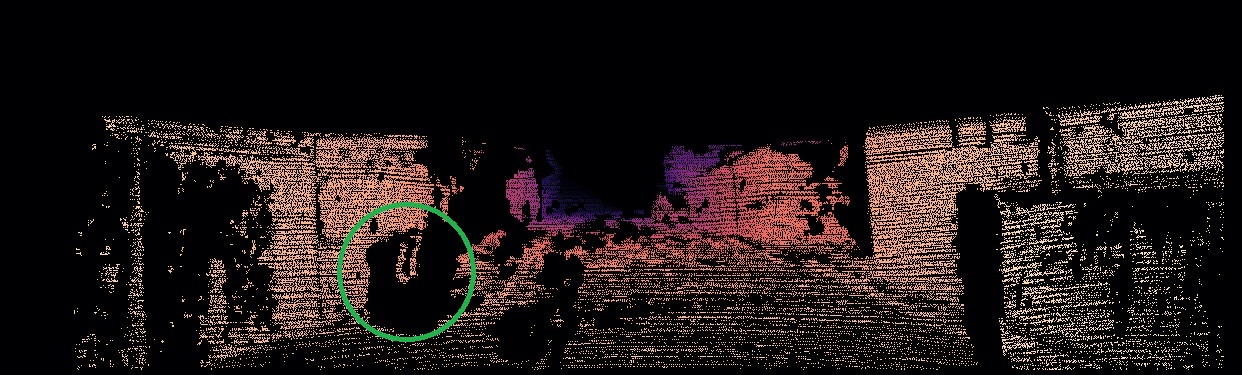} &
    \includegraphics[width=0.166\textwidth]{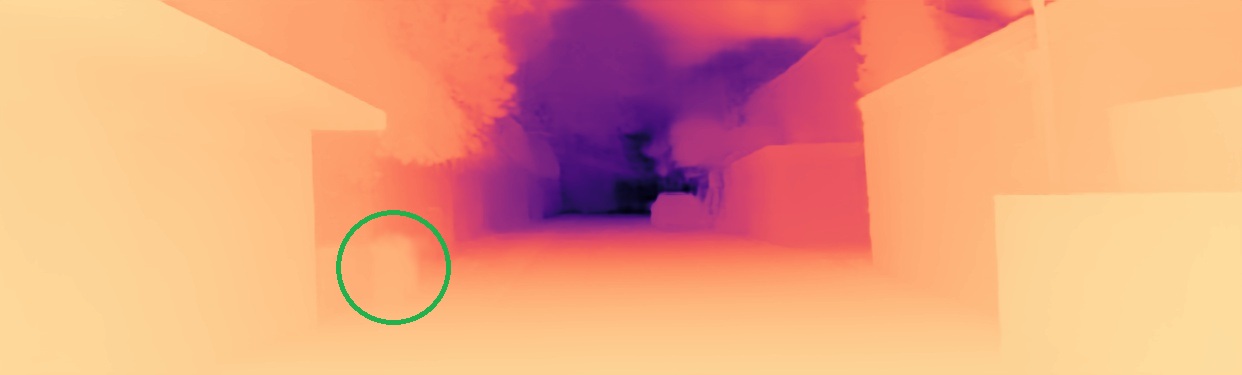} &
    \includegraphics[width=0.166\textwidth]{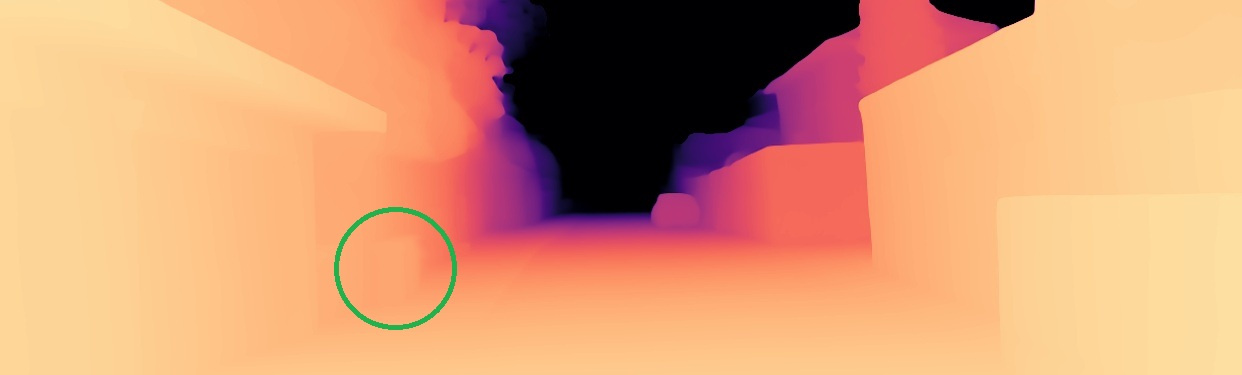} &
    \includegraphics[width=0.166\textwidth]{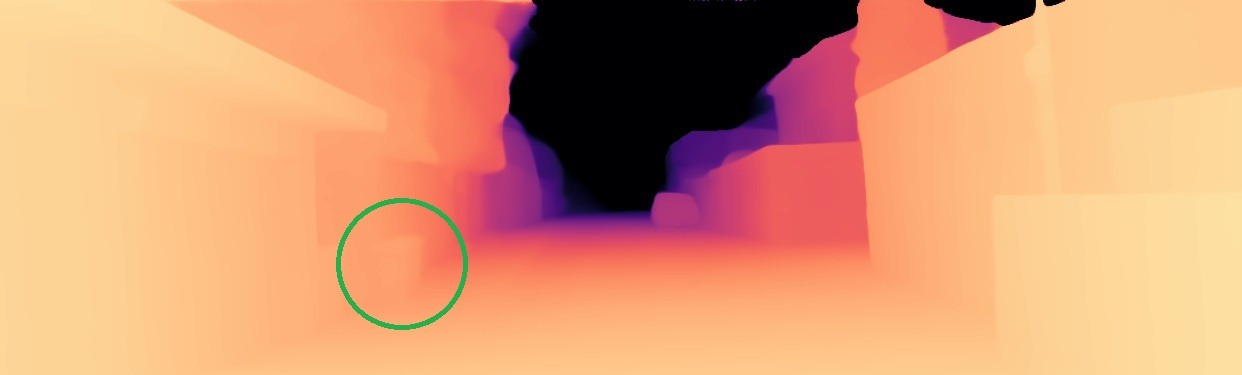} &
    \includegraphics[width=0.166\textwidth]{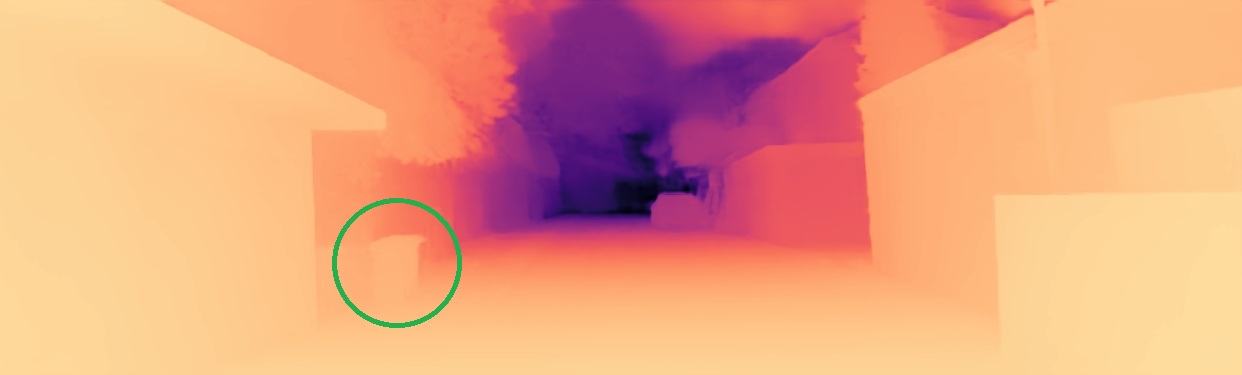} \\
    \includegraphics[width=0.166\textwidth]{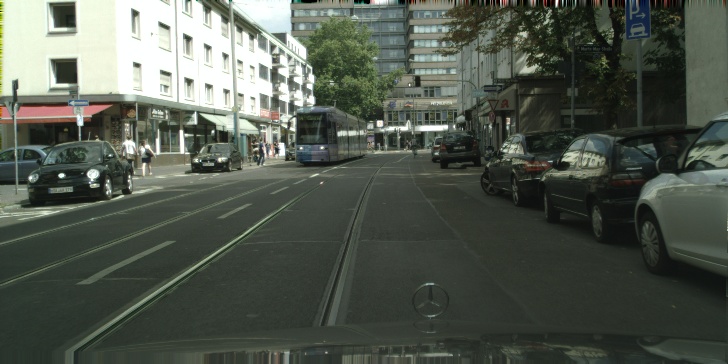} &
    \includegraphics[width=0.166\textwidth]{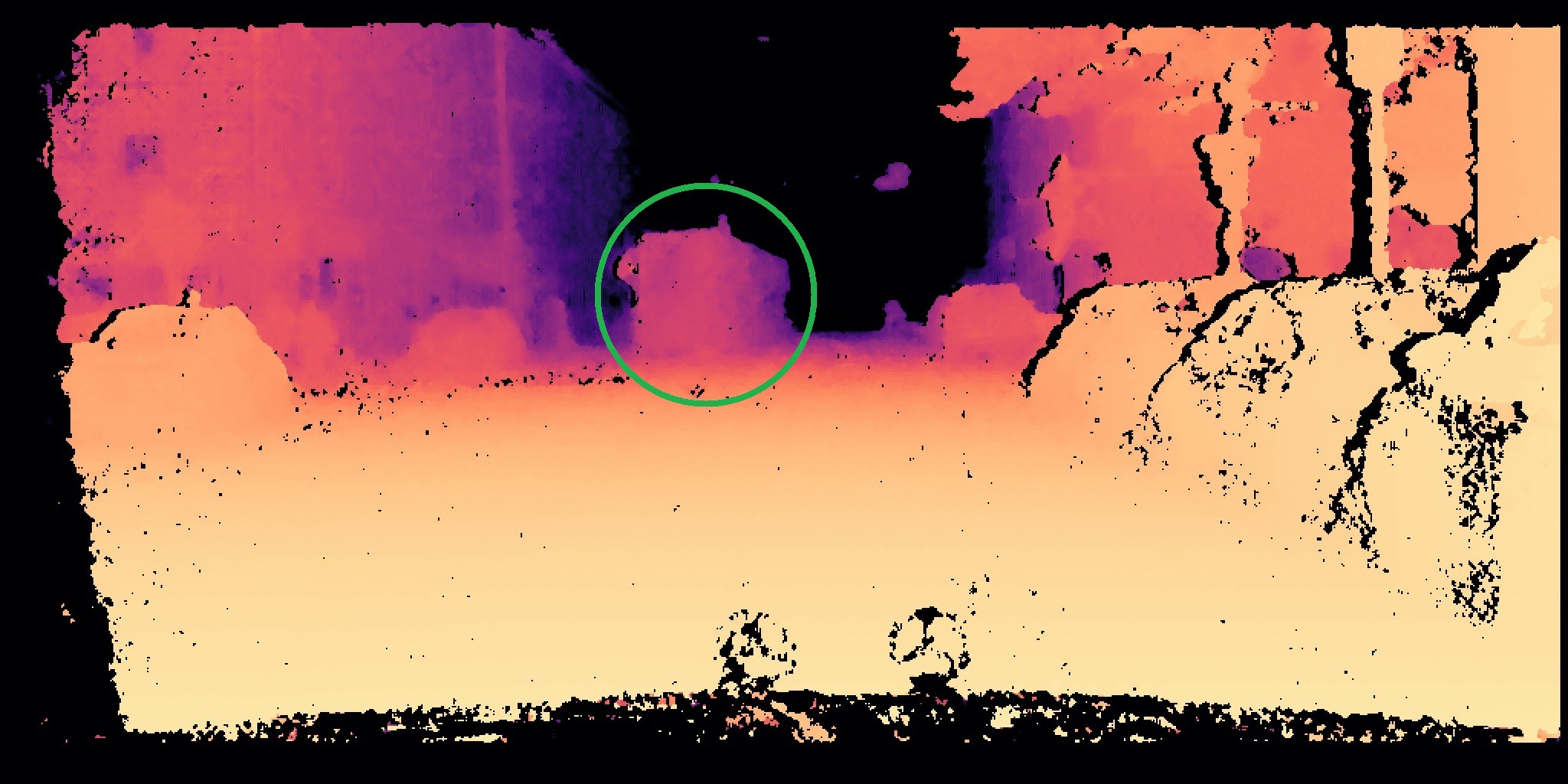} &
    \includegraphics[width=0.166\textwidth]{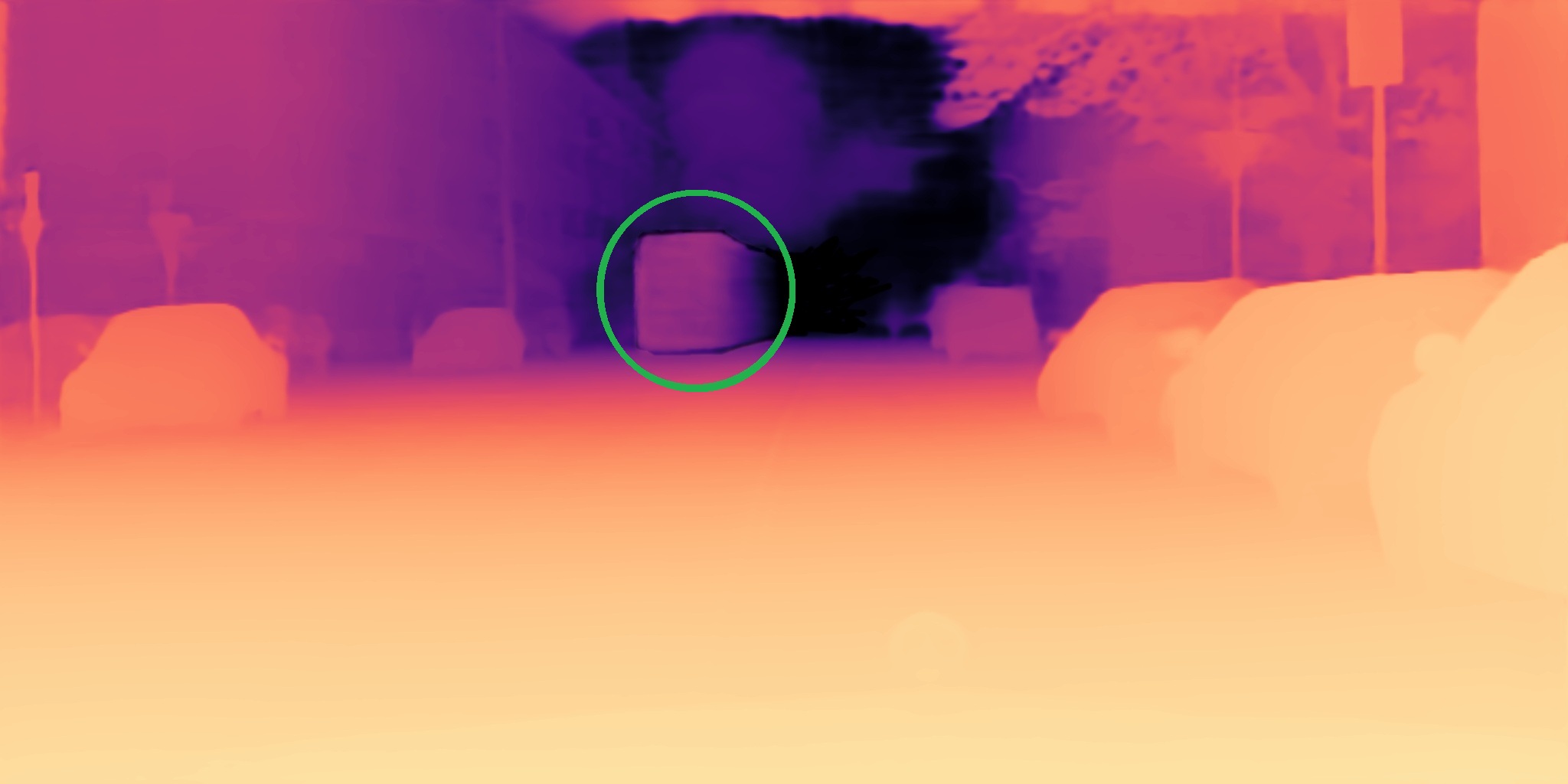} &
    \includegraphics[width=0.166\textwidth]{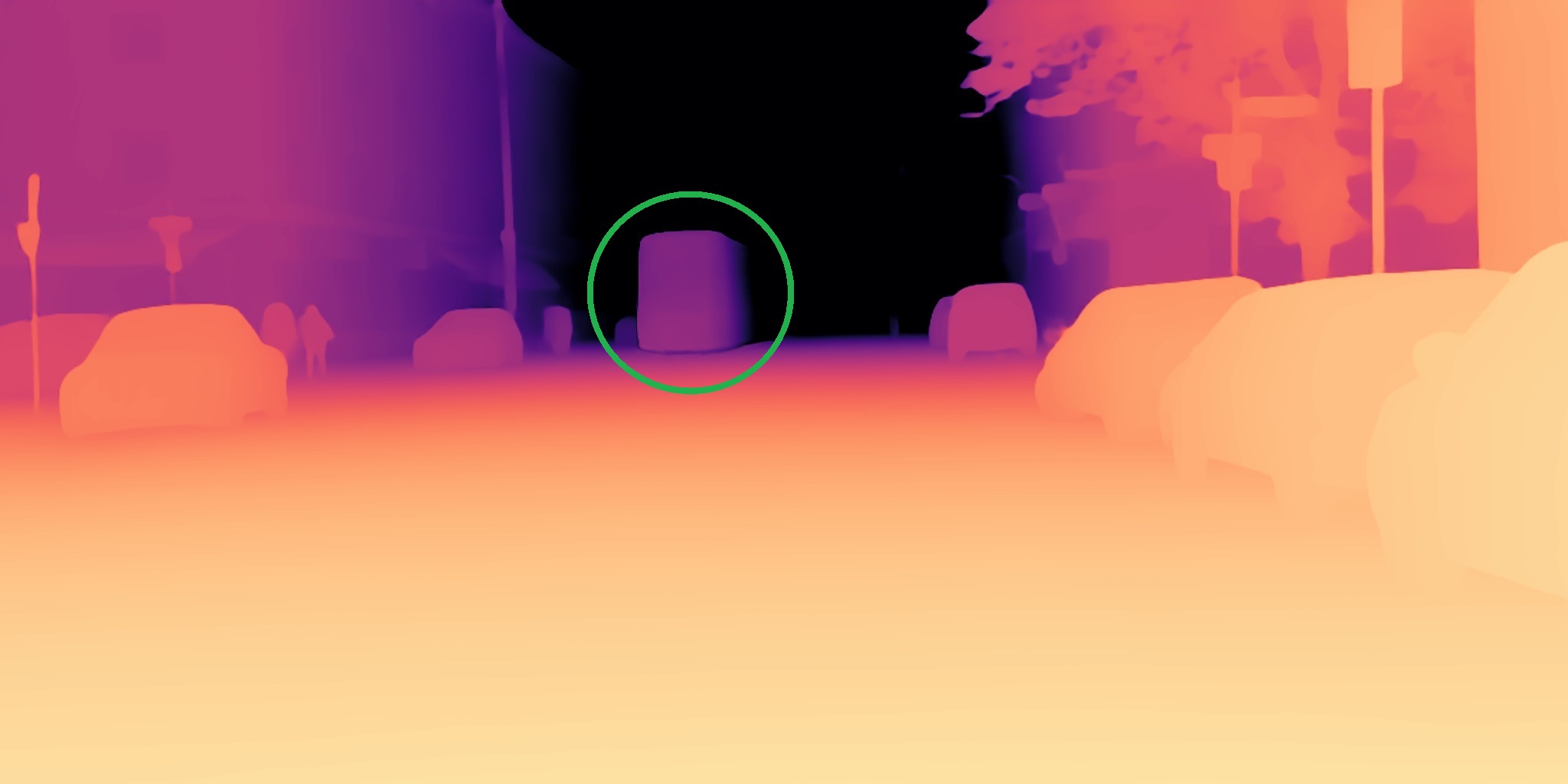} &
    \includegraphics[width=0.166\textwidth]{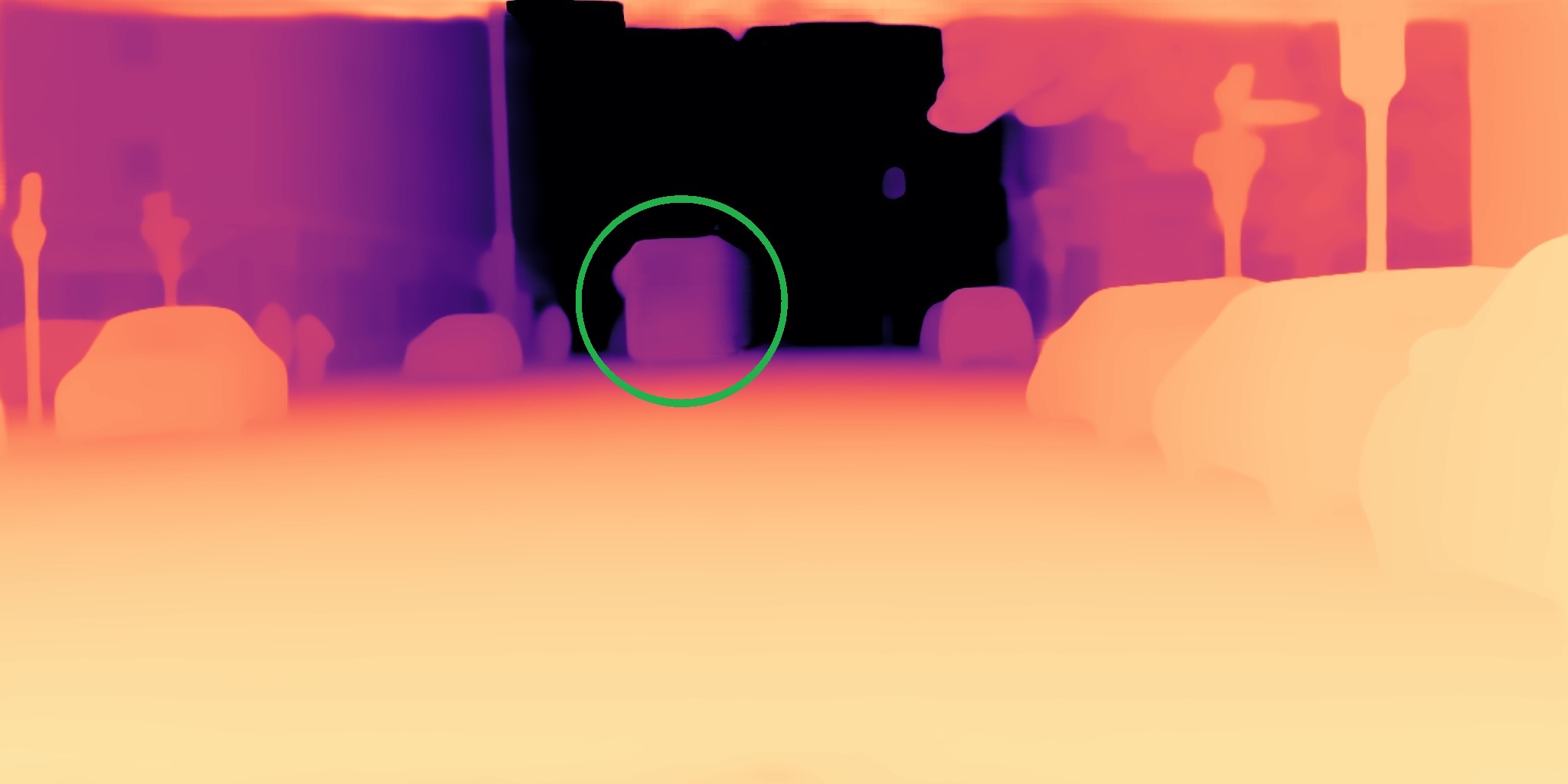} &
    \includegraphics[width=0.166\textwidth]{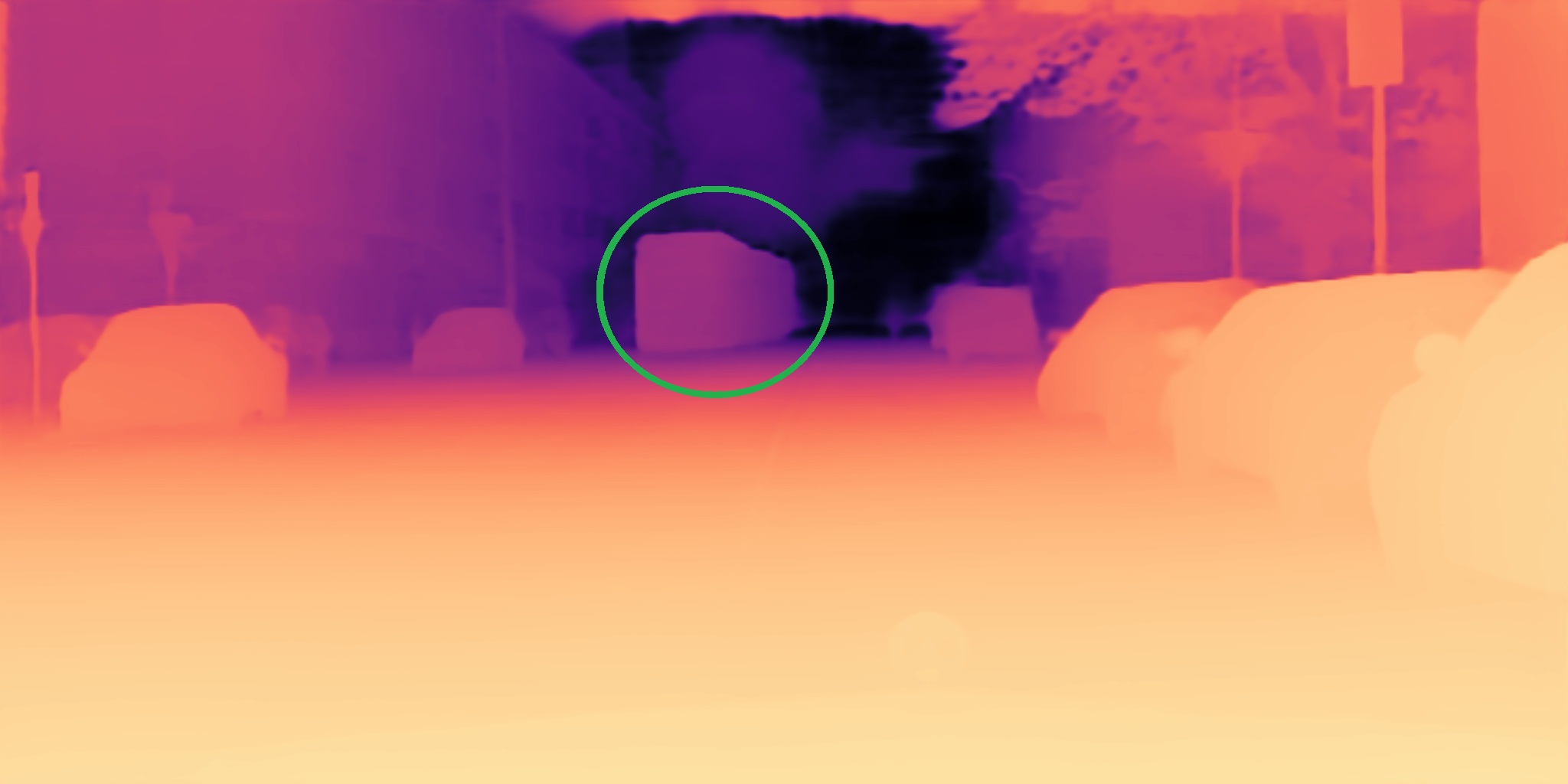} \\
    \includegraphics[width=0.166\textwidth]{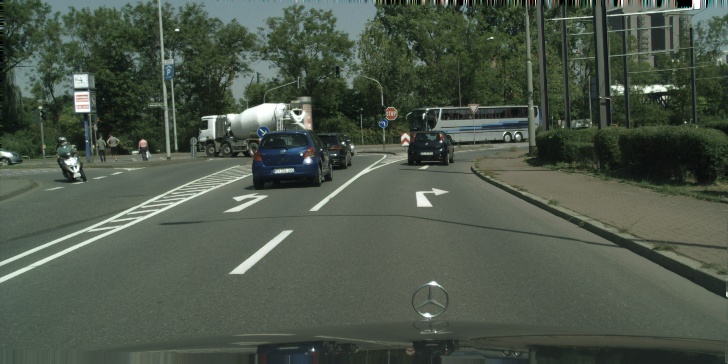} &
    \includegraphics[width=0.166\textwidth]{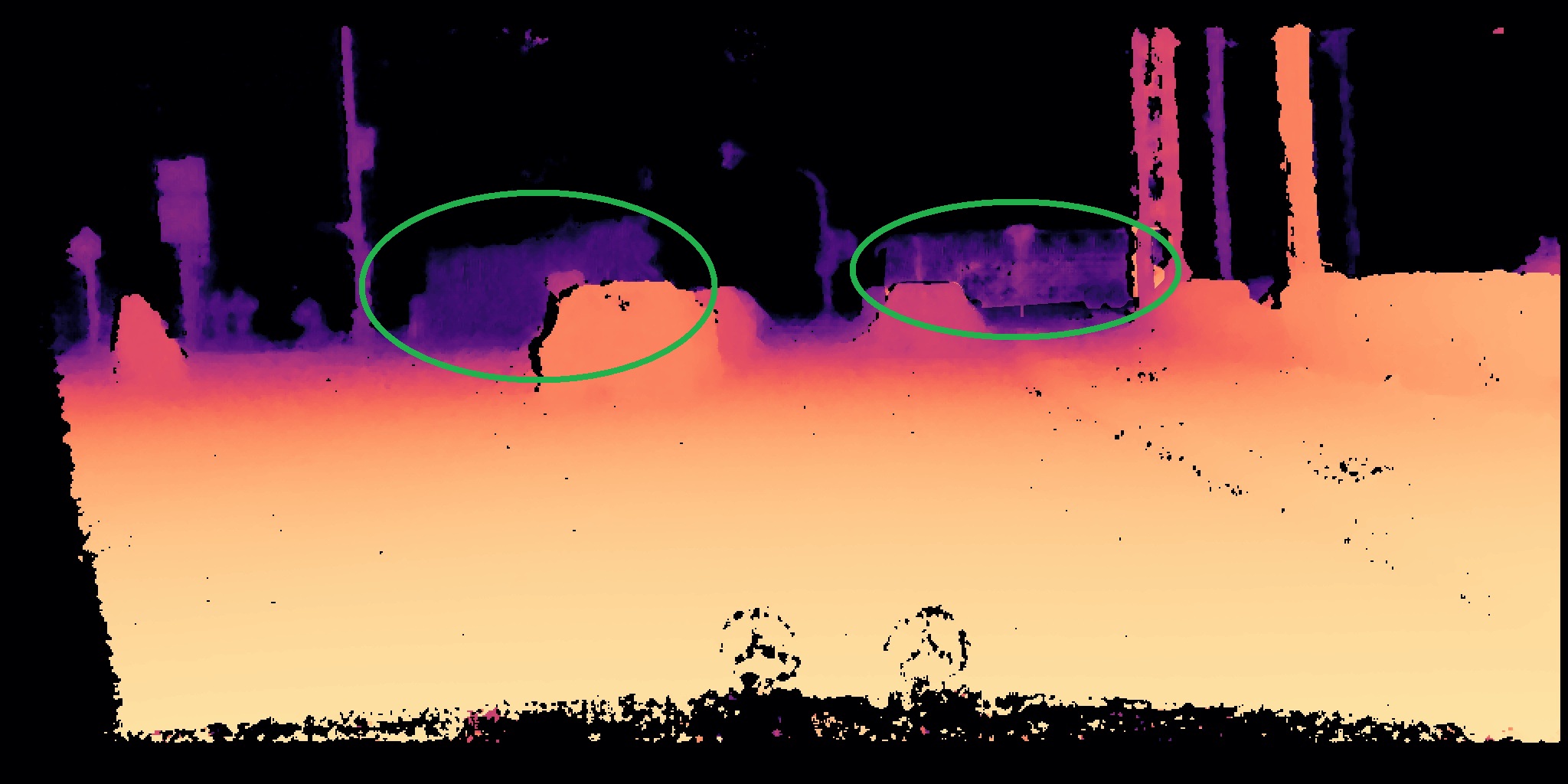} &
    \includegraphics[width=0.166\textwidth]{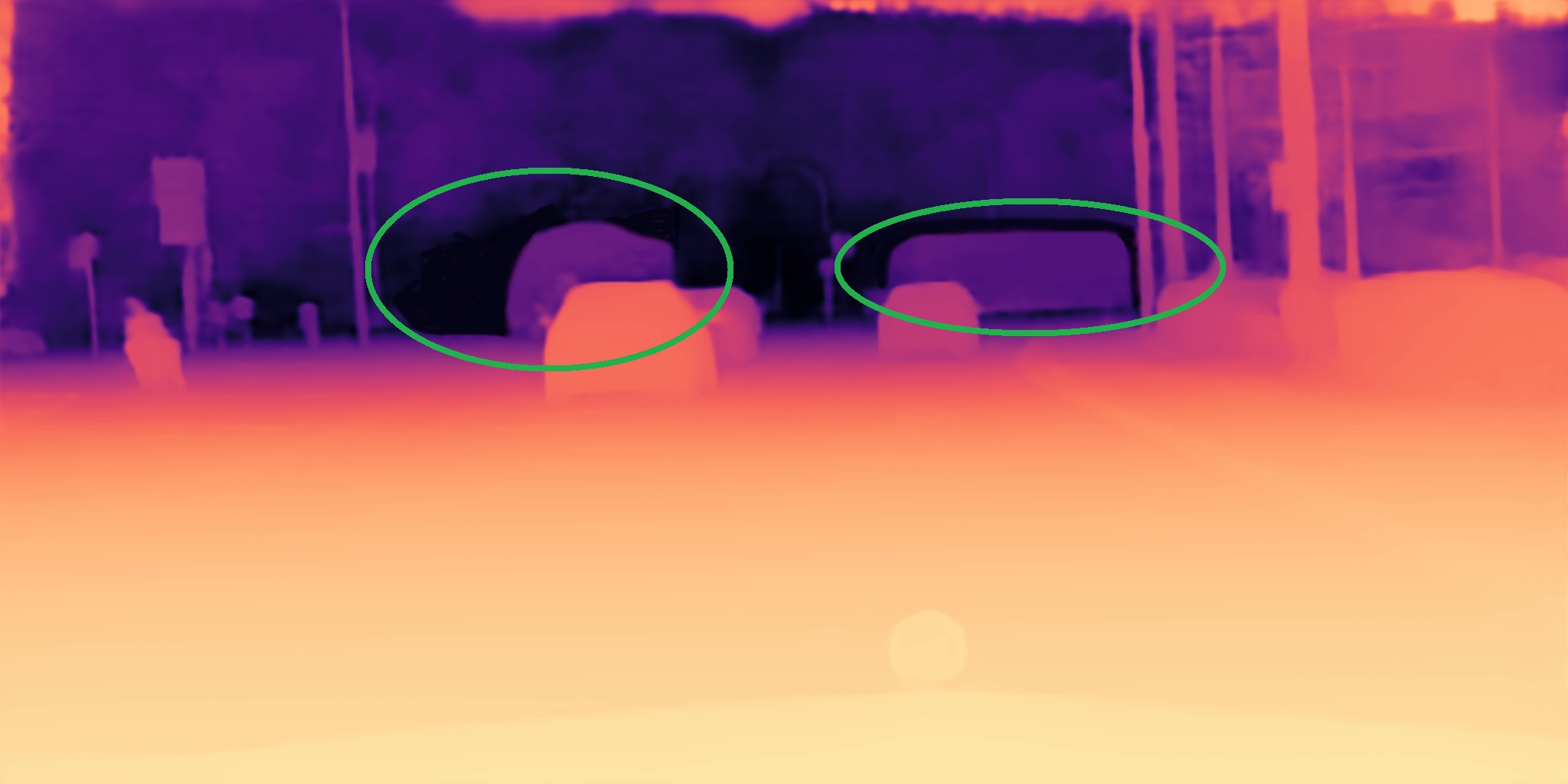} &
    \includegraphics[width=0.166\textwidth]{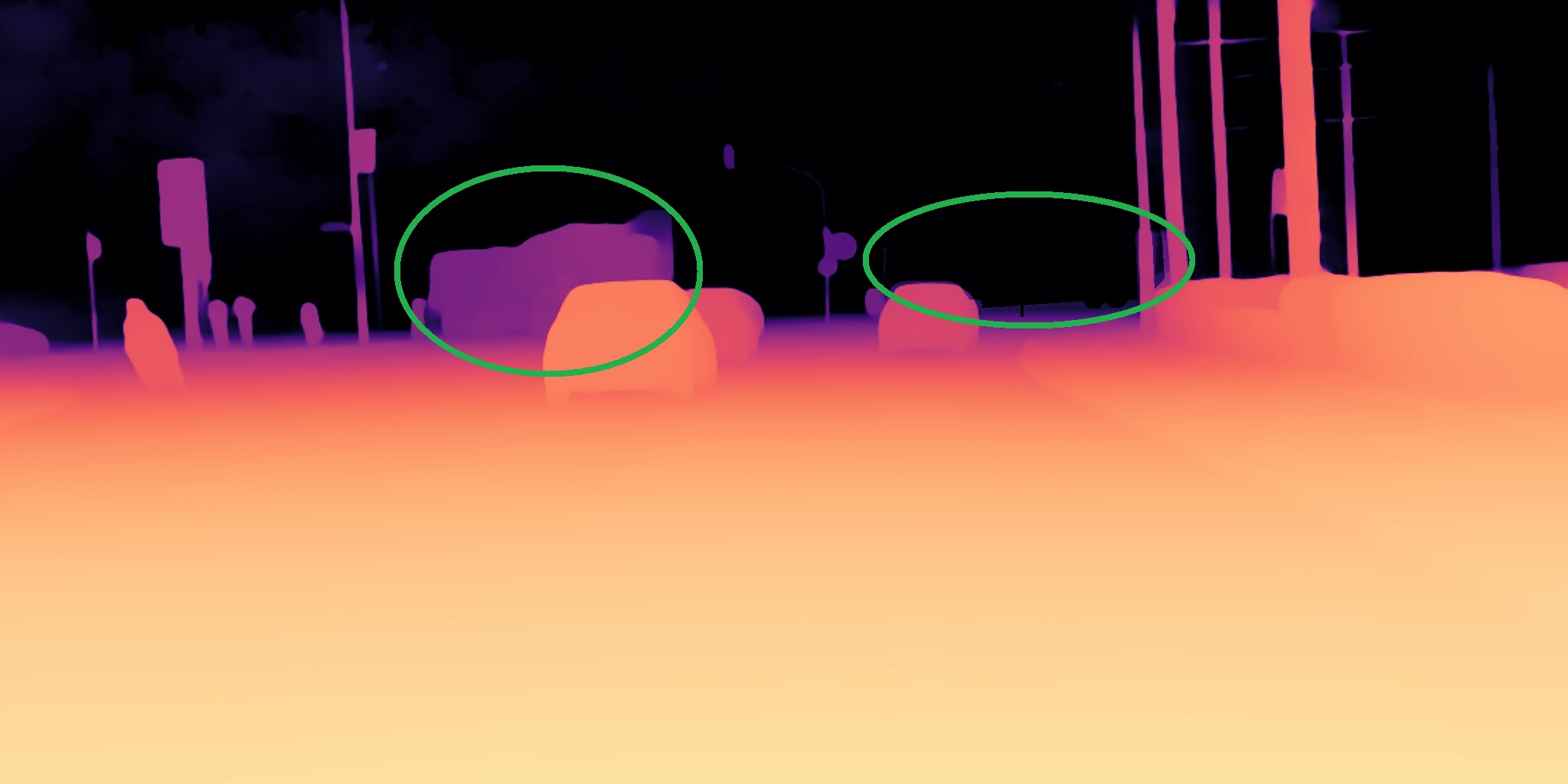} &
    \includegraphics[width=0.166\textwidth]{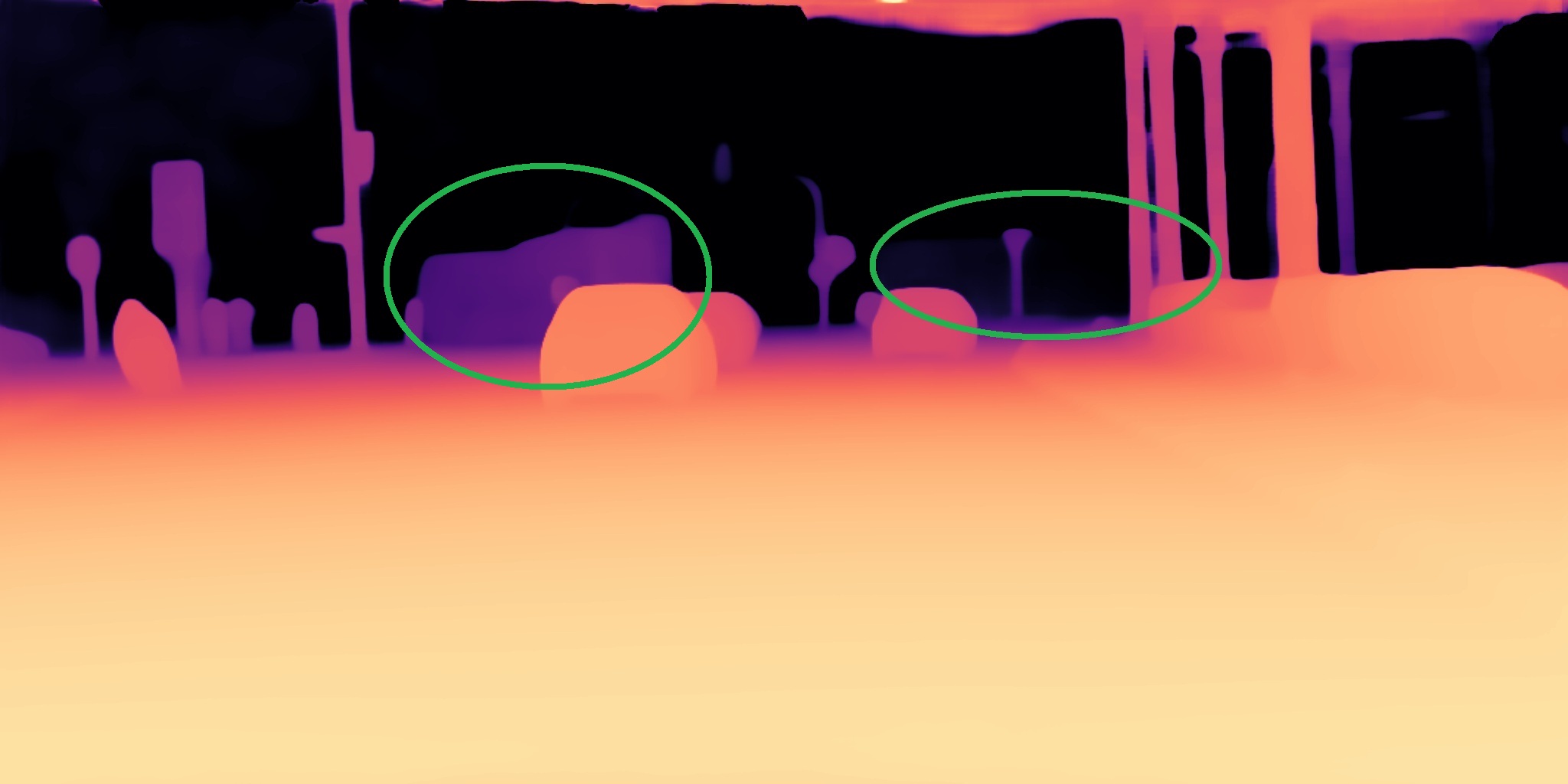} &
    \includegraphics[width=0.166\textwidth]{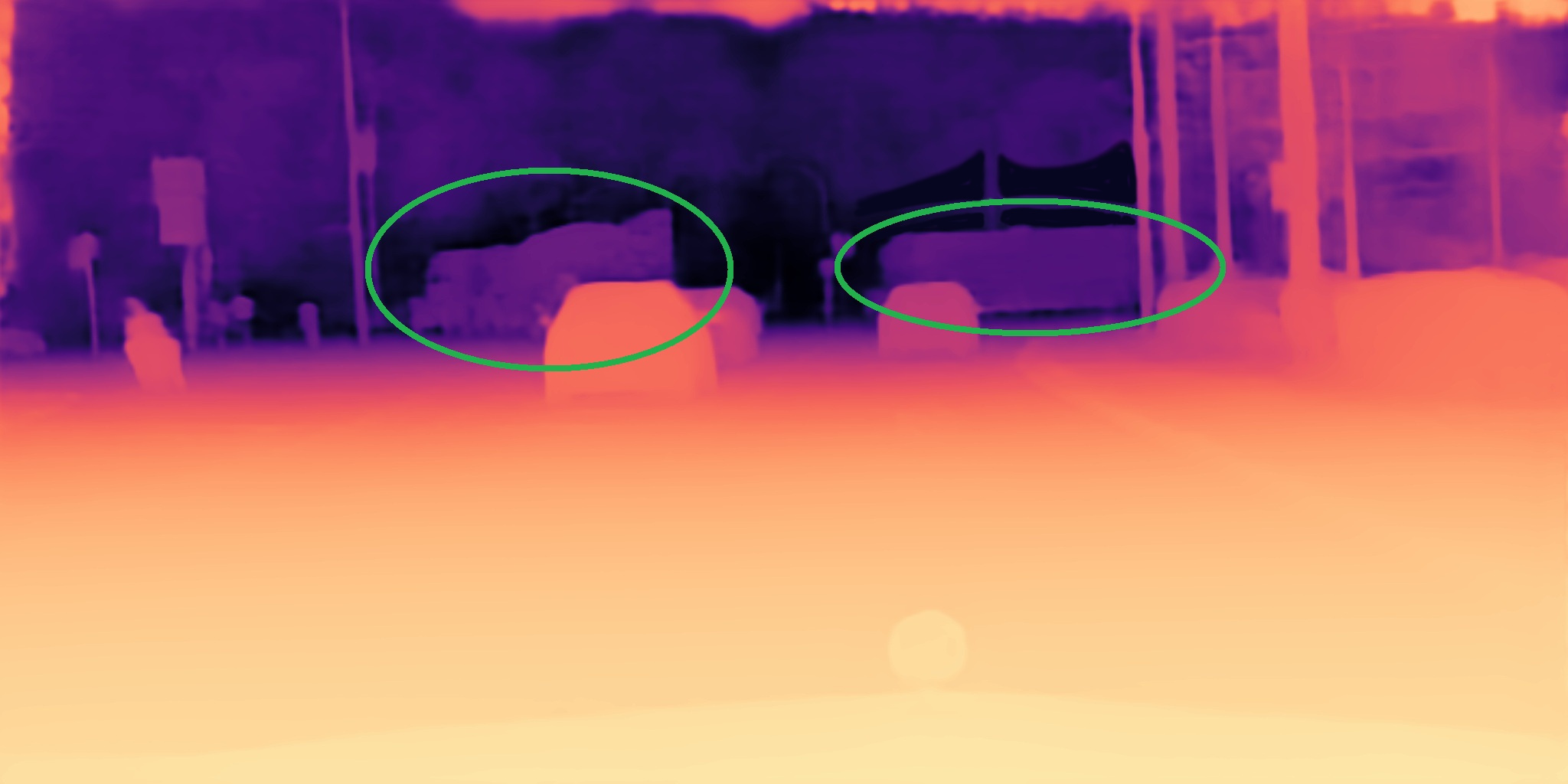} \\ [0.5ex]  %
    Input & Ground truth & DepthAnything v2 & UniDepth v2 & Metric3D v2& RAD
  \end{tabular}
  \vspace{-0.2cm}
  \caption{\textbf{Qualitative results} for NYU Depth v2 (top two rows), KITTI (middle two rows) and Cityscapes (bottom two rows). We compare our method (RAD) to baselines DepthAnything v2 \cite{DepthAnything_V2}, UniDepth v2 \cite{UniDepth_V2} and Metric3D v2 \cite{metric3d_v2}. Best viewed zoomed in.}
  \vspace{-0.1cm}
  \label{fig:qualitative results}
\end{figure*}

\section{Experiments}
\label{sec:results}

We demonstrate the effectiveness of RAD through quantitative and qualitative evaluations. We stress that in all experiments, our retrieval context dataset $D_{pool}$ is from the training set, making comparison to baselines fair. 
We then conduct experiments to visualize and analyze the behavior of RAD’s core components. Finally, we present a series of ablation studies.
Limitations are in the supplementary. 

\subsection{Experimental Setup}

\noindent \textbf{Datasets.} \quad 
We evaluate our method on three widely used depth estimation datasets: NYU Depth v2 \cite{dataset-nyud_v2}, KITTI \cite{dataset-kitti} (Eigen split), and Cityscapes \cite{dataset-cityscapes}. 
Since our focus is on underrepresented classes, we use semantic segmentation annotations to determine the location of underrepresented classes (detailed below) in test images, and calculate metrics (detailed below) on pixels of such classes in an isolated manner. Both NYU and Cityscapes provide ground-truth semantic segmentation annotations (for Cityscapes, we use the fine annotations from the validation split). KITTI, however, lacks semantic labels (a limitation that motivates the inclusion of Cityscapes as an outdoor dataset with manual annotated segmentation). To compensate for the absence of semantic annotations in KITTI, we generate segmentations using Detectron2 \cite{detectron2}, employing the R101-FPN checkpoint trained on COCO panoptic segmentation data. 

NYU includes 894 classes, and KITTI (with Detectron2 segmentation) has 133. We define underrepresented classes as those appearing in fewer than 10\% of training images but occurring more than five times, enabling analysis of the long tail while preserving natural distributions. For NYU, there are 231 underrepresented classes, while for KITTI, we have 22.
Interestingly, we find that these long-tail cases occur in 93\% of test examples in NYU and 51\% in KITTI.

For Cityscapes, we designate the following as underrepresented classes: truck, bus, caravan, trailer, and train, based on their substantially lower instance counts, roughly an order of magnitude fewer than more common categories. The test sets include 611 samples for NYU, 334 for KITTI, and 133 for Cityscapes. Notably, NYU contains a large number of sparsely represented classes, making it particularly well-suited for evaluating our task.

\noindent \textbf{Metrics.} \quad 
We adopt standard depth estimation metrics \cite{eigen2014depth}, including threshold accuracy (\(\delta_n\)), relative absolute error (RelAbs), root mean squared error (RMS), RMS in logarithmic space (\(\mathrm{RMS_{log}}\)), and average \(\mathrm{Log}_{10}\) error (\(\mathrm{Log_{10}}\)). Using these metrics, we conduct two types of evaluations: (1) underrepresented classes evaluation, where metrics are computed exclusively on pixels belonging to underrepresented semantic classes in the test images; and (2) all classes evaluation, where metrics are computed across all pixels in the test images, corresponding to the standard test set.

\noindent \textbf{Baselines.} \quad 
We refer to our method as RAD-{X}, where {X} denotes the size of the ViT encoder. We compare it against two categories of baselines: fine-tuning and zero-shot approaches (other methods in Sec.~\ref{sec:related work} operate in different settings). In the fine-tuning category, we compare to BTS \cite{bts}, AdaBins \cite{adabins}, NewCRF \cite{newcrf}, IEBins \cite{iebins}, iDisc \cite{idisc}, and DepthAnything v2 \cite{DepthAnything_V2}; in the zero-shot category, we compare to ZoeDepth \cite{ZoeDepth}, DepthPro \cite{DepthPro}, Metric3D v2 \cite{metric3d_v2}, and UniDepth v2 \cite{UniDepth_V2}, which represent the state-of-the-art for MMDE. See supplementary for further details. 

\begin{figure*}
    \centering
    \renewcommand{\arraystretch}{0.25}  %
    \begin{tabular}{@{}c@{}c@{}c@{}c@{}c@{}c@{}c@{}c@{}}
         \includegraphics[width=0.125\textwidth]{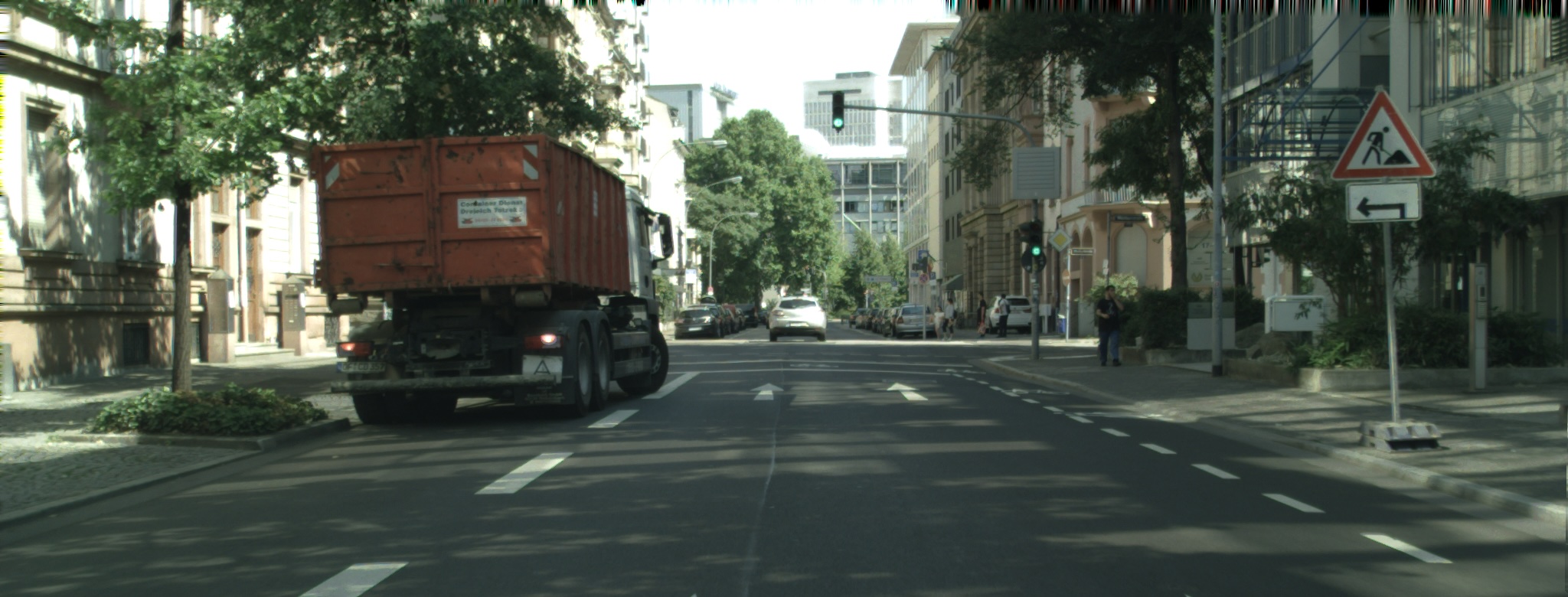} &
         \includegraphics[width=0.125\textwidth]{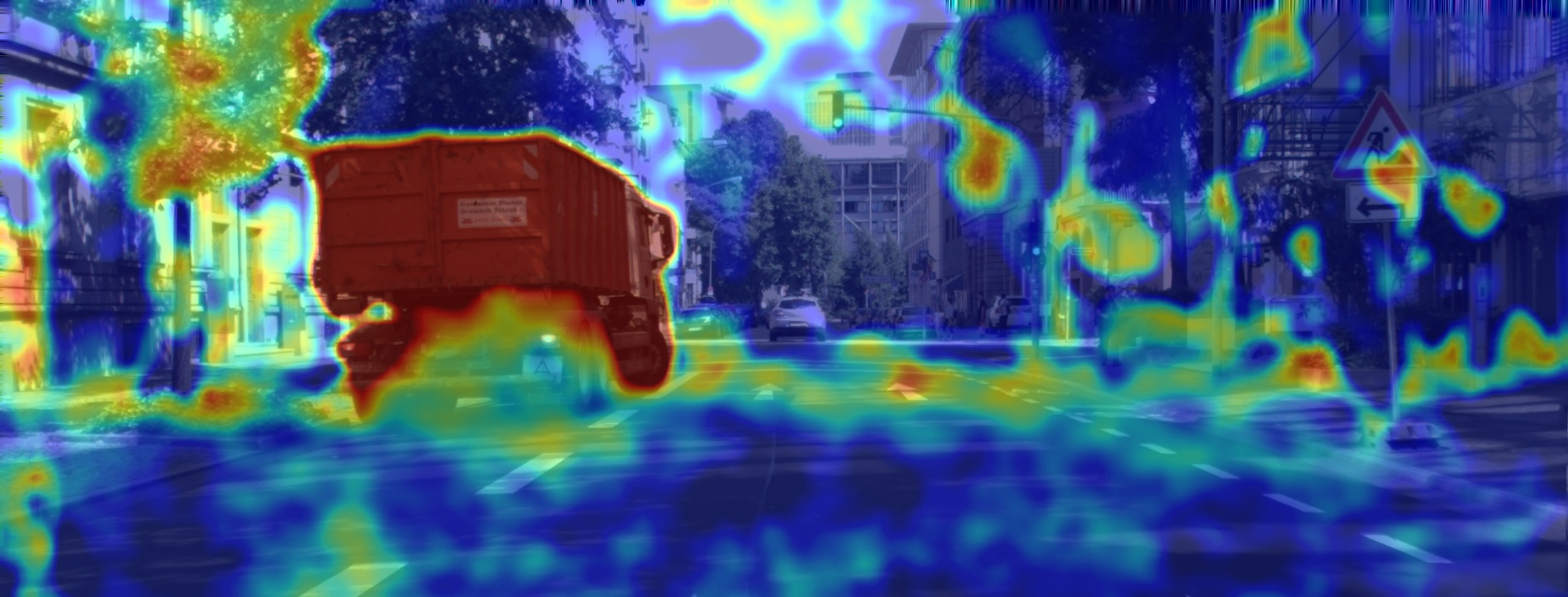} &
         \includegraphics[width=0.125\textwidth]{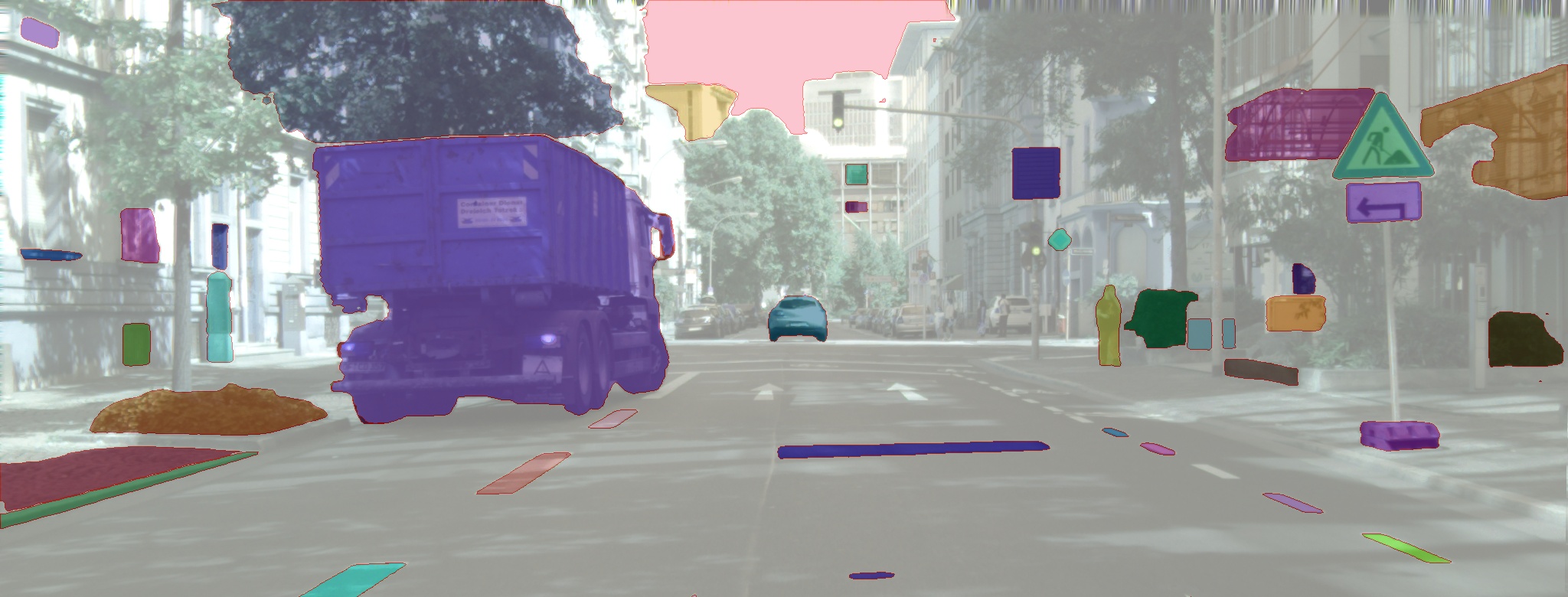} &
         \includegraphics[width=0.125\textwidth]{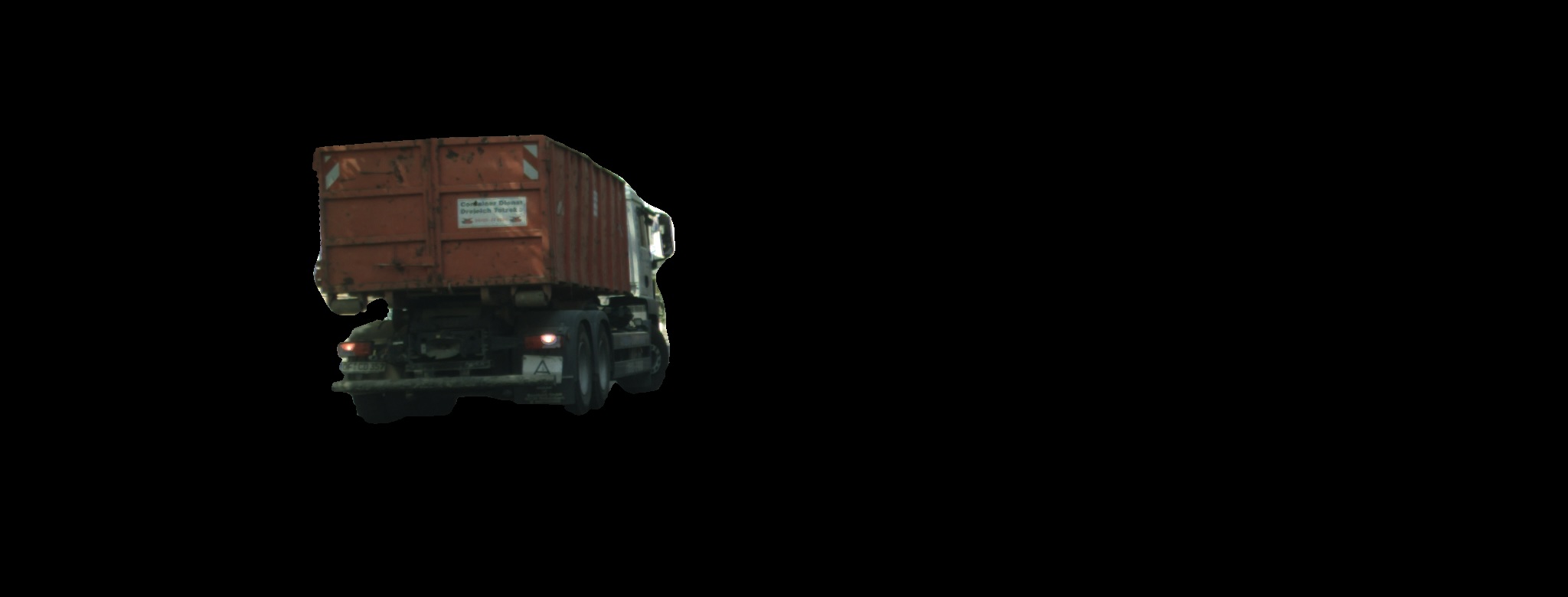} &
         \includegraphics[width=0.125\textwidth]{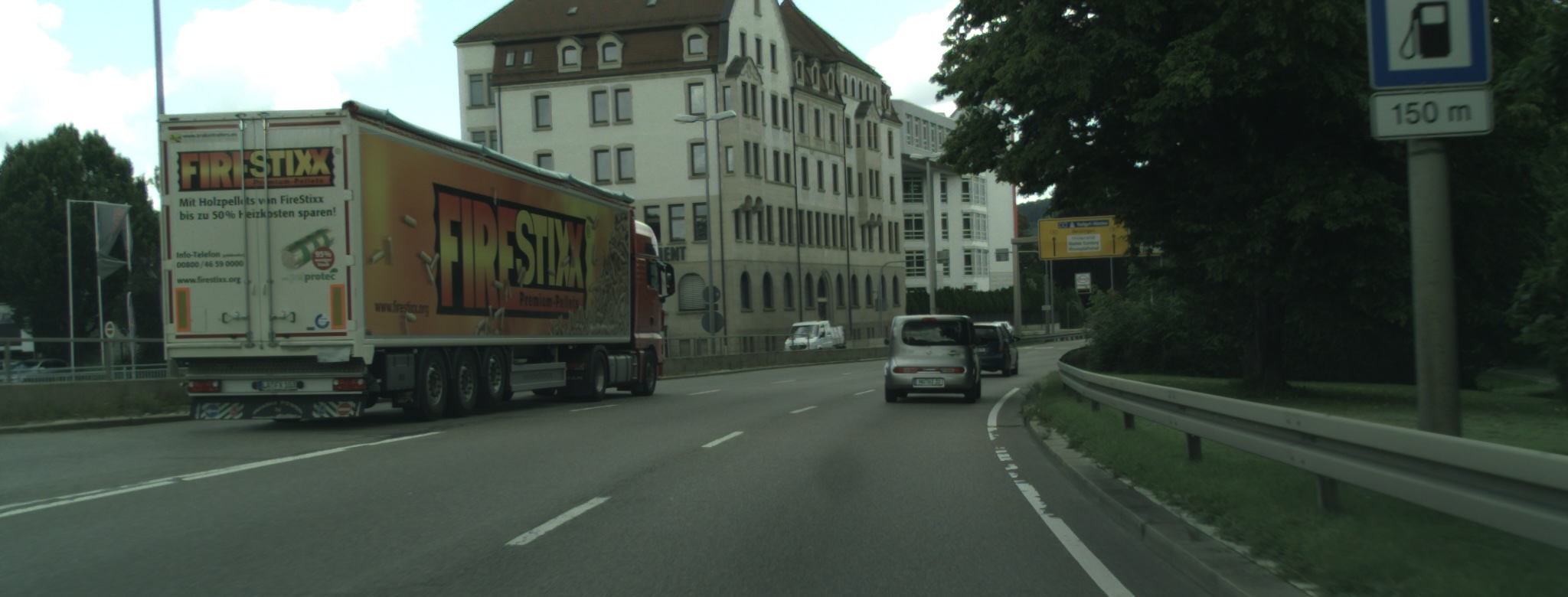} &
         \includegraphics[width=0.125\textwidth]{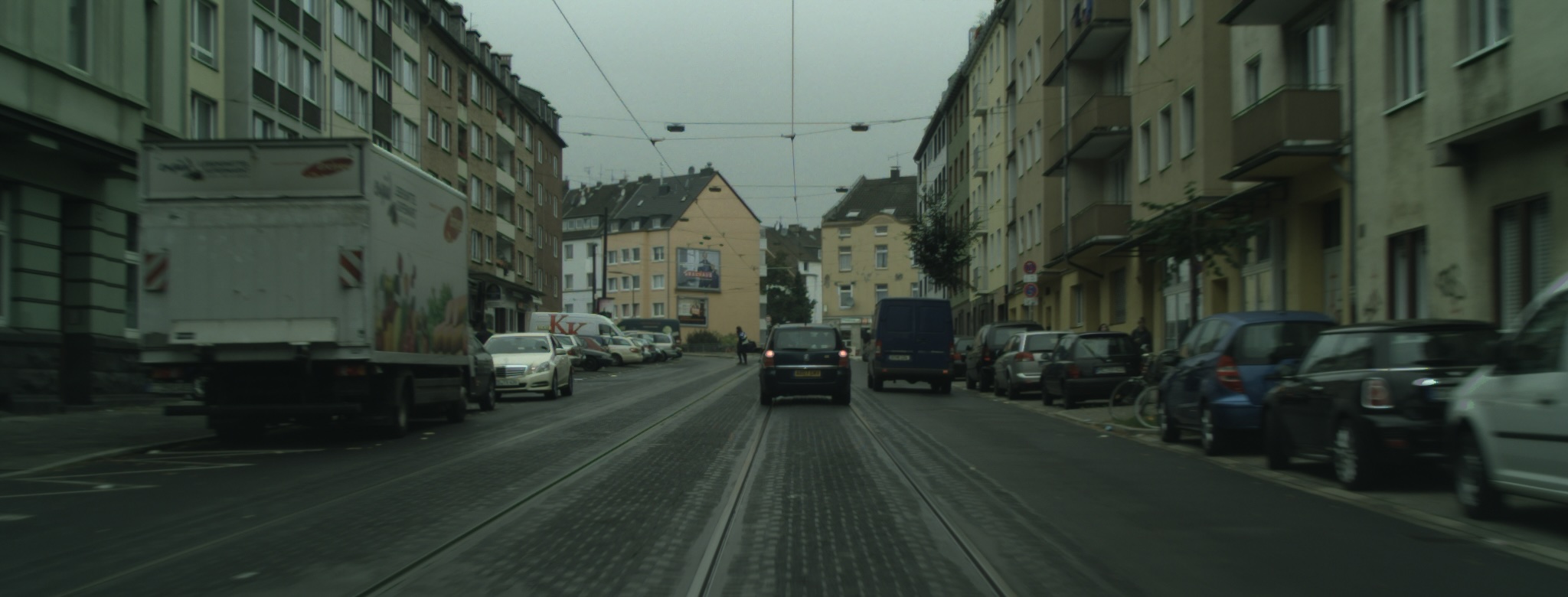} &
         \includegraphics[width=0.125\textwidth]{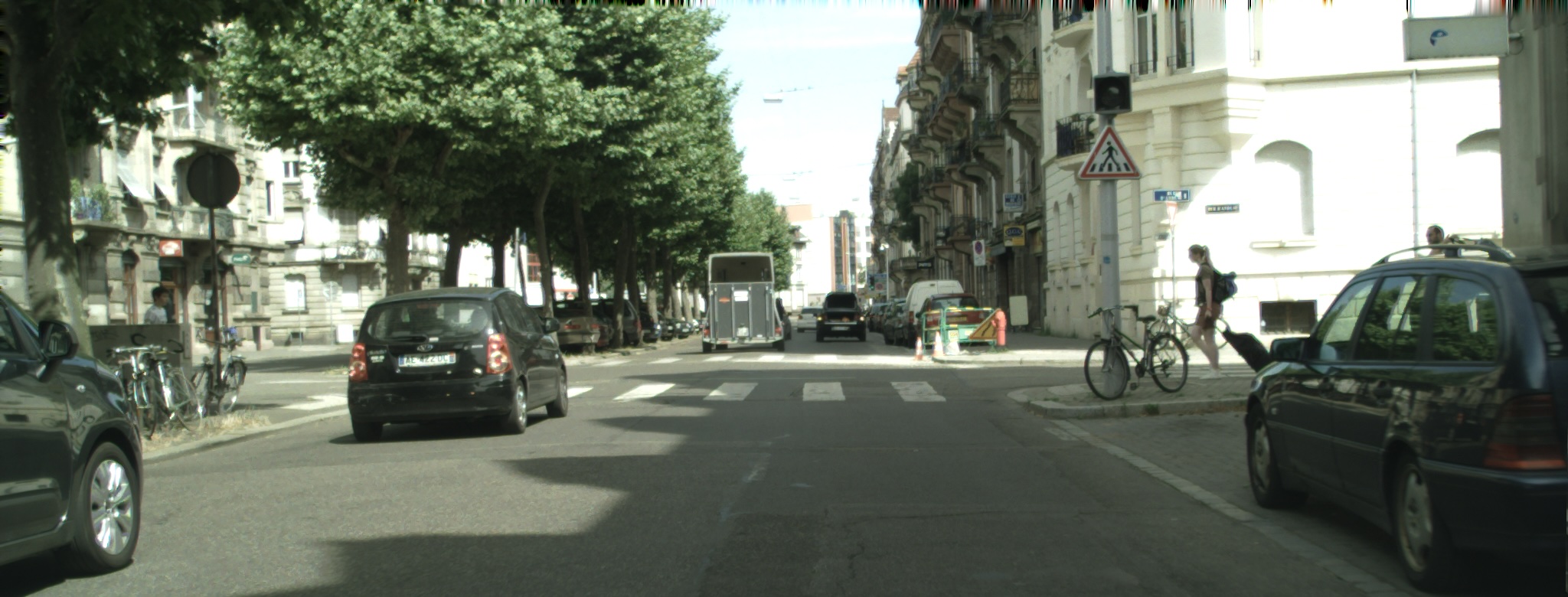} &
         \includegraphics[width=0.125\textwidth]{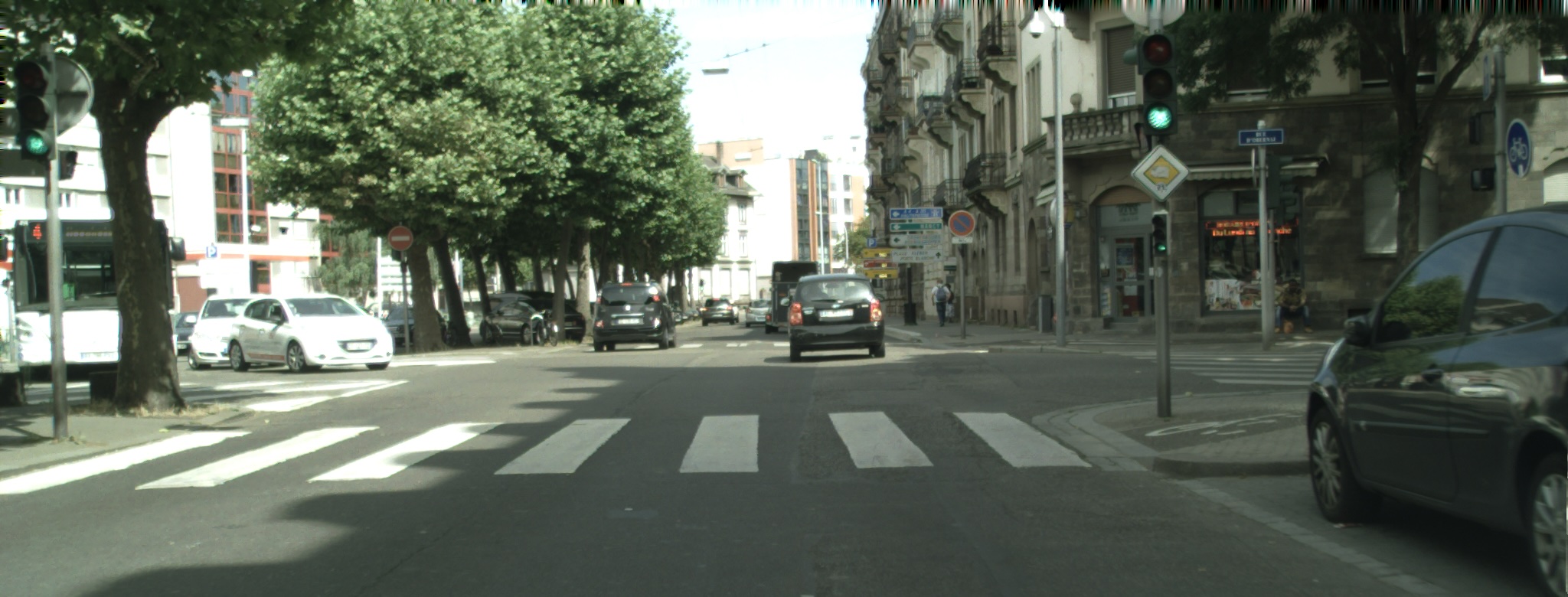} \\
         \includegraphics[width=0.125\textwidth]{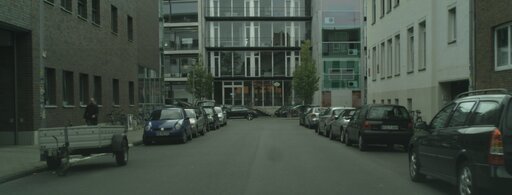} &
         \includegraphics[width=0.125\textwidth]{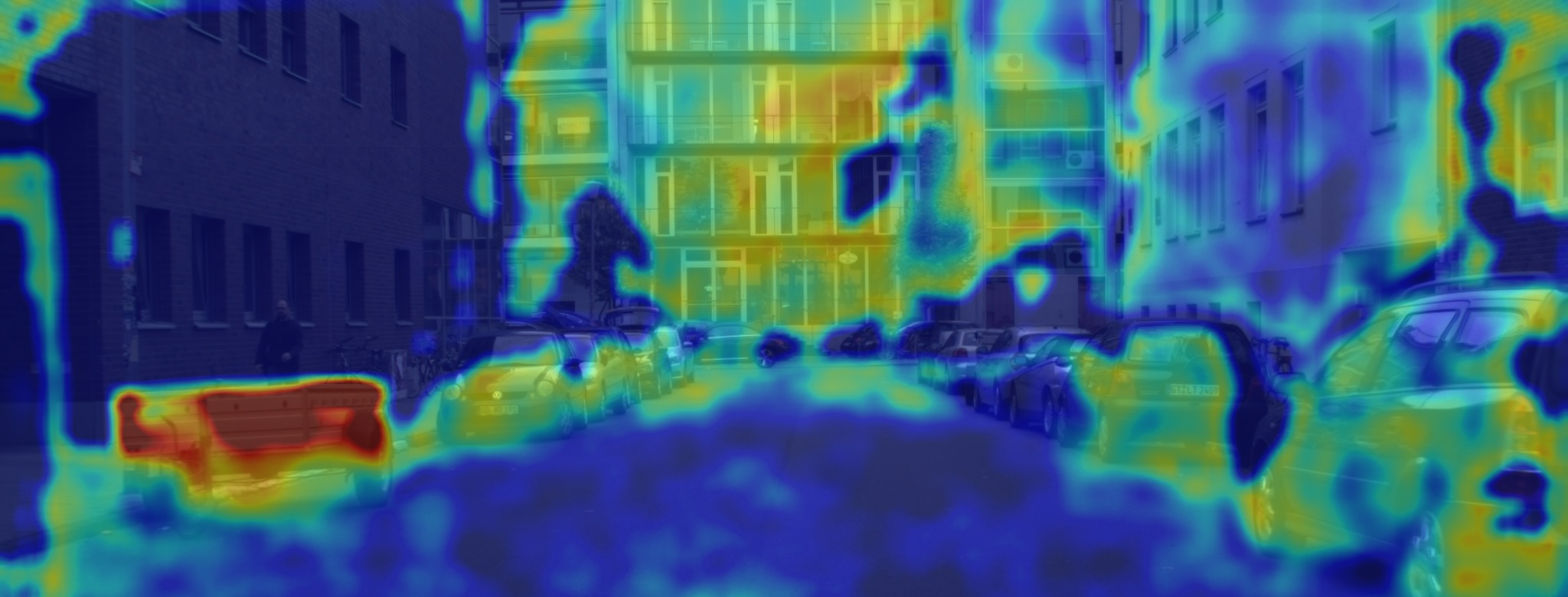} &
         \includegraphics[width=0.125\textwidth]{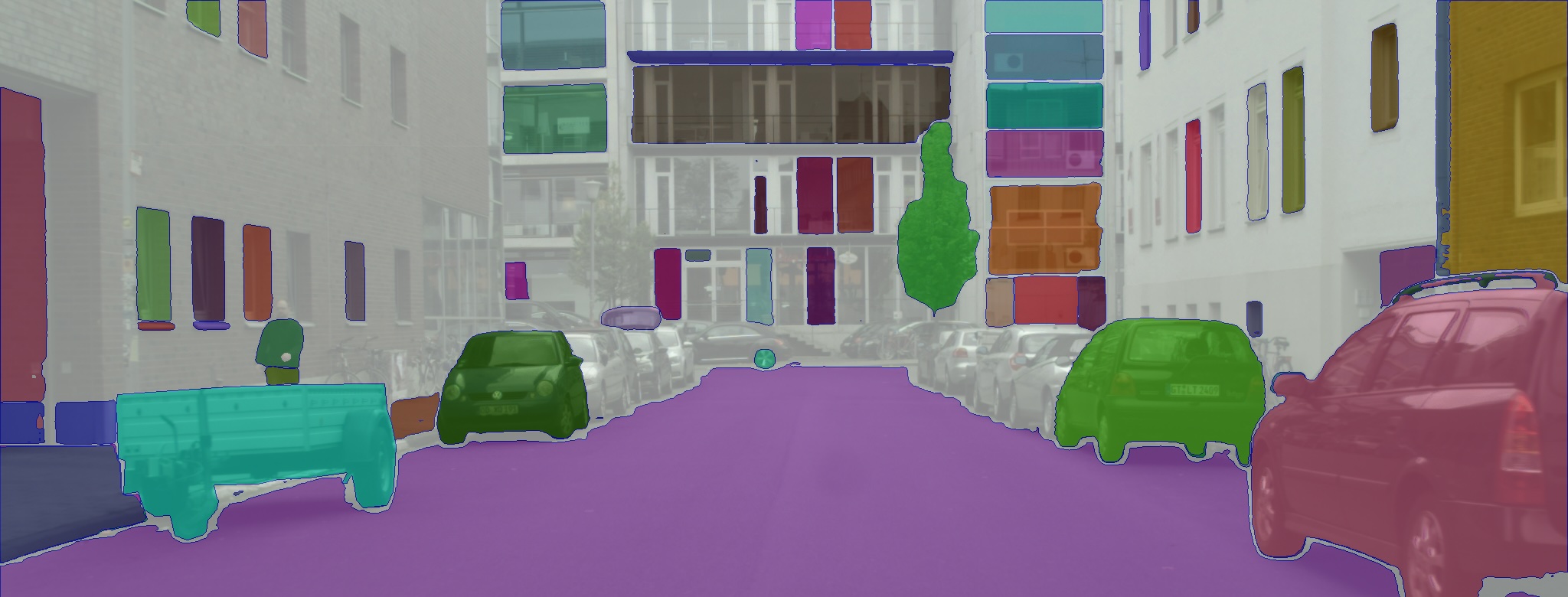} &
         \includegraphics[width=0.125\textwidth]{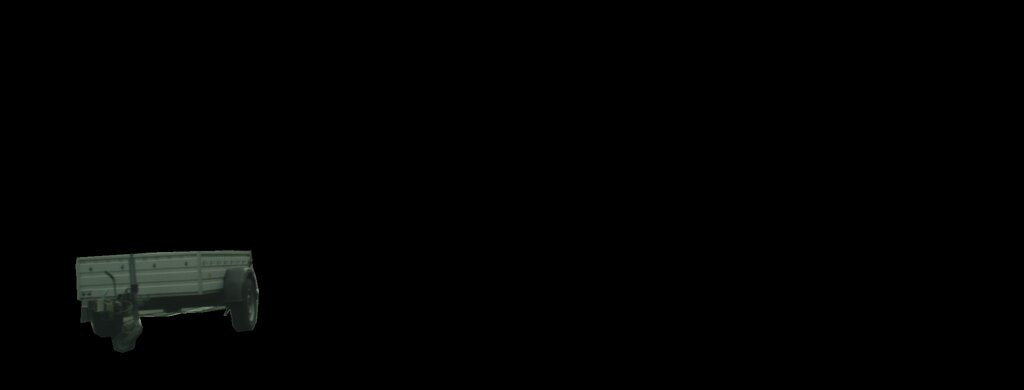} &
         \includegraphics[width=0.125\textwidth]{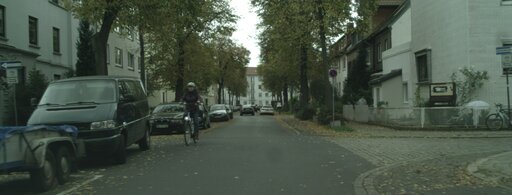} &
         \includegraphics[width=0.125\textwidth]{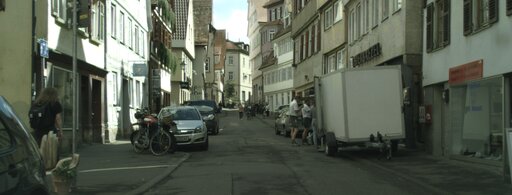} &
         \includegraphics[width=0.125\textwidth]{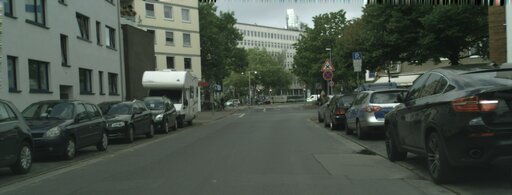} &
         \includegraphics[width=0.125\textwidth]{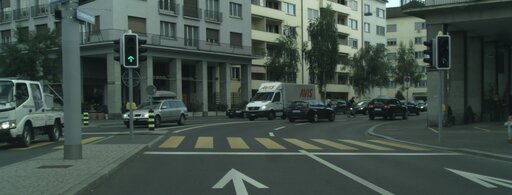} \\
         \includegraphics[width=0.125\textwidth]{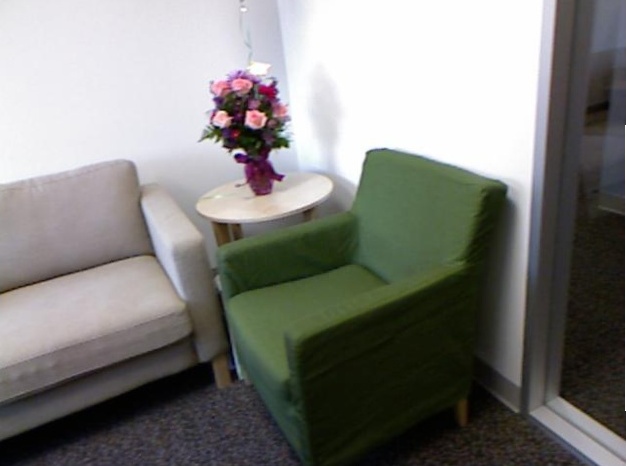} &
         \includegraphics[width=0.125\textwidth]{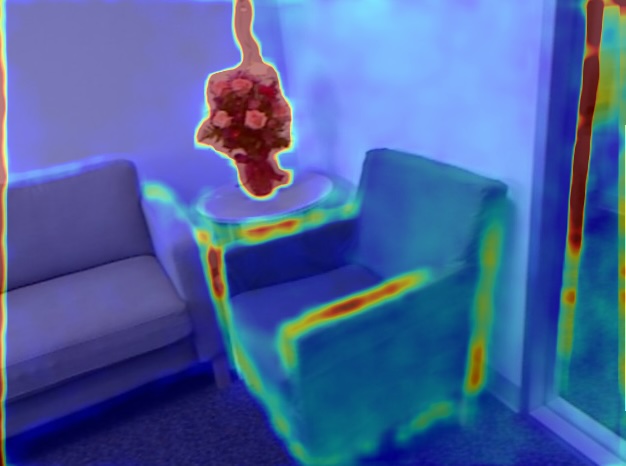} &
         \includegraphics[width=0.125\textwidth]{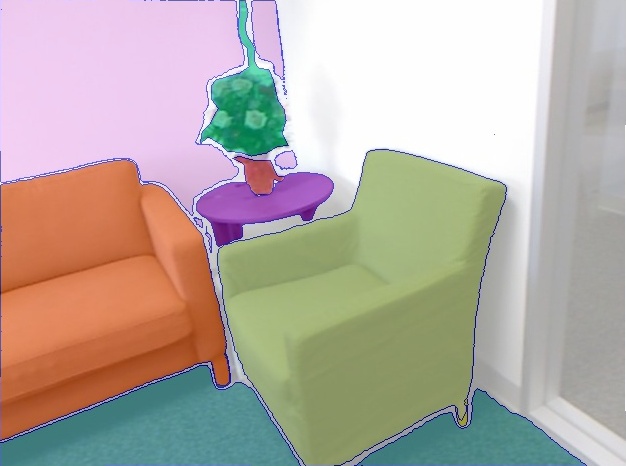} &
         \includegraphics[width=0.125\textwidth]{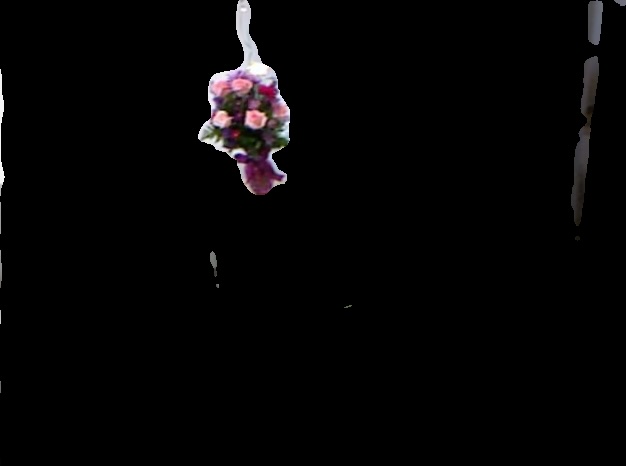} &
         \includegraphics[width=0.125\textwidth]{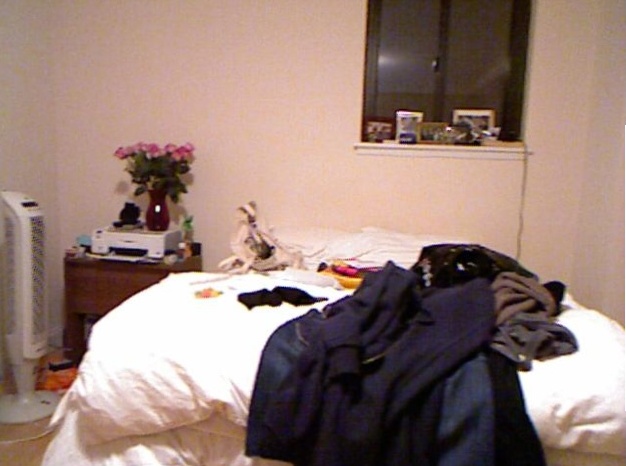} &
         \includegraphics[width=0.125\textwidth]{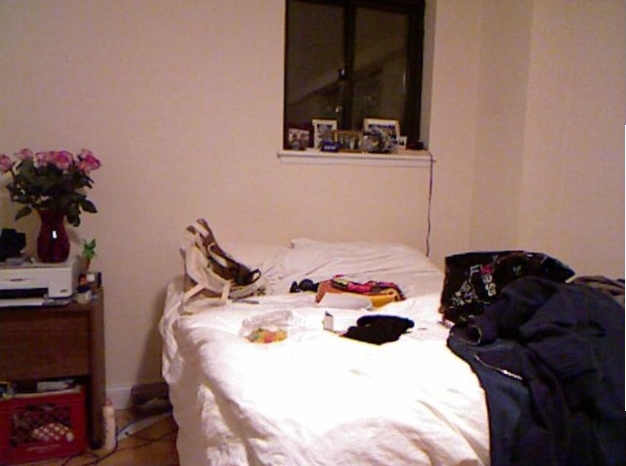} &
         \includegraphics[width=0.125\textwidth]{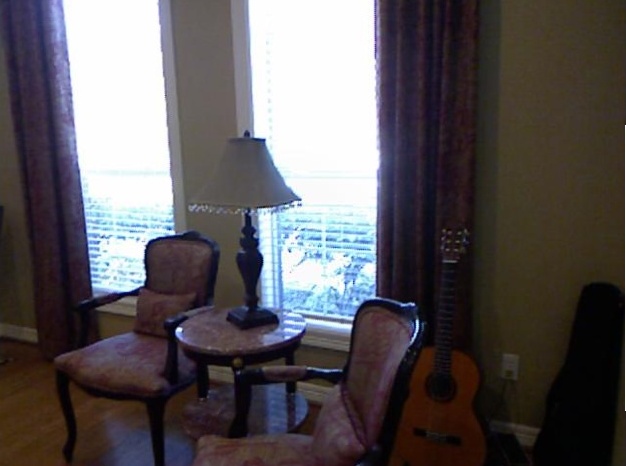} &
         \includegraphics[width=0.125\textwidth]{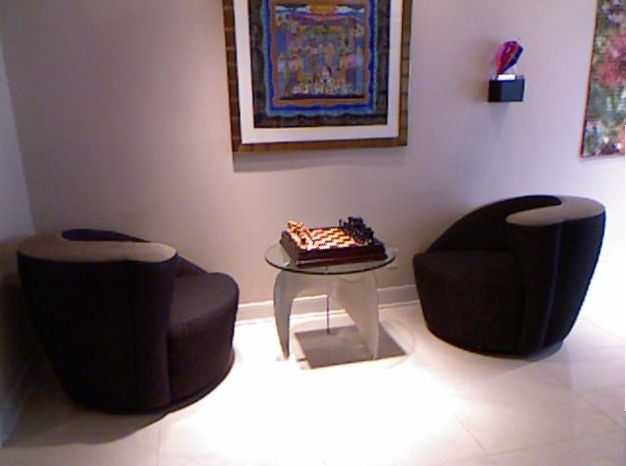} \\
         {\footnotesize Query} &
         {\footnotesize Uncertainty} &
         {\footnotesize Segmentation} &
         \parbox{0.125\textwidth}{\centering \footnotesize  Masked \\ query} &
         \parbox{0.125\textwidth}{\centering \footnotesize Uncertainty-aware \\ retrieved 1 (ours)} &
         \parbox{0.125\textwidth}{\centering \footnotesize Uncertainty-aware \\ retrieved 2 (ours)} &
         \parbox{0.125\textwidth}{\centering \footnotesize DINO retrieved 1 \\ (baseline)} &
         \parbox{0.125\textwidth}{\centering \footnotesize DINO retrieved 2 \\ (baseline)}
    \end{tabular}
    \vspace{-0.2cm}
    \caption{\textbf{Uncertainty-aware retrieval.} Comparison of image retrieval using our \textit{uncertainty-aware} approach to the baseline \textit{DINO}-based retrieval. Our approach identifies segments of high uncertainty in the query image and retrieves examples containing similar objects, rather than images with similar general layout. Best viewed zoomed in.}
    \label{fig:retrieval examples}
    \vspace{-0.4cm}
\end{figure*}

\begin{figure*}[htbp]
    \centering
    \begin{subfigure}[b]{0.33\textwidth}
        \includegraphics[width=\textwidth]{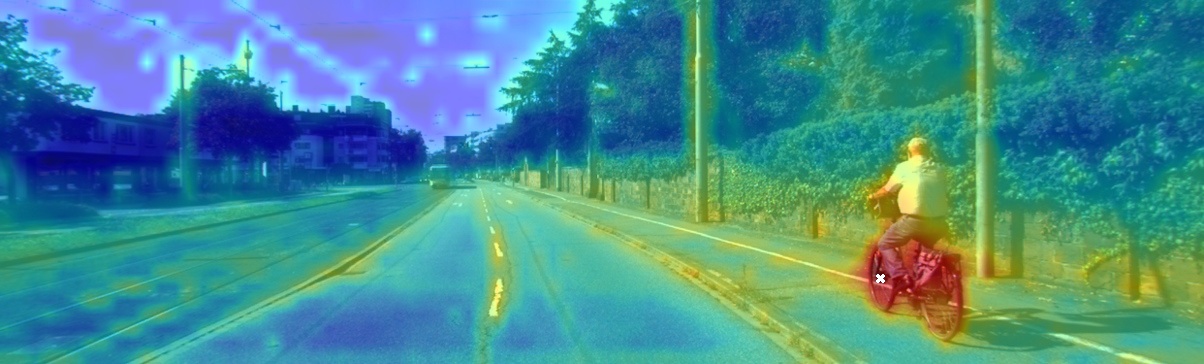}
    \end{subfigure}
    \begin{subfigure}[b]{0.66\textwidth}
        \includegraphics[width=\textwidth]{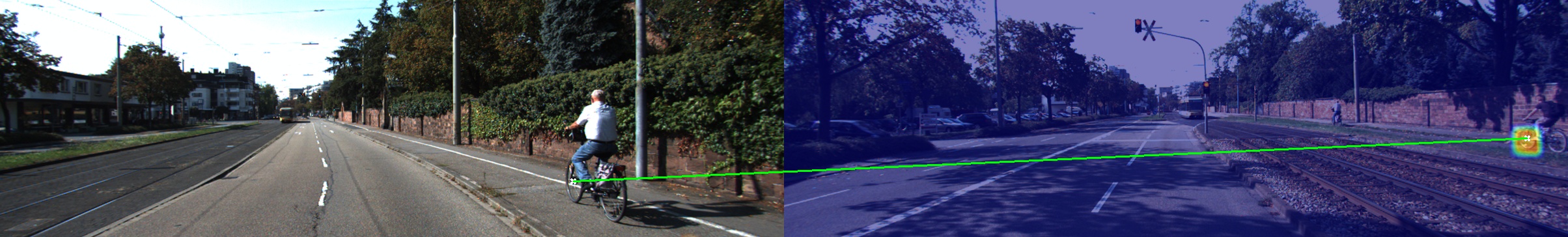}
    \end{subfigure} \\
    \begin{subfigure}[b]{0.33\textwidth}
        \includegraphics[width=\textwidth]{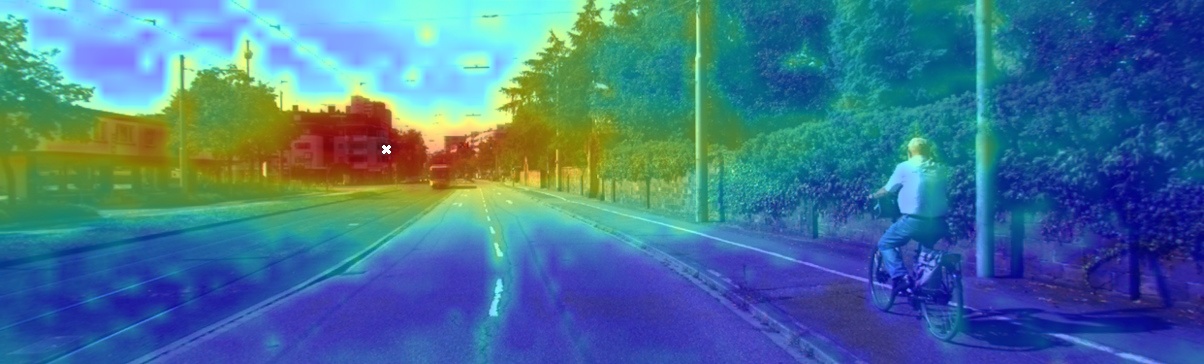}
        \caption{}
    \end{subfigure}
    \begin{subfigure}[b]{0.66\textwidth}
        \includegraphics[width=\textwidth]{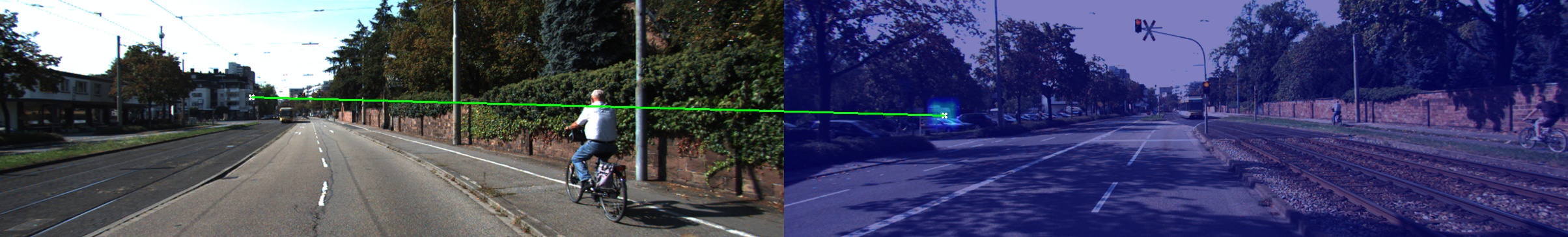}
        \caption{}
    \end{subfigure}
    \vspace{-0.3cm}
    \caption{\textbf{Visualization of matched cross-attention}. For a selected patch in the input image, indicated by a white marker, and its corresponding matched point in the context image, we separately visualize the attention directed toward the input (a) and the context (b). 
    When the match is correct, strong cross-attention emerges within the local neighborhood of the matched point in the context image. In contrast, incorrect matches yield weak cross-attention responses. Colormaps are kept consistent across images. }
    \vspace{-0.4cm}
    \label{fig:attention}
\end{figure*}

\subsection{Quantitative Evaluation}
We compare RAD against other state-of-the-art methods, reporting results based on publicly available code and pretrained models, or as documented in the literature.

\noindent \textbf{Underrepresented classes evaluation.} \quad
Tab.~\ref{tab:underrepresented classes - all datasets} presents quantitative comparison of depth estimation methods evaluated on underrepresented data from NYU, KITTI, and Cityscapes. For methods with multiple network sizes, we report results for the most comparable to a large ViT (the best variant). Overall, our RAD-Large consistently achieves the strongest performance across benchmarks.

On NYU, RAD-Large achieves the highest overall performance, with relative improvements of 29.2\% in AbsRel and 11.9\% in RMS compared to the best-performing non-RAD method, DepthAnything v2, the foundation of our approach. Additionally, it achieves a 5.1\% gain in $\delta_1$ over UniDepthV2, which ranks second in that metric, excluding other RAD models.
On KITTI, RAD-Large again leads, demonstrating relative improvements of 13.3\% in RelAbs, 39.5\% in $\mathrm{RMS_{log}}$, and 4.7\% in $\delta_1$ over the best non-RAD method, DepthAnything v2. Similarly, on Cityscapes, RAD-Large delivers the strongest results, with relative improvements of 7.2\% in AbsRel and 28.0\% in RMS compared to Metric3Dv2, the best non-RAD baseline.

\noindent \textbf{All classes evaluation.} \quad 
Tab.~\ref{tab:all classes - all datasets} presents a quantitative comparison to baselines evaluated across all semantic classes on NYU, KITTI, and Cityscapes, corresponding to the standard MMDE test protocol. On NYU and Cityscapes, RAD-Large performs on par with state-of-the-art methods such as UniDepthV2 and Metric3Dv2, while closely ranking second on KITTI. Notably, across all datasets, RAD-Large slightly outperforms DepthAnything v2, the backbone of our approach. 

\subsection{Qualitative Evaluation}
Fig.~\ref{fig:qualitative results} showcases qualitative results across the three datasets. RAD demonstrates a clear advantage in handling underrepresented objects. 
Notably, DepthAnything v2 and RAD exhibit similar performance in in-domain regions of the image, a natural outcome given their shared backbone. 
RAD detects underrepresented elements with high uncertainty, delivering improved depth estimations in the lamp in the first image, the stroller in the third image, and the side of the train in the fifth image.

\subsection{Component Visualization}
Fig.~\ref{fig:retrieval examples} analyzes the uncertainty-aware image retrieval mechanism (Sec.~\ref{sec:data preparation}). 
Uncertainty effectively localizes underrepresented classes. Combined with segmentation, it enables specific retrieval of context images containing visually similar object instances.
In contrast, the baseline method, based on Nearest-Neighbour retrieval using global DINO v2 \cite{dinov2} descriptors, tends to retrieve images with similar overall layouts and semantics but fails to capture the targeted underrepresented objects.

Fig.~\ref{fig:attention} illustrates matched cross-attention behavior (Sec.~\ref{sec:depth estimation network}) by showing attention maps for input (a) and context (b) patches (all attention maps are computed by averaging across attention heads and ViT blocks). White markers denote reference patches; green lines show matches. The top row reflects a correct match, the bottom an incorrect one. In the input image, attention concentrates on semantically relevant regions for the selected patch (e.g., the bicycle). For a correct match, strong activation appears in the neighborhood of the matching location in the context image (e.g., the bicycle's wheel), while incorrect matches result in weak attention responses, demonstrating the model’s ability to suppress unreliable correspondences, a direct consequence of our matching attention design.  Additional examples are provided in the supplementary.

The supplementary material presents visualizations of point matching instances, showing that the method effectively identifies correspondences between visually similar underrepresented objects across diverse scenes.

\begin{table}[tbh]
    \centering
        \caption{\textbf{Ablations} of RAD-Large, evaluated on Cityscapes. 
    }
    \small
    \begin{tabular}{lcc}
         \hline
         Ablation & $\delta_1$ & AbsRel \\
         \hline
         RAD-Large & 93.5 & 0.090 \\
         \ding{56} 3D data augmentation in train & 92.8 & 0.096 \\
         \ding{56} Retrieved data in train & 91.7 & 0.099 \\
         \ding{56} Uncertainty-aware retrieval in train & 91.9 & 0.102 \\
         \ding{56} Uncertainty-aware retrieval in test & 90.1 & 0.107 \\
         \ding{56} Matched cross-attention & 90.3 & 0.106 \\
         \ding{56} Dual-stream (i.e., concat(in, retrieval)) & 89.4 & 0.109 \\
         \ding{56} Retrieval (i.e., DepthAnyV2 \cite{DepthAnything_V2}) & 89.2 & 0.110 \\
         \hline
    \end{tabular}
    \label{tab:ablations}
    \vspace{-0.2cm}
\end{table}

\begin{figure}[tbh]
    \centering
    \includegraphics[width=0.45\linewidth]{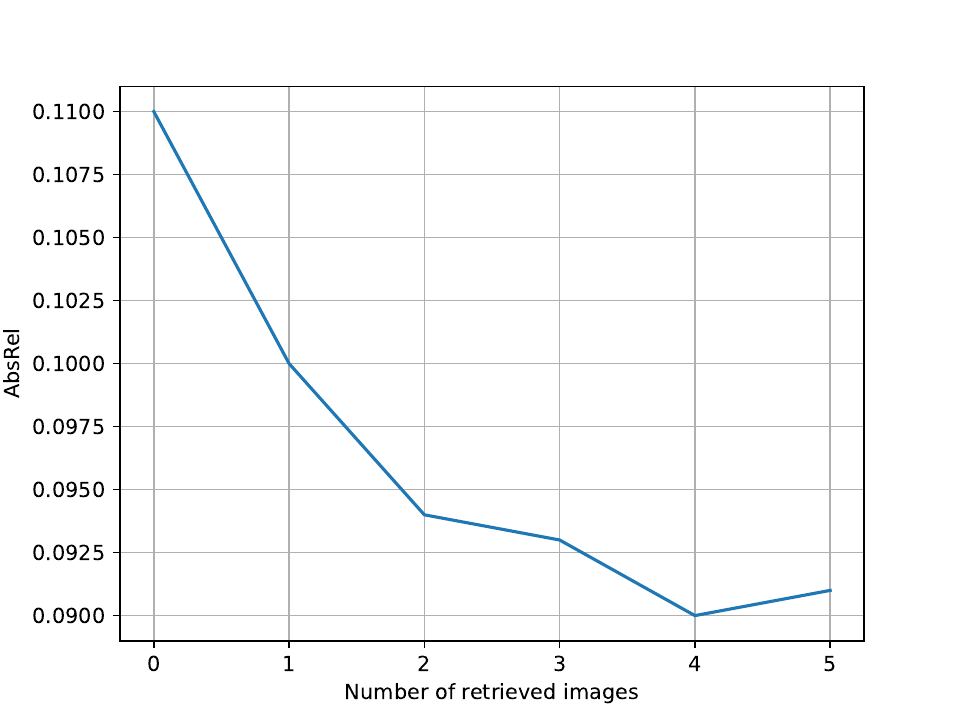}
    \includegraphics[width=0.45\linewidth]{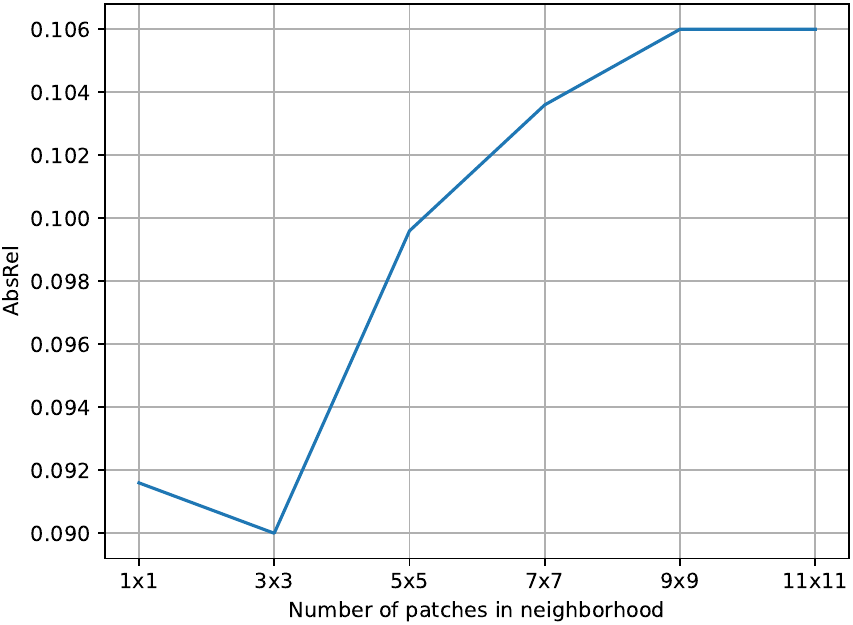}
    \vspace{-0.2cm}
    \caption{\textbf{Ablation on number of retrieved images and neighborhood size.} Absolute relative error as a function of the number of retrieved images (left) and neighborhood size in matched cross-attention (right) for RAD-Large, evaluated on Cityscapes. 
    }
    \label{fig:absrel vs num retrieved images and patch neighborhood size}
    \vspace{-0.4cm}
\end{figure}

\subsection{Ablation Study}
Our ablation study is conducted on the RAD-Large model, evaluated on the Cityscapes dataset.
Table~\ref{tab:ablations} confirms the necessity of all components. Removing \textit{3D augmentation} or \textit{retrieval} during training increases AbsRel by 6.7\% and 10.0\% respectively. The most critical component is the \textit{uncertainty-aware retrieval}; disabling it  (i.e. using global DINO KNN on the full image) at inference causes the largest drop (18.9\% increase in AbsRel), confirming that targeted context is superior to global retrieval. Furthermore, replacing our \textit{matched cross-attention} with standard full-context attention increases errors by 17.8\%, validating the need for localized information transfer. Finally, eliminating the dual-stream design and instead using a single stream on the concatenated input and retrievals yields the worst performance, degrading AbsRel by over 22\%.

We now examine a key parameter: the number of retrieved images. As shown in Fig.~\ref{fig:absrel vs num retrieved images and patch neighborhood size} (left), AbsRel steadily improves with more retrievals, reaching its lowest value at four. However, using five images introduces more irrelevant or misaligned content, slightly degrading performance.

Finally, we analyze the effect of neighborhood size in matched cross-attention. As shown in Fig.~\ref{fig:absrel vs num retrieved images and patch neighborhood size} (right), a $3\times 3$ neighborhood yields optimal performance. %

\section{Conclusion}
\label{sec:conclusion}
We presented RAD, a retrieval-augmented framework addressing the challenge of monocular depth estimation for underrepresented classes. By identifying regions of high uncertainty and retrieving semantically similar context, our method effectively supplements the network's internal priors with external proxy geometric guidance. Through a dual-stream architecture and matched cross-attention, we demonstrate that this targeted context can be reliably fused to correct depth in long-tail scenarios. Our approach achieves state-of-the-art performance on rare classes across NYUv2, KITTI, and Cityscapes, validating our approach. 

\clearpage
\setcounter{page}{1}
\maketitlesupplementary

\appendix


\begin{figure}
    \centering
    \includegraphics[width=\linewidth]{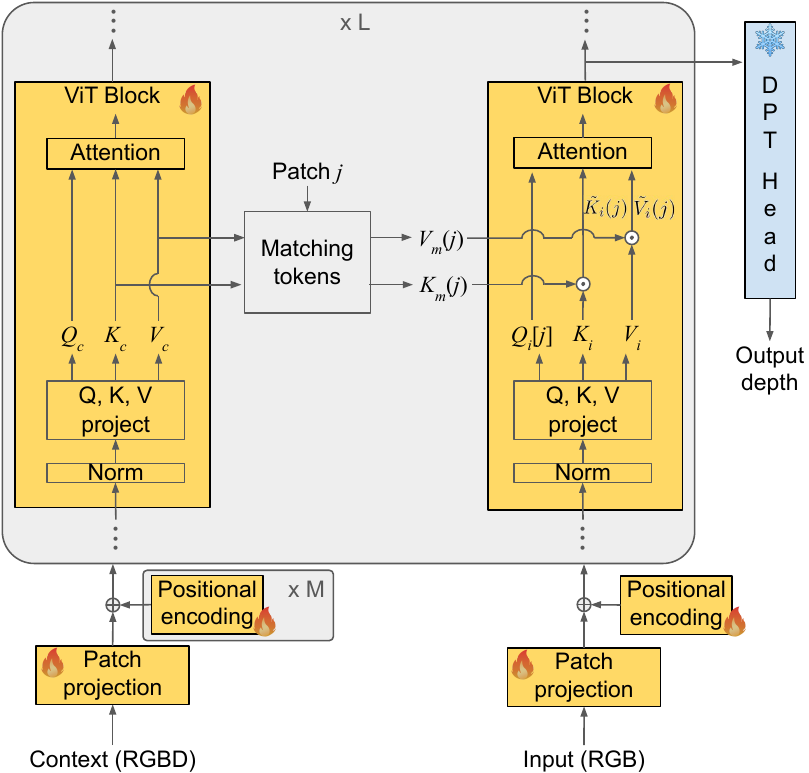}
    \caption{\textbf{Full network architecture.} The matched cross-attention mechanism allows the input stream to pull information from the context at the matched locations. Yellow blocks are optimized during training, while the blue block is frozen. Plus signs correspond to addition, the ``dot'' operation is concatenation.}
    \label{fig:full_network architecture}
\end{figure}

\begin{figure*}[tbh]
  \centering
  \renewcommand{\arraystretch}{0}  
  \begin{tabular}{@{}c@{}c@{}c@{}c@{}c@{}}
    \includegraphics[width=0.2\textwidth]{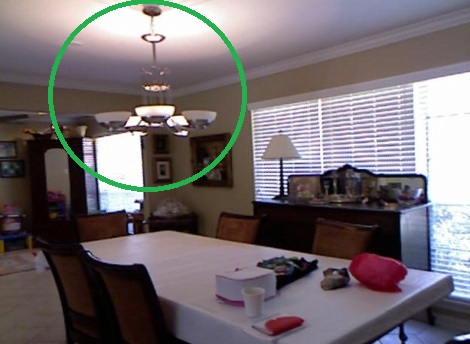} &
    \includegraphics[width=0.2\textwidth]{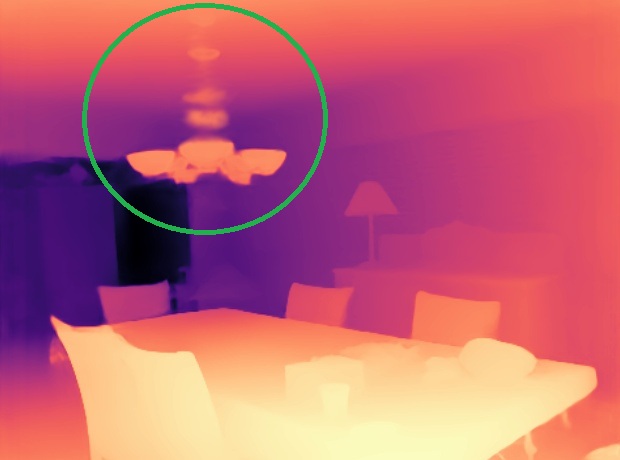} &
    \includegraphics[width=0.2\textwidth]{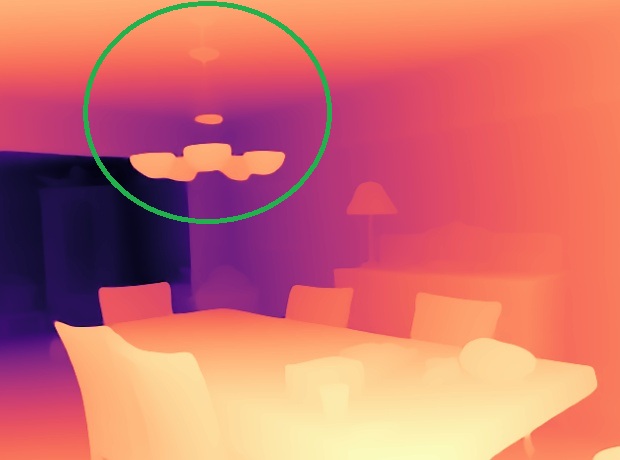} &
    \includegraphics[width=0.2\textwidth]{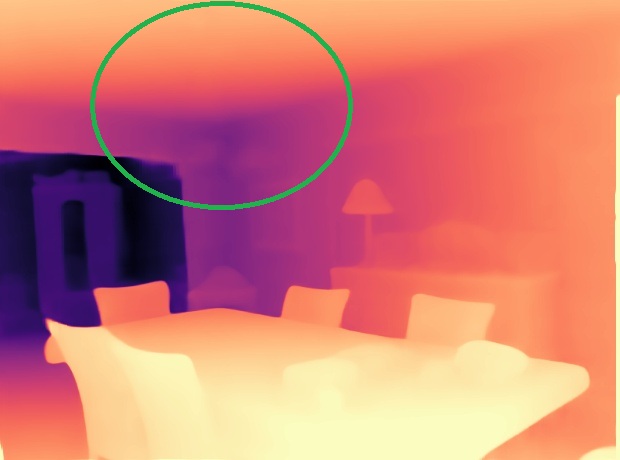} &
    \includegraphics[width=0.2\textwidth]{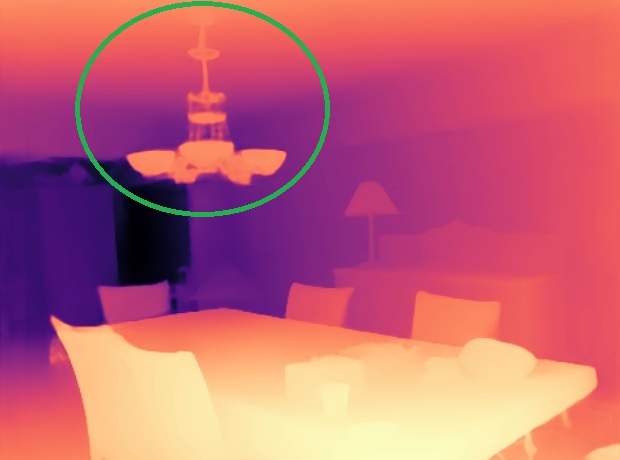} \\
    \includegraphics[width=0.2\textwidth]{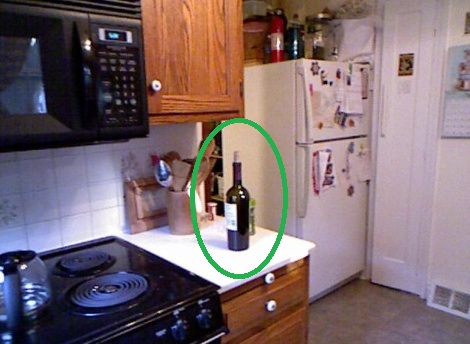} &
    \includegraphics[width=0.2\textwidth]{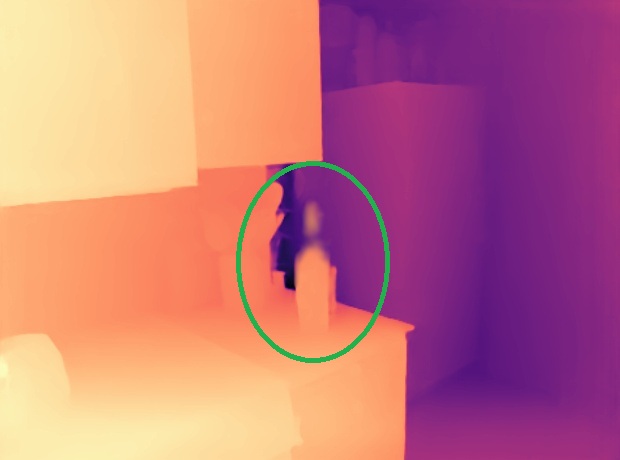} &
    \includegraphics[width=0.2\textwidth]{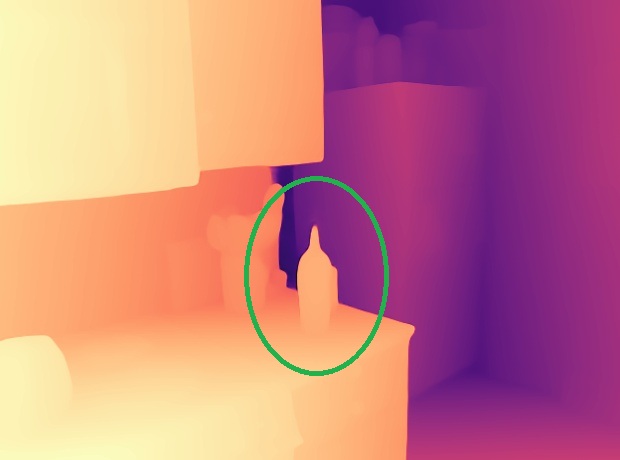} &
    \includegraphics[width=0.2\textwidth]{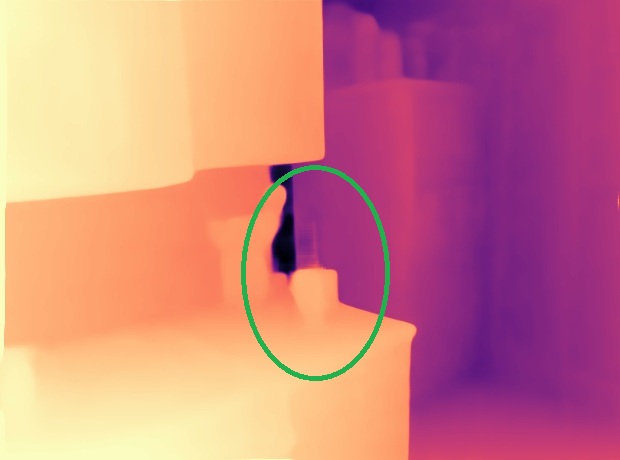} &
    \includegraphics[width=0.2\textwidth]{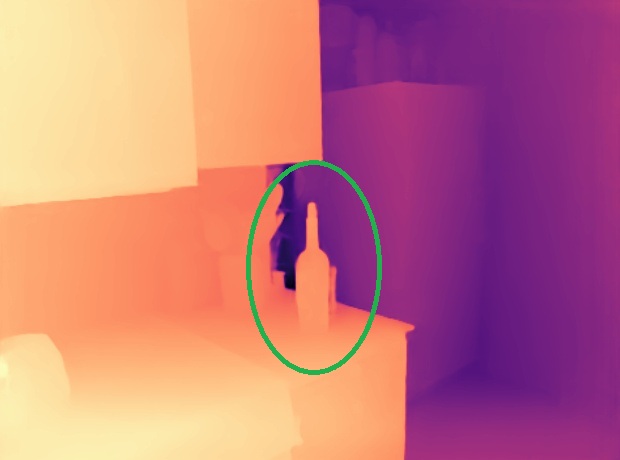} \\
    \includegraphics[width=0.2\textwidth]{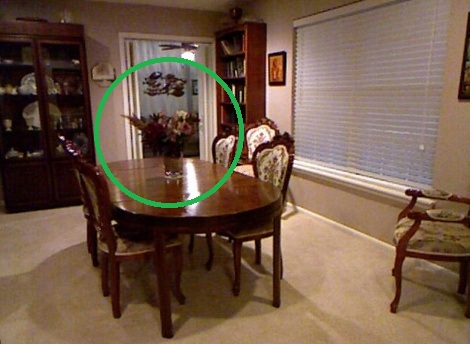} &
    \includegraphics[width=0.2\textwidth]{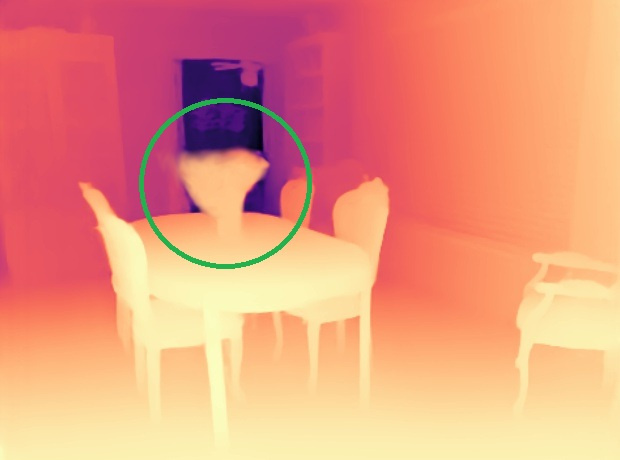} &
    \includegraphics[width=0.2\textwidth]{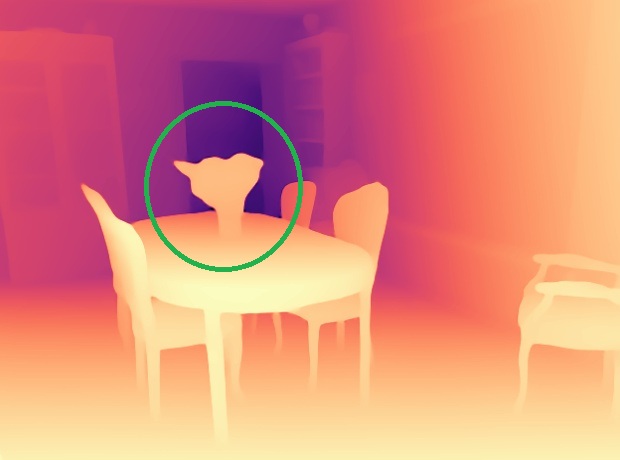} &
    \includegraphics[width=0.2\textwidth]{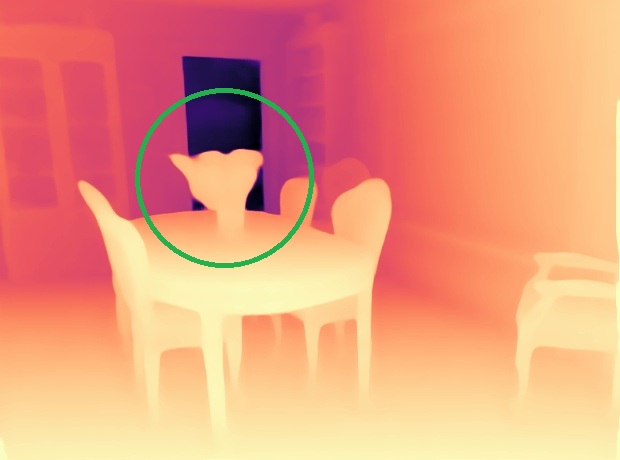} &
    \includegraphics[width=0.2\textwidth]{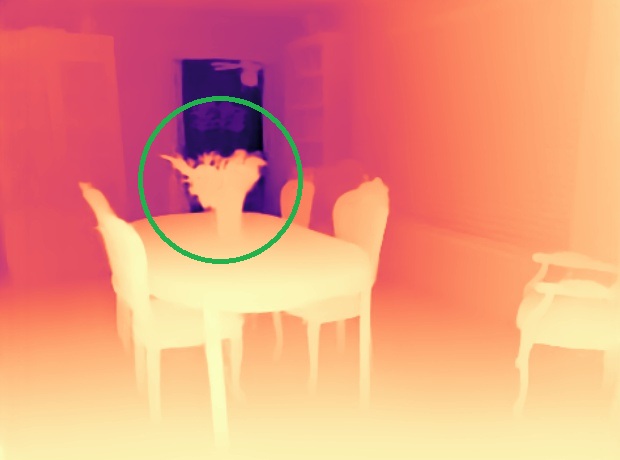} \\
    \includegraphics[width=0.2\textwidth]{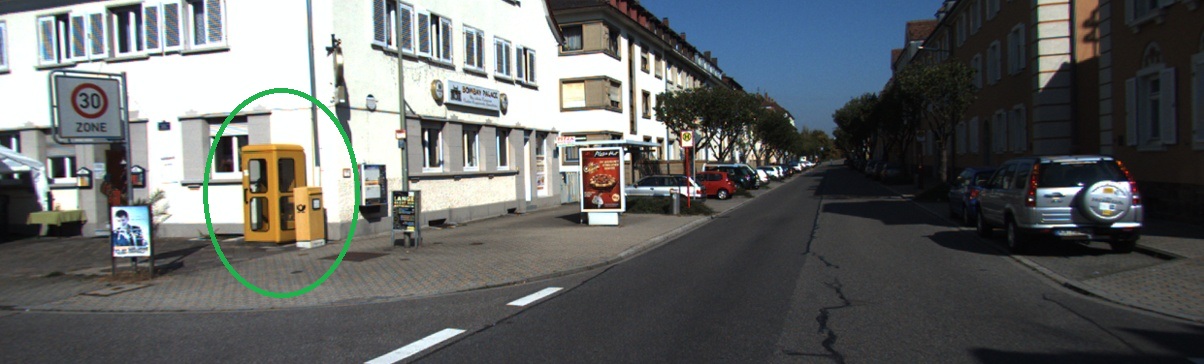} &
    \includegraphics[width=0.2\textwidth]{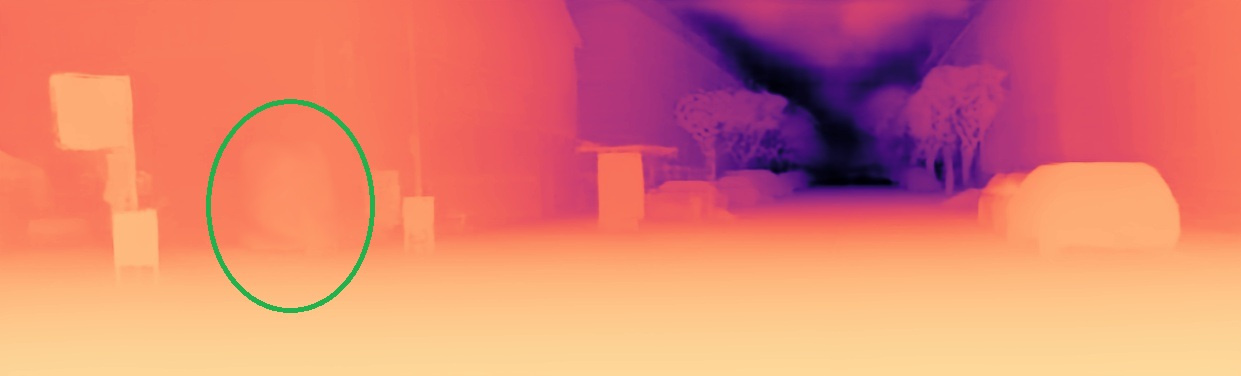} &
    \includegraphics[width=0.2\textwidth]{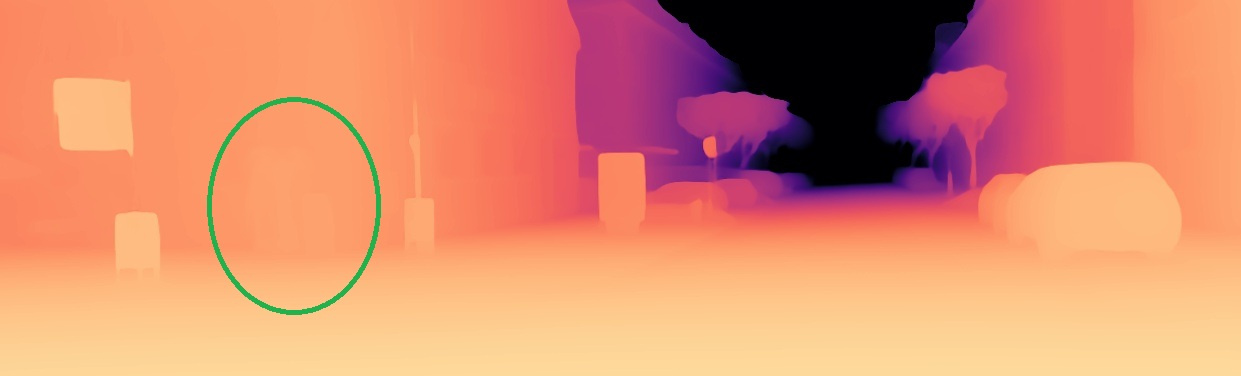} &
    \includegraphics[width=0.2\textwidth]{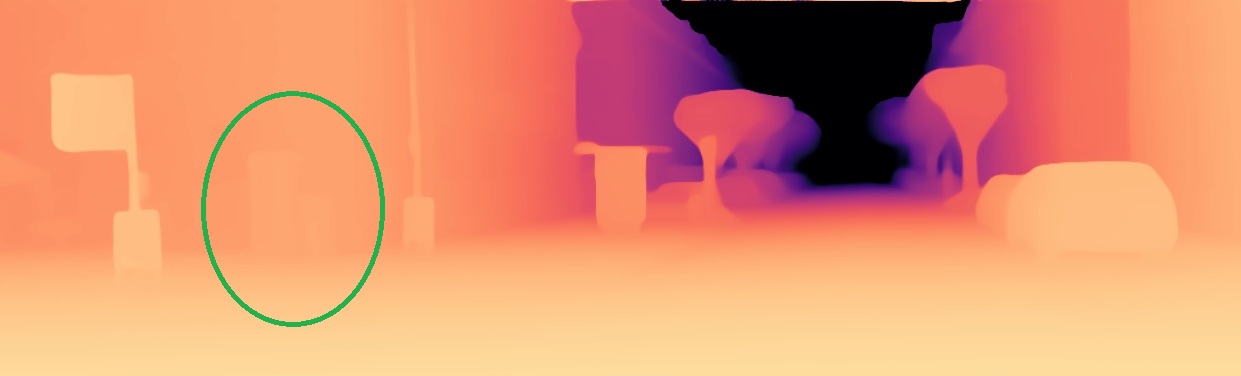} &
    \includegraphics[width=0.2\textwidth]{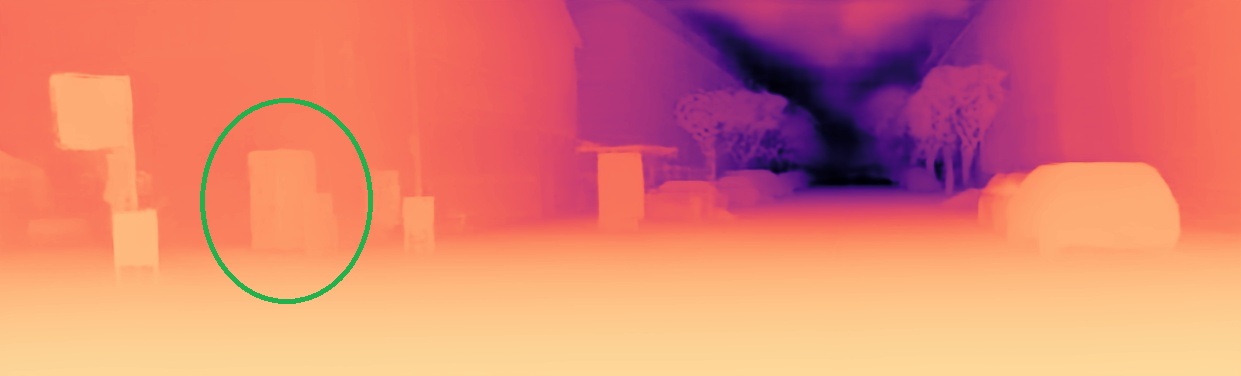} \\
    \includegraphics[width=0.2\textwidth]{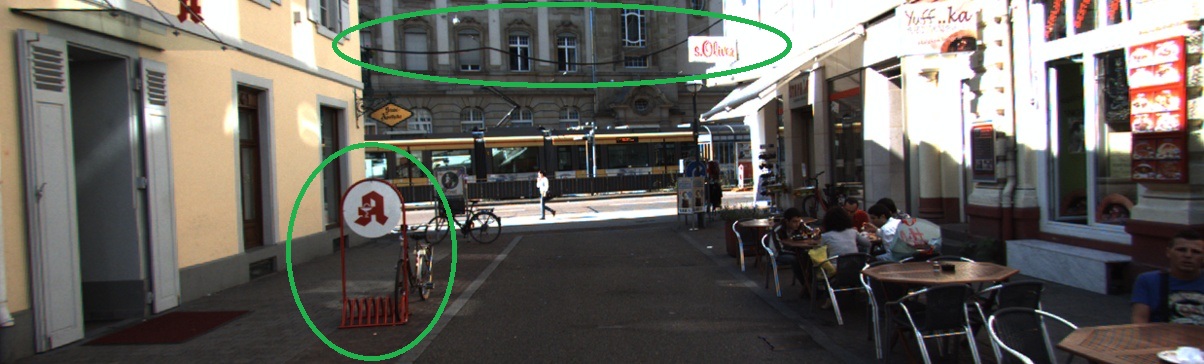} &
    \includegraphics[width=0.2\textwidth]{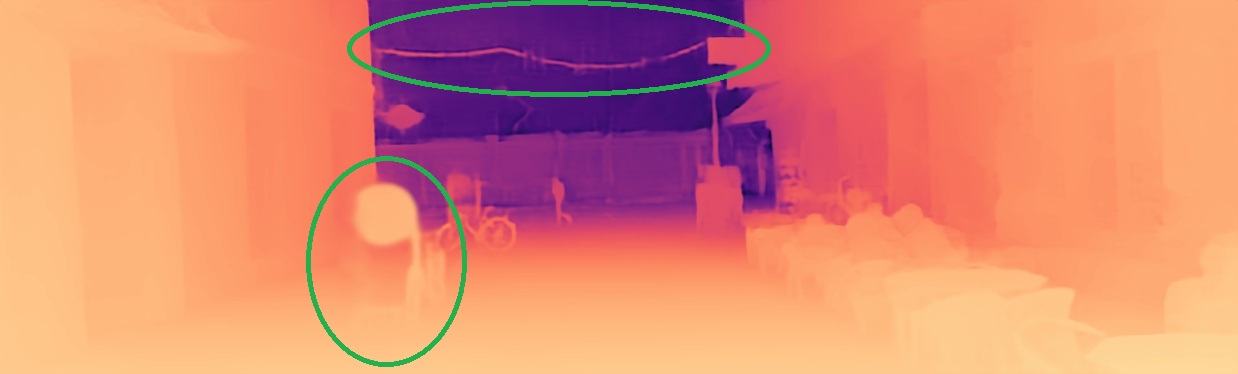} &
    \includegraphics[width=0.2\textwidth]{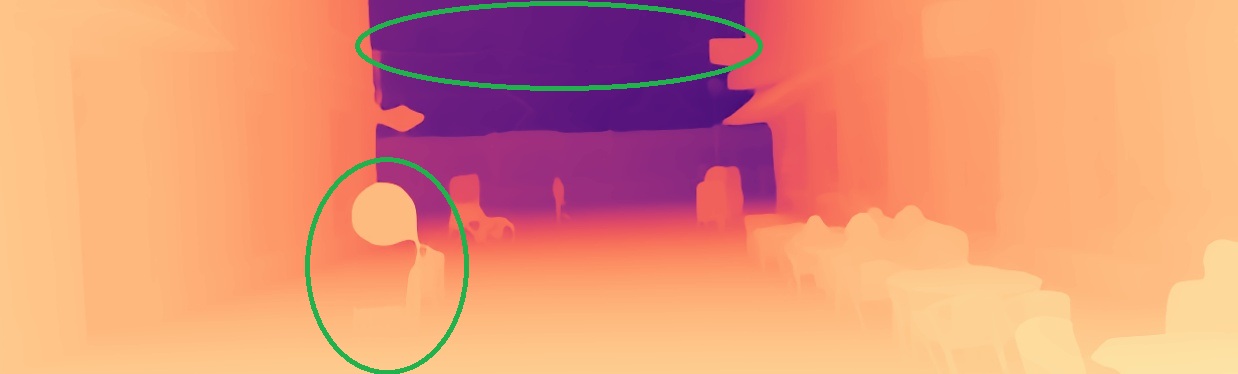} &
    \includegraphics[width=0.2\textwidth]{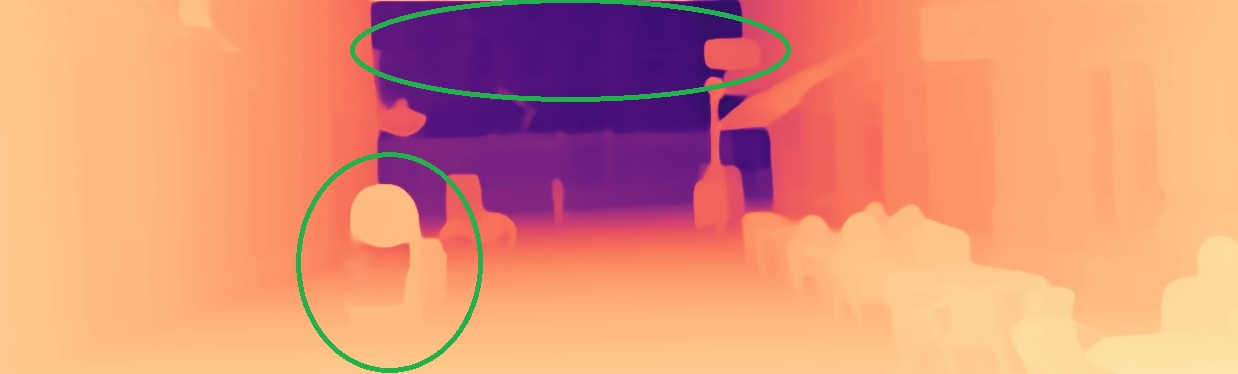} &
    \includegraphics[width=0.2\textwidth]{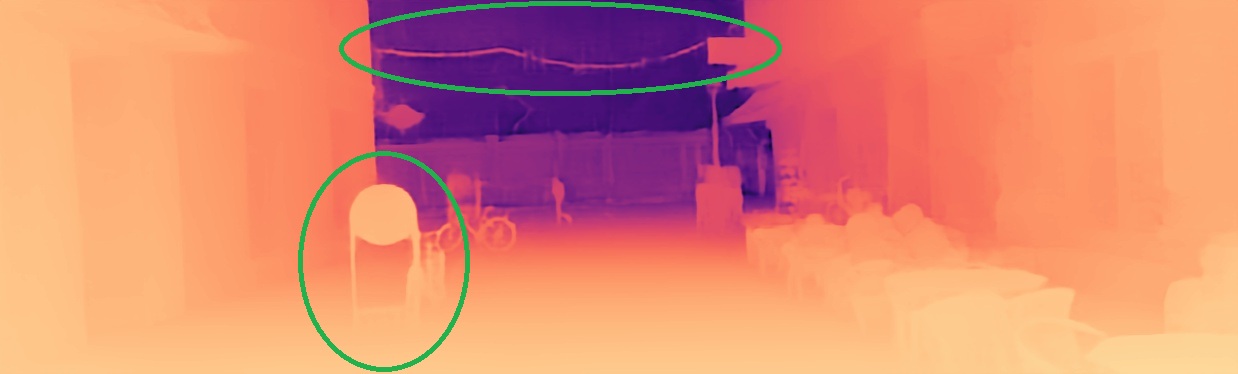} \\
    \includegraphics[width=0.2\textwidth]{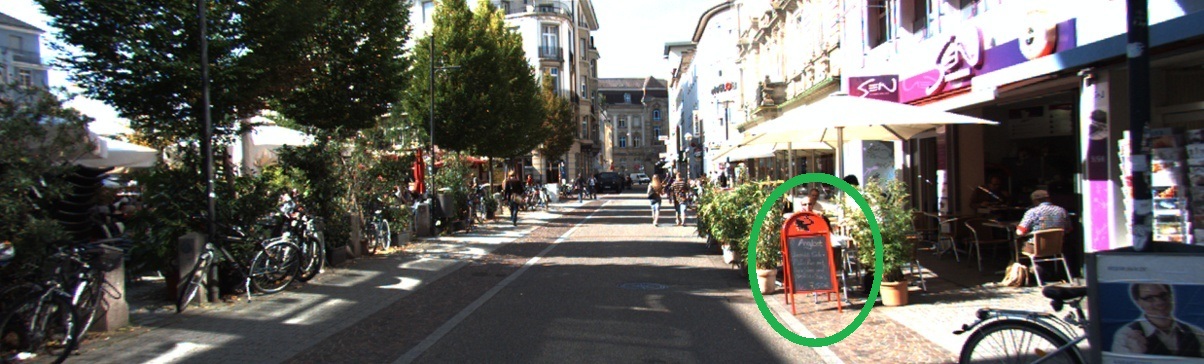} &
    \includegraphics[width=0.2\textwidth]{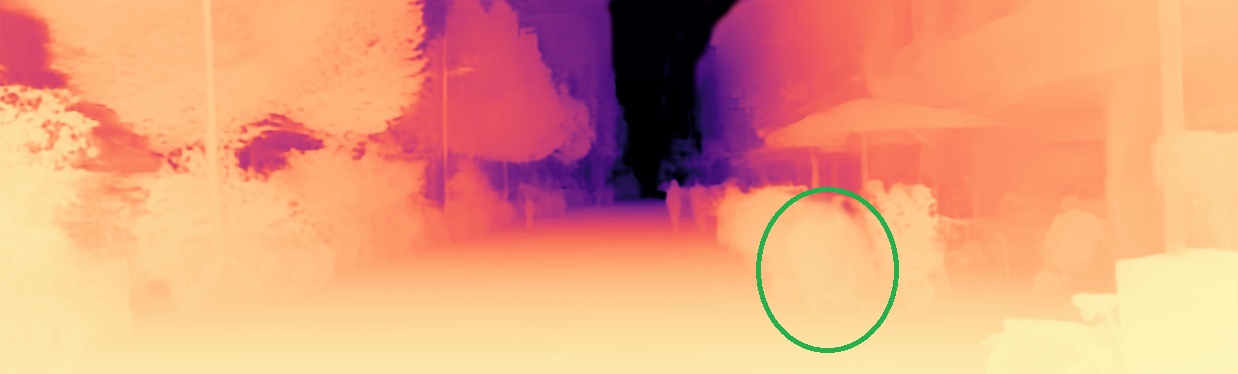} &
    \includegraphics[width=0.2\textwidth]{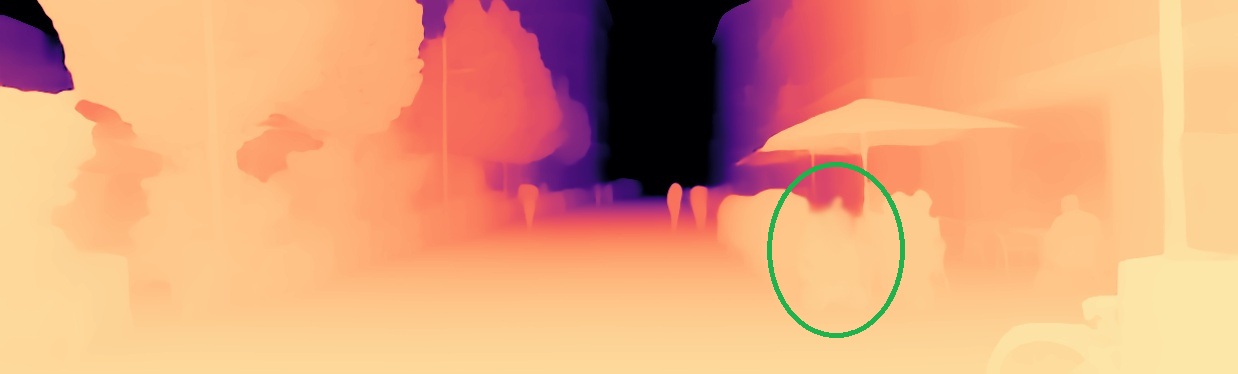} &
    \includegraphics[width=0.2\textwidth]{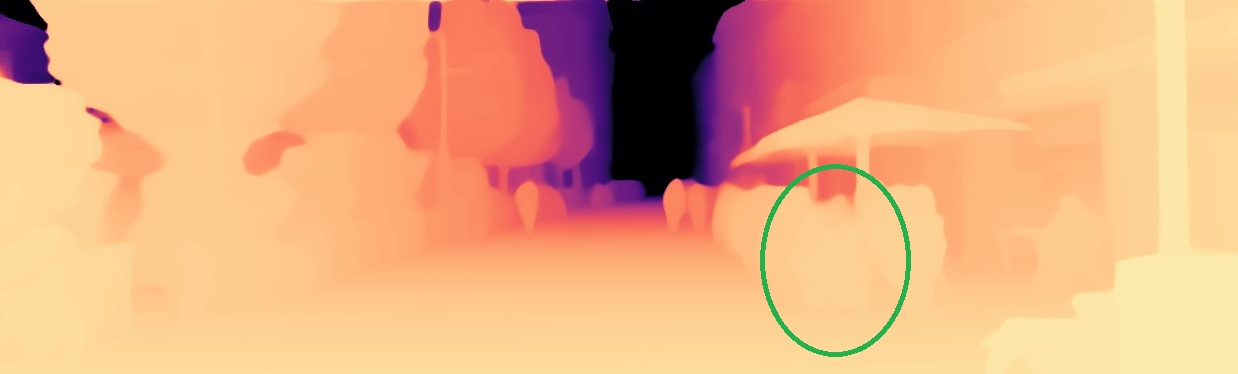} &
    \includegraphics[width=0.2\textwidth]{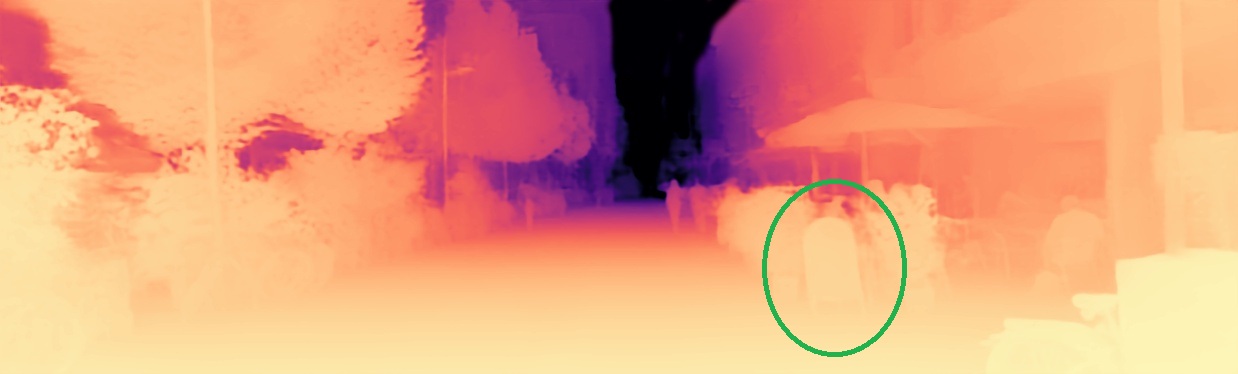} \\
    \includegraphics[width=0.2\textwidth]{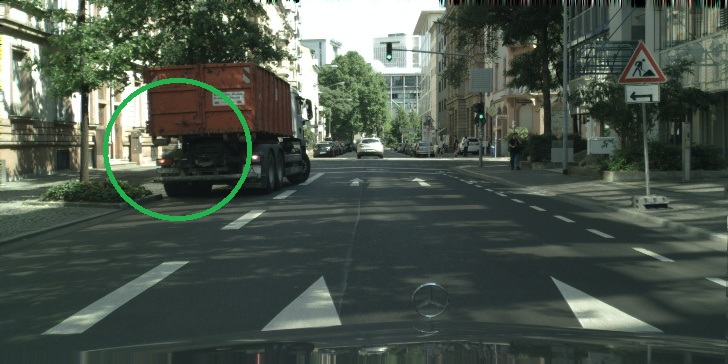} &
    \includegraphics[width=0.2\textwidth]{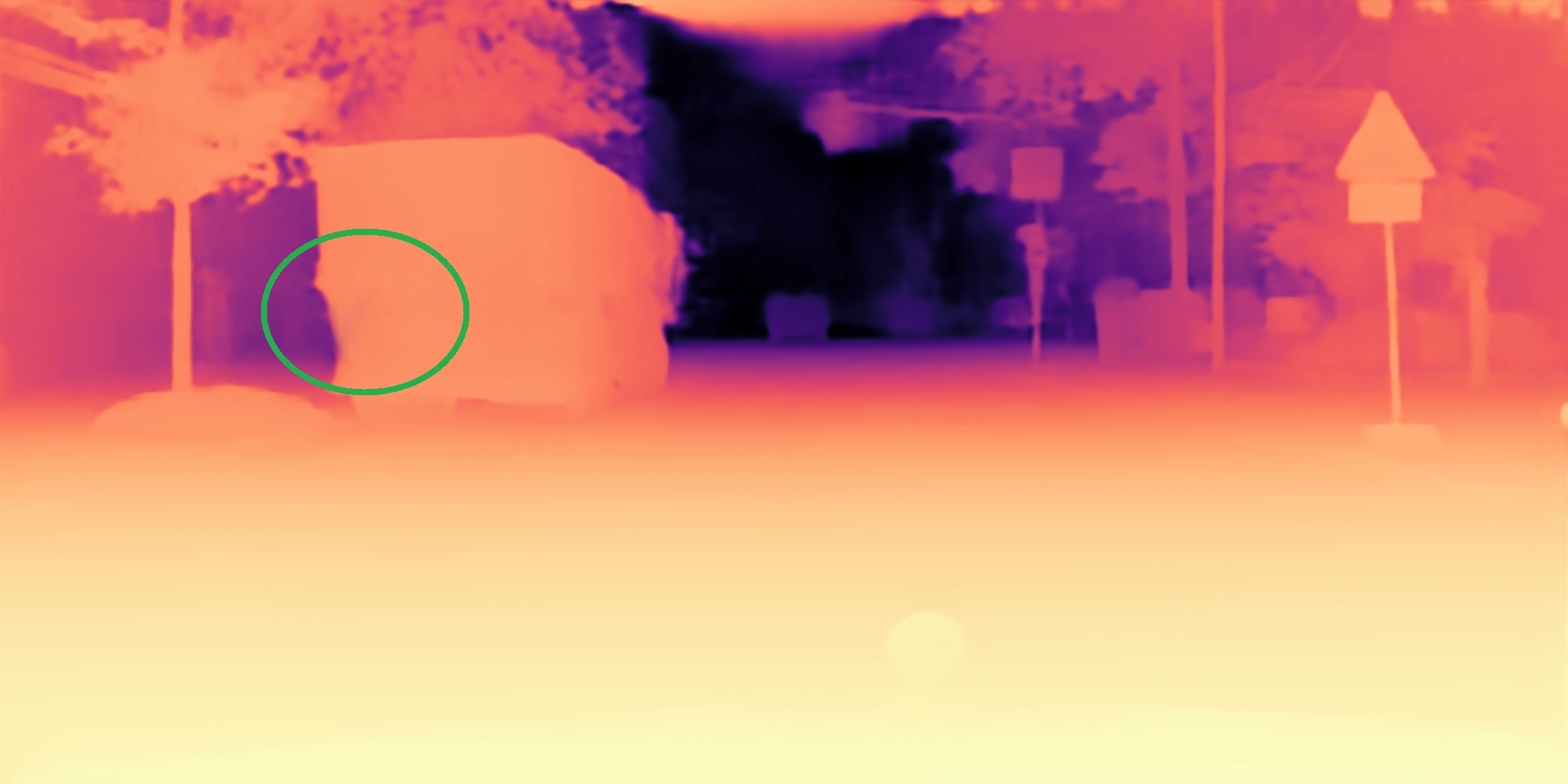} &
    \includegraphics[width=0.2\textwidth]{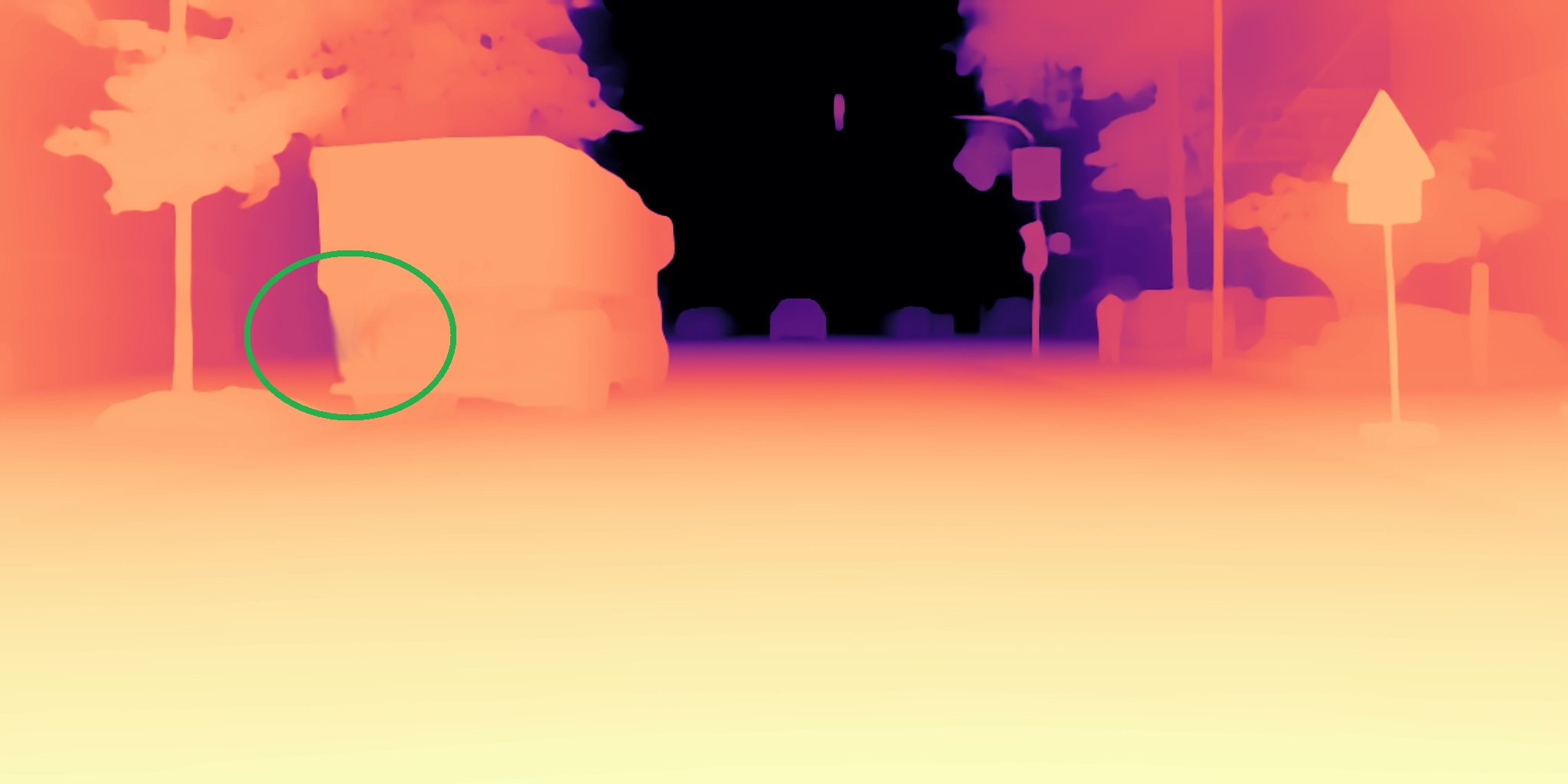} &
    \includegraphics[width=0.2\textwidth]{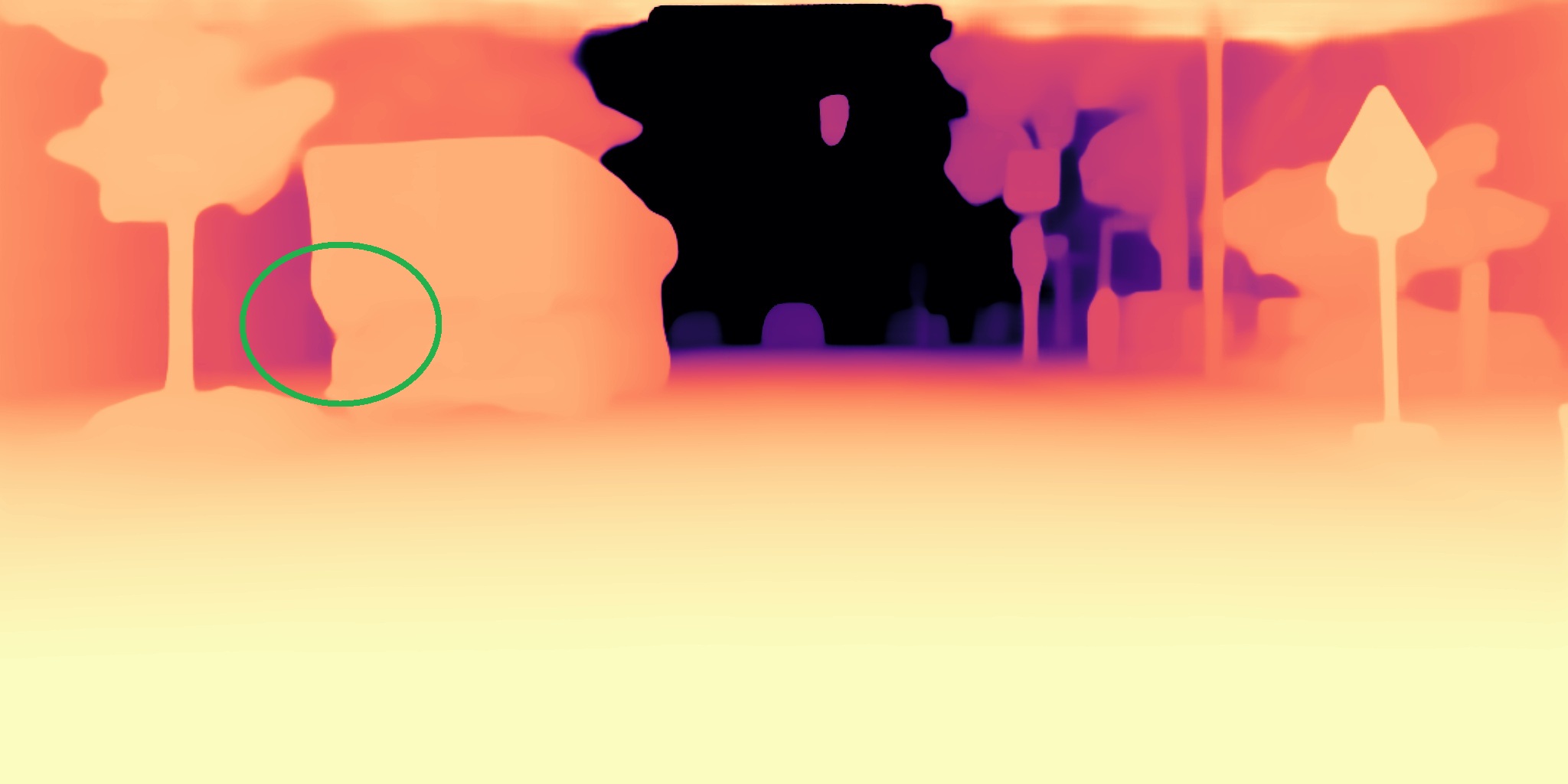} &
    \includegraphics[width=0.2\textwidth]{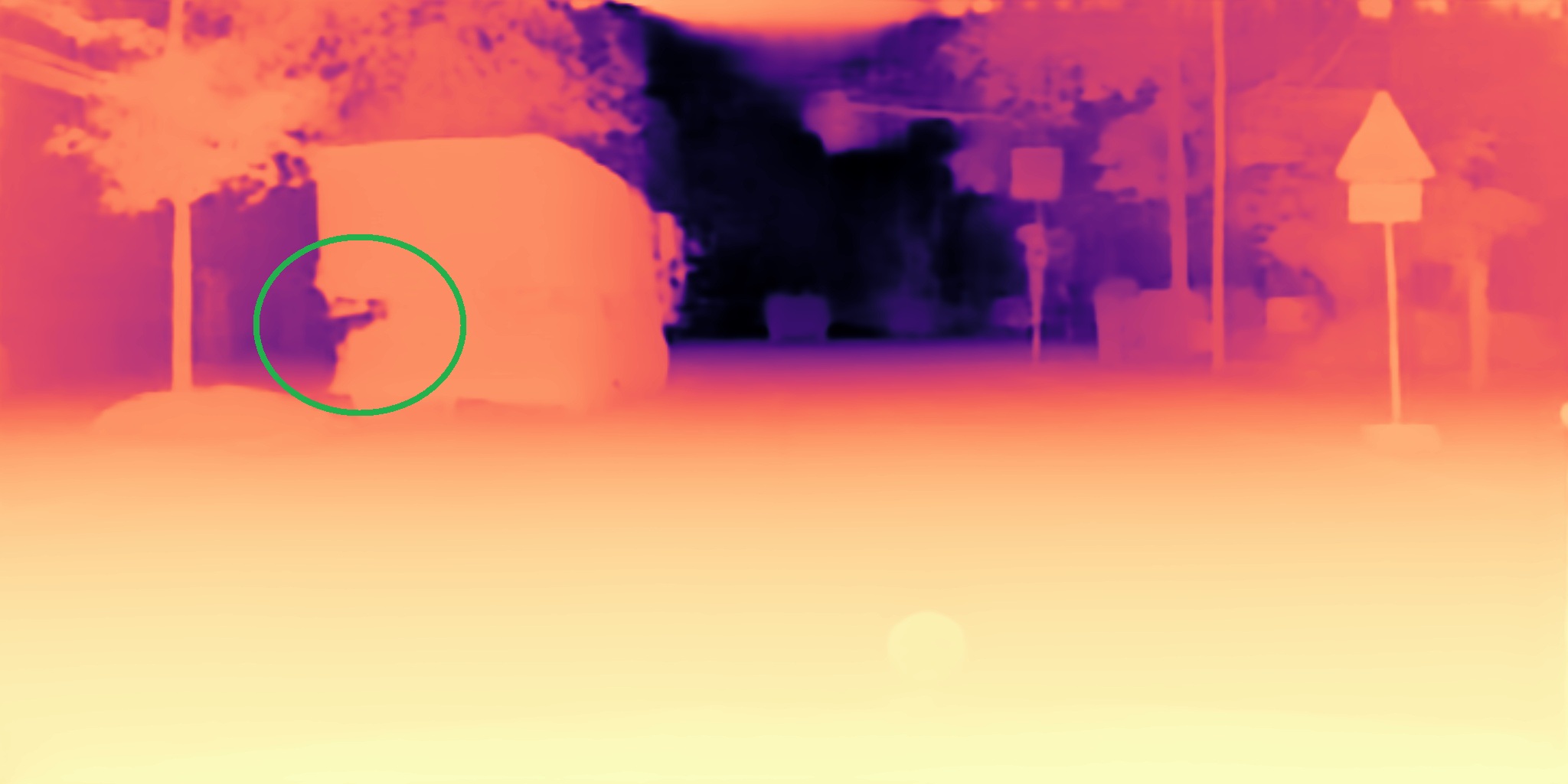} \\
    \includegraphics[width=0.2\textwidth]{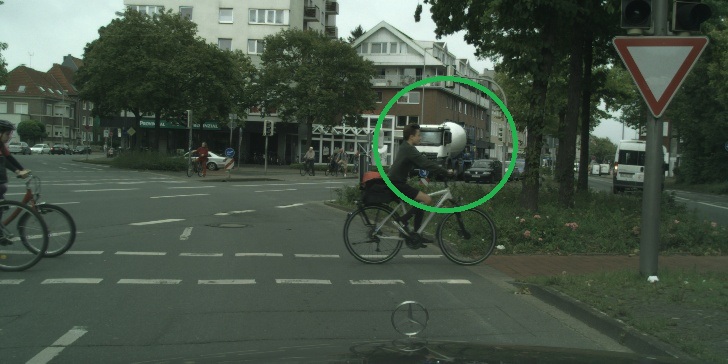} &
    \includegraphics[width=0.2\textwidth]{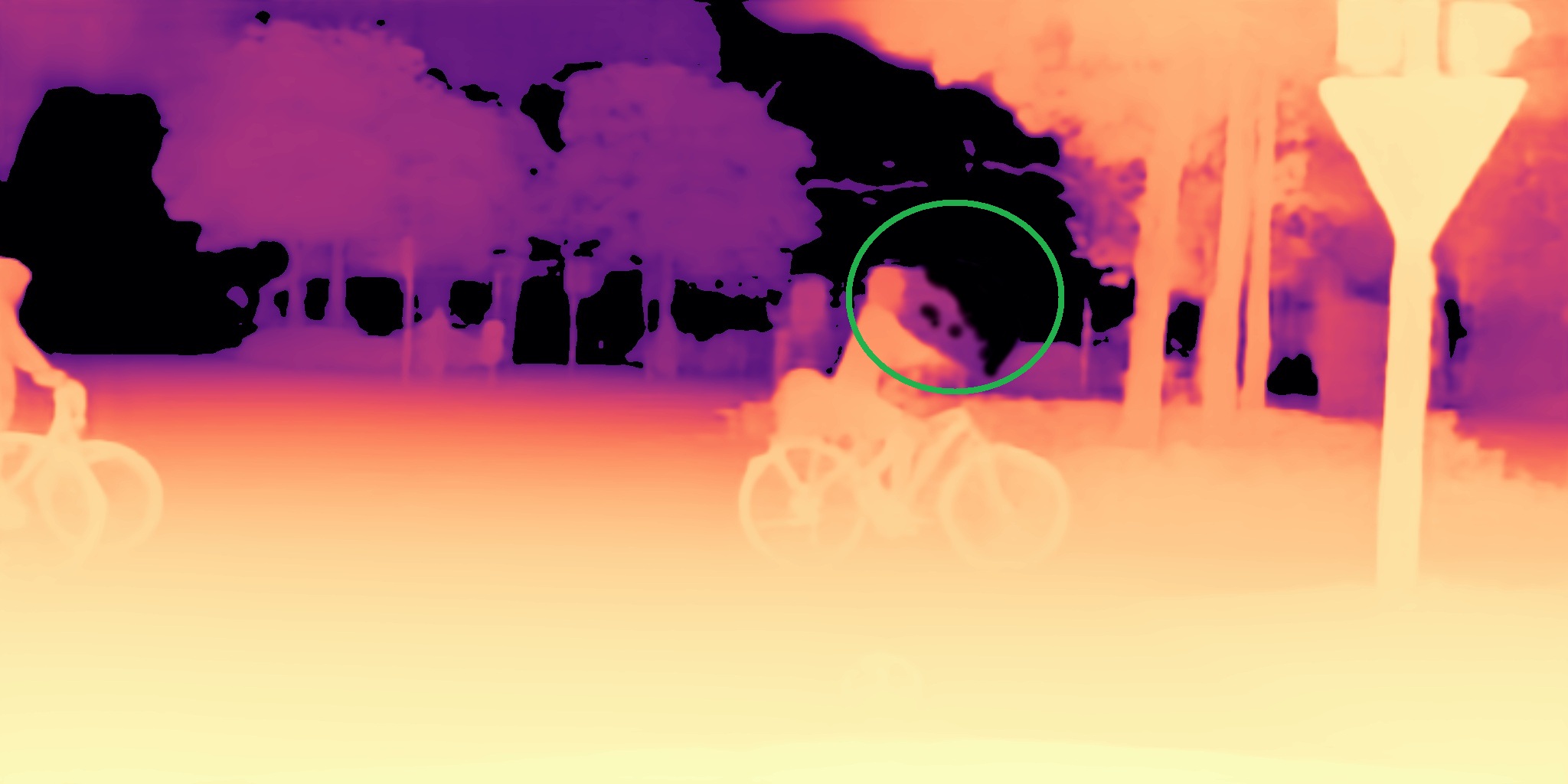} &
    \includegraphics[width=0.2\textwidth]{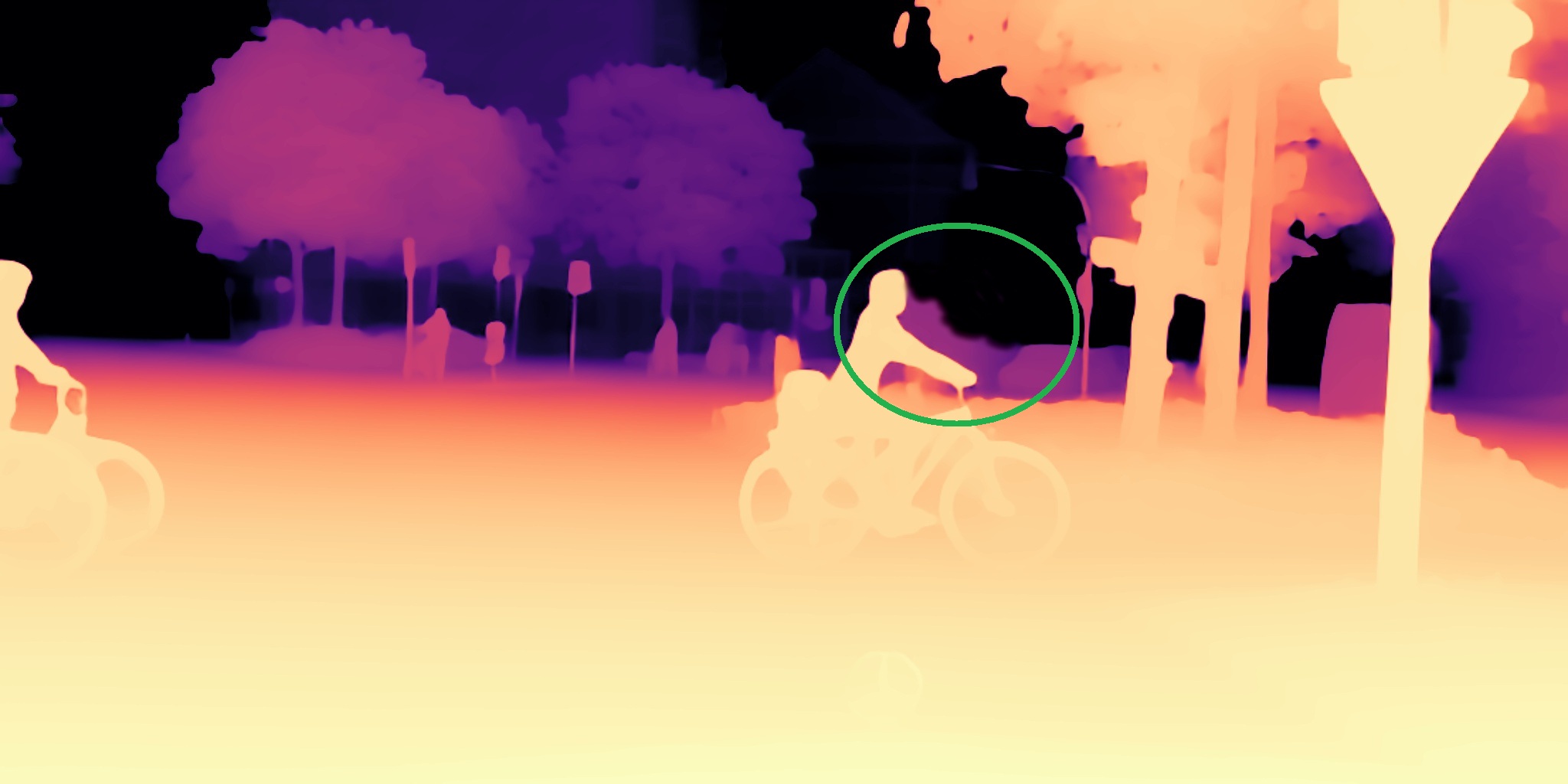} &
    \includegraphics[width=0.2\textwidth]{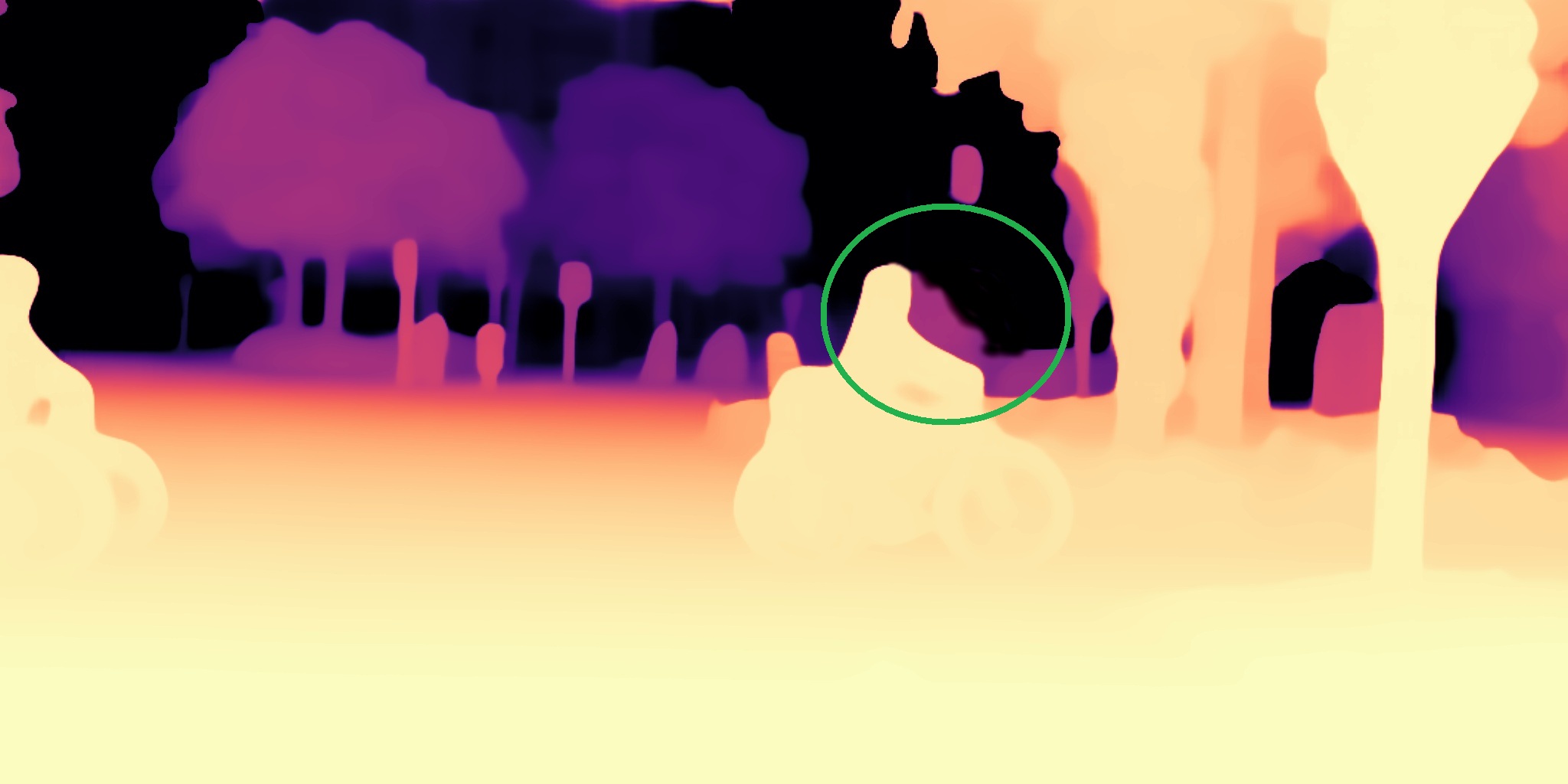} &
    \includegraphics[width=0.2\textwidth]{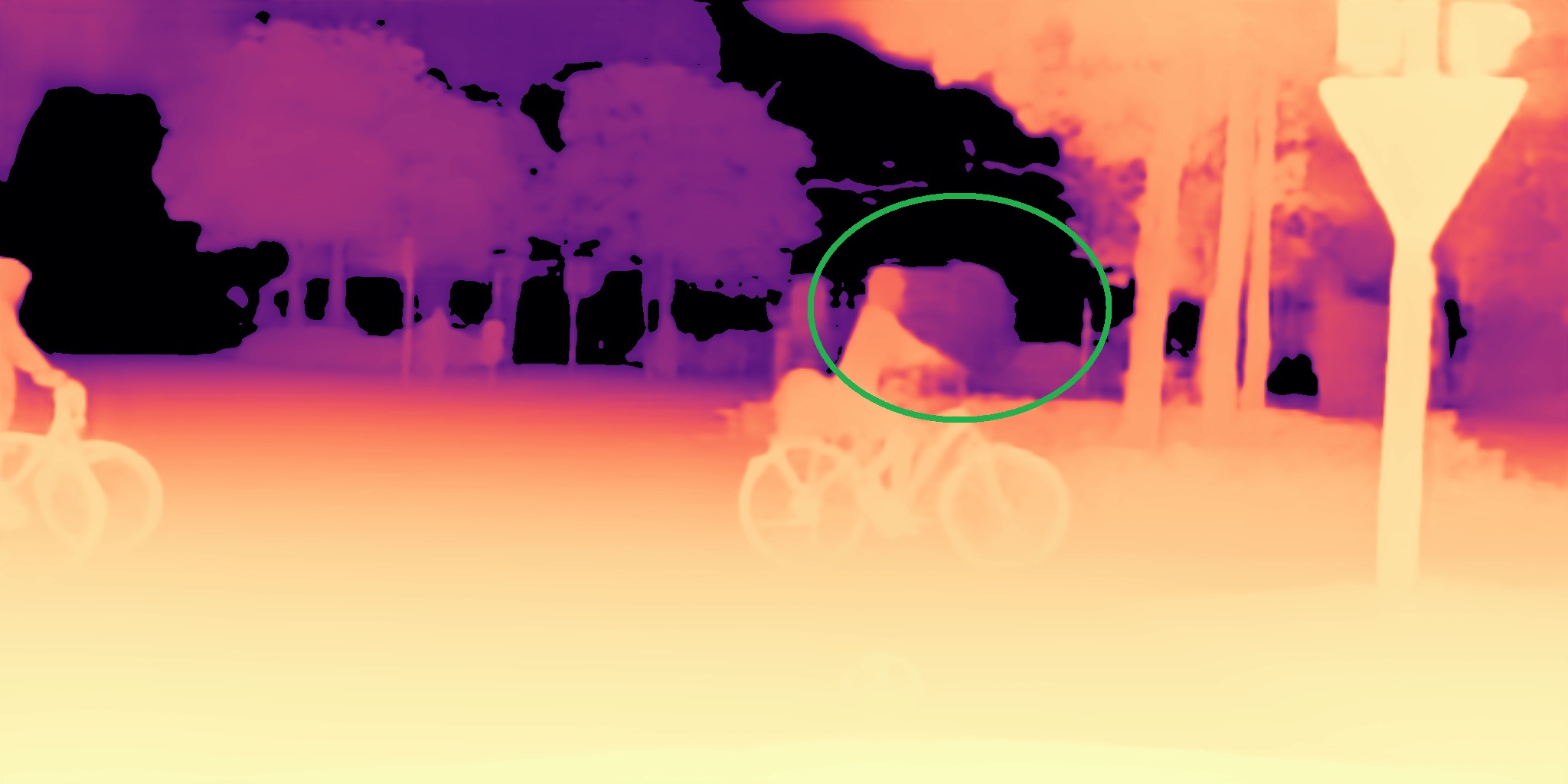} \\
    \includegraphics[width=0.2\textwidth]{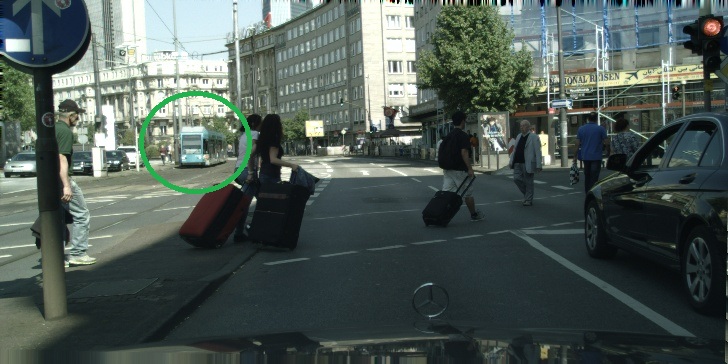} &
    \includegraphics[width=0.2\textwidth]{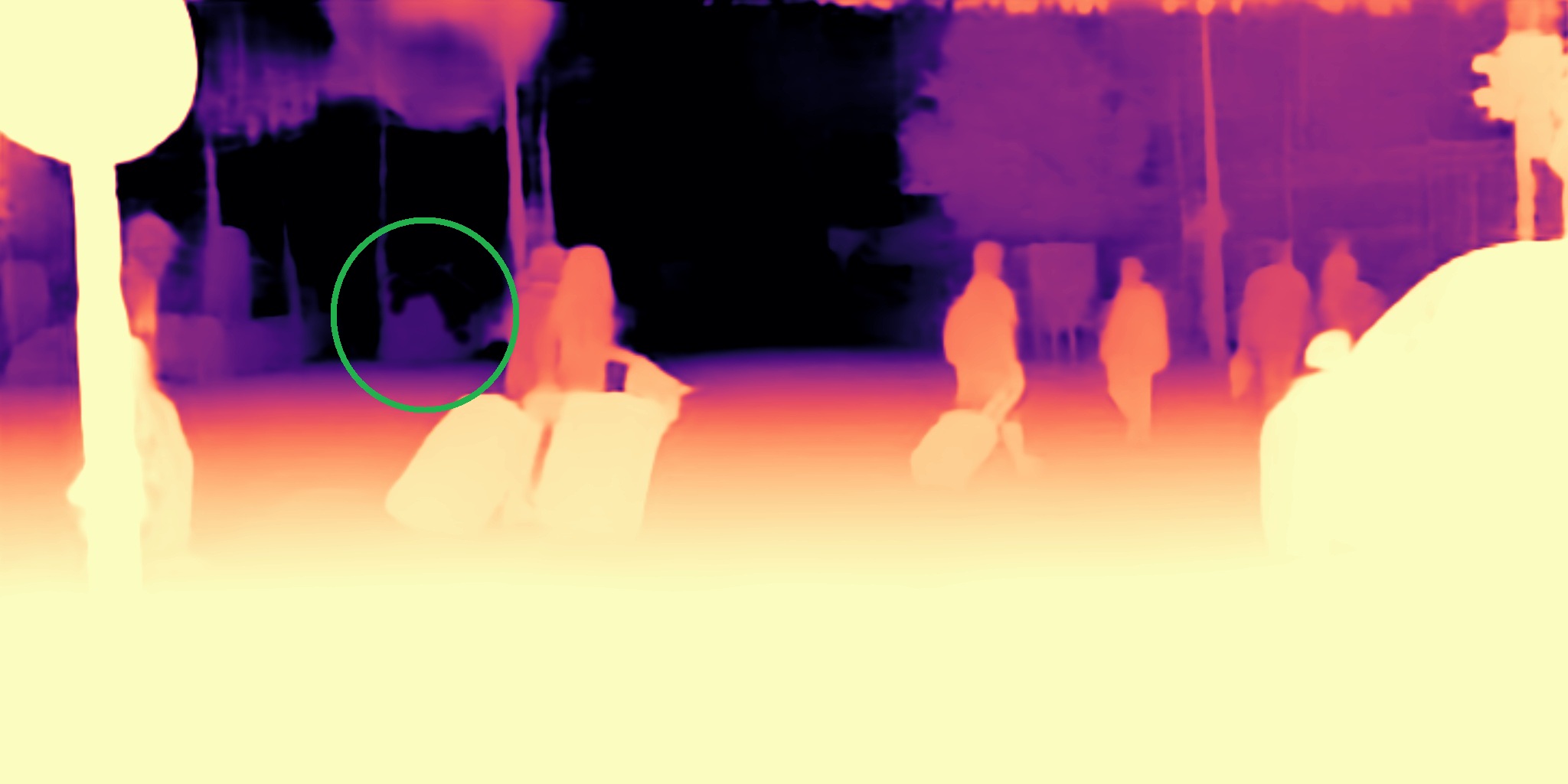} &
    \includegraphics[width=0.2\textwidth]{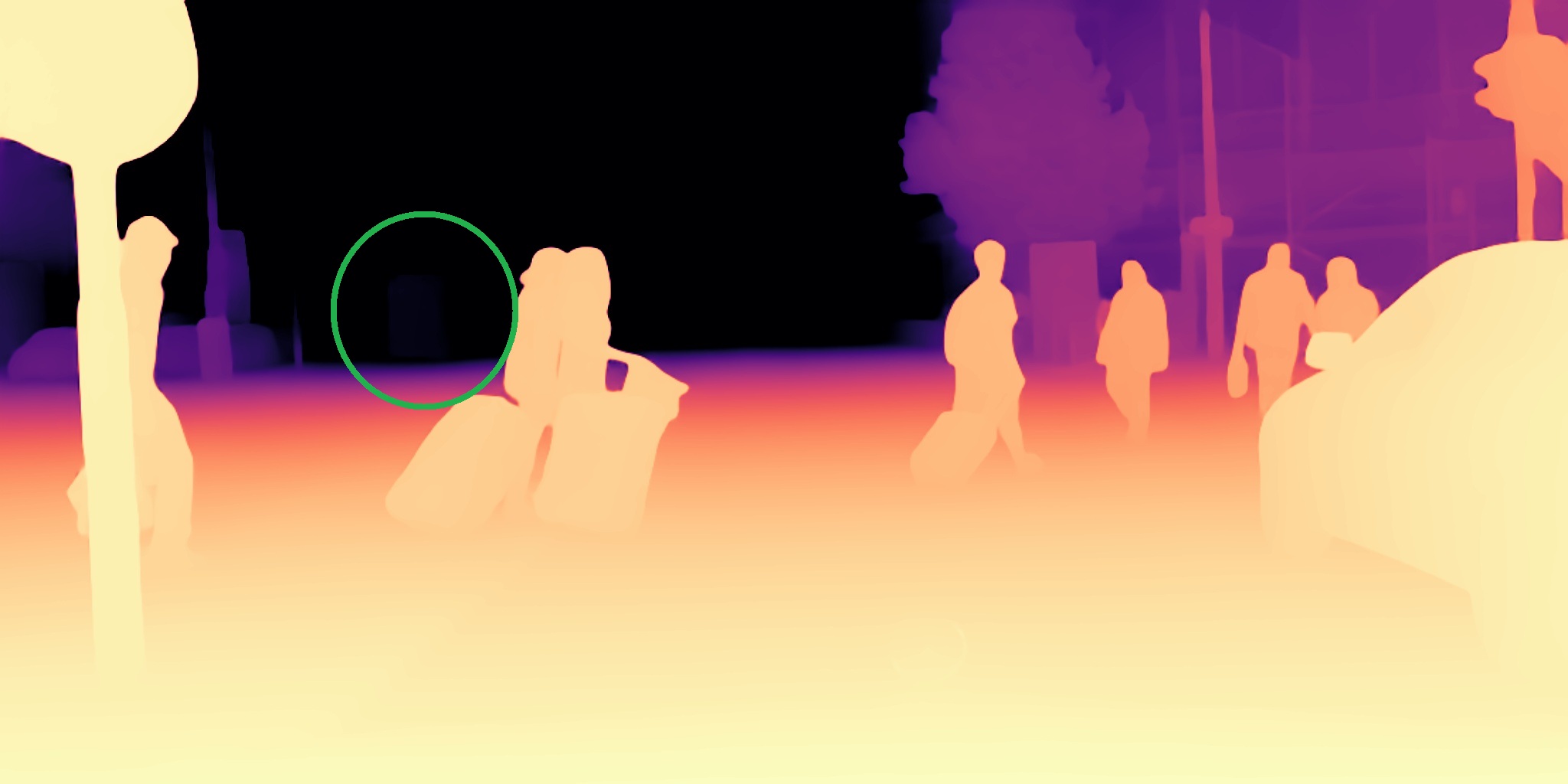} &
    \includegraphics[width=0.2\textwidth]{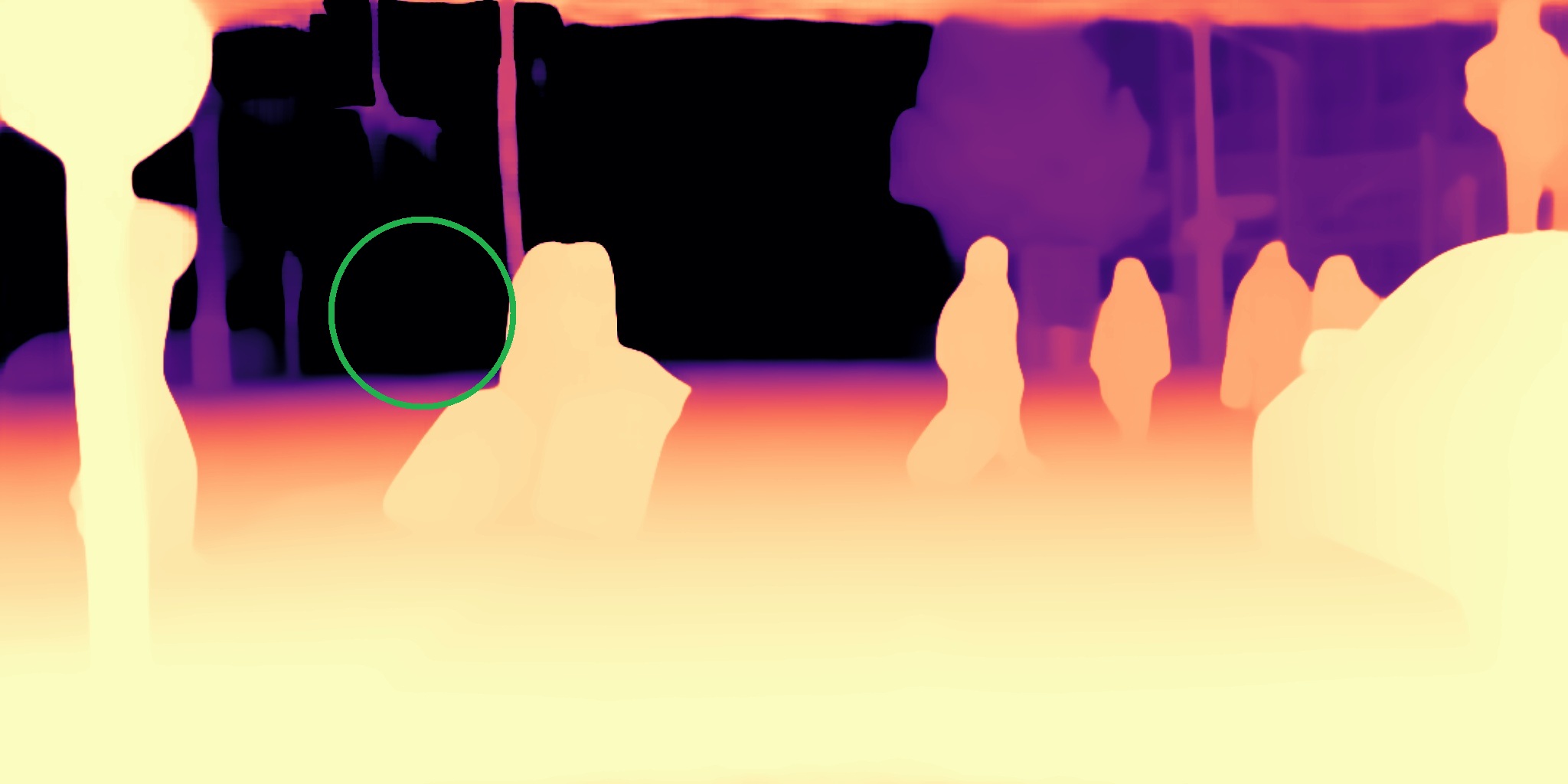} &
    \includegraphics[width=0.2\textwidth]{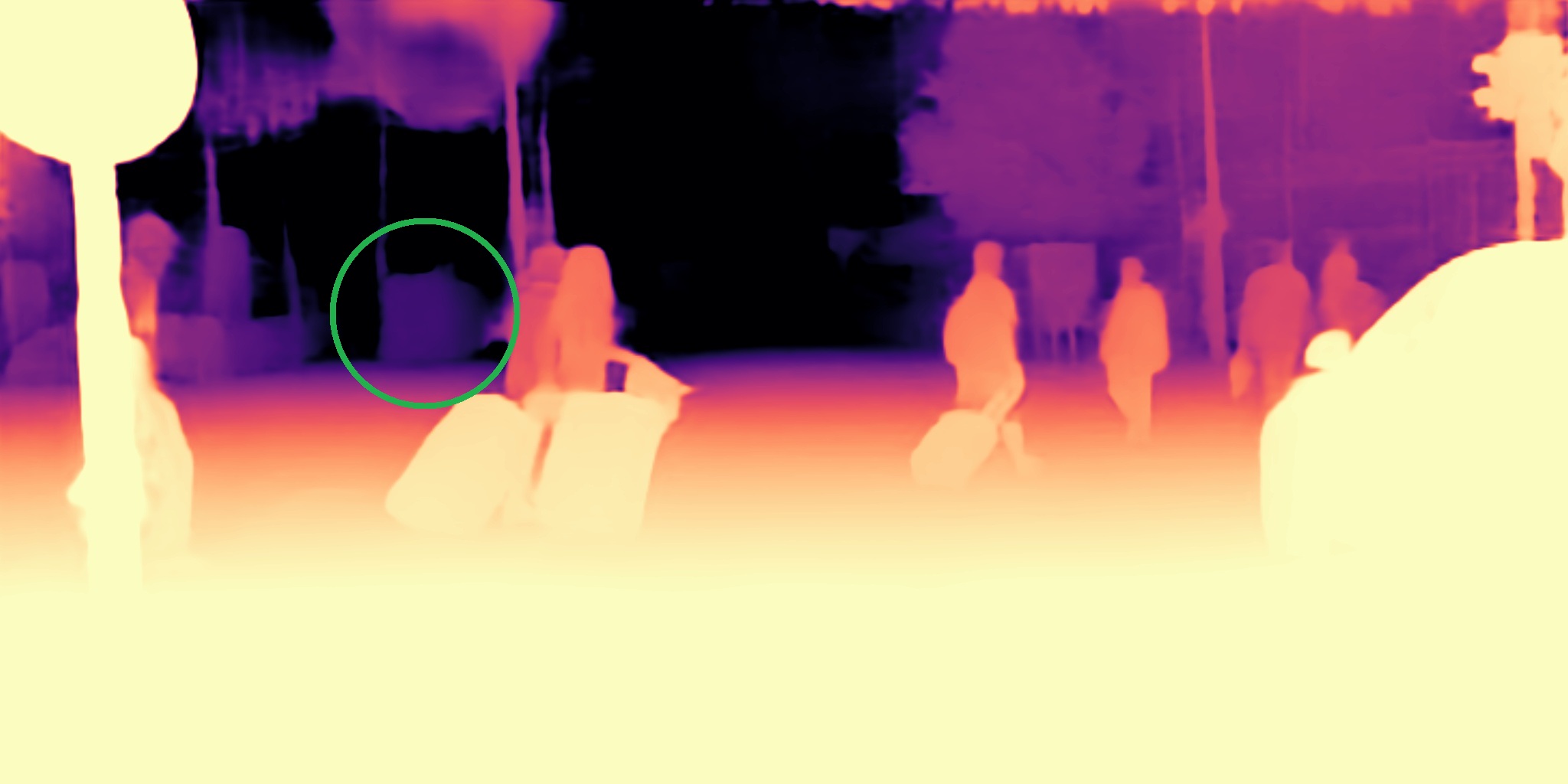} \\ [0.5ex]  
    Input & DepthAnything v2 & UniDepth v2 & Metric3D v2& RAD
  \end{tabular}
  \caption{\textbf{Qualitative results} for NYU Depth v2 (top three rows), KITTI (middle three rows) and Cityscapes (bottom three rows). We compare our method (RAD) to baselines DepthAnything v2 \cite{DepthAnything_V2}, UniDepth v2 \cite{UniDepth_V2} and Metric3D v2 \cite{metric3d_v2}. Best viewed zoomed in.}
  \label{fig:additional qualitative results}
\end{figure*}

\section{Overall Network Architecture}

Fig.~\ref{fig:full_network architecture} illustrates the complete network architecture employed in RAD.
The design is based on the DepthAnything v2 framework \cite{DepthAnything_V2}, with one key modification: the original attention module is replaced by our matched cross-attention mechanism, which connects between the context and input streams. The remainder of the encoder's Vision Transformer (ViT) architecture remains unchanged.

The system comprises a dual-stream encoder, which is trainable during optimization, and a frozen DPT head \cite{dpt}. For each token $j$ in the input stream, the \textit{matching tokens} block retrieves matched keys $K_m(j)$ and values $V_m(j)$, derived from the context keys $K_c$ and values $V_c$ at corresponding matching positions. These matched representations are concatenated with the input keys $K_i$ and values $V_i$, forming the input to the standard scaled dot-product attention mechanism within the ViT. The input and context queries, $Q_i[j]$ and $Q_c$, are processed following the original DepthAnything v2 approach.

Importantly, each stream employs its own \textit{positional encoding}, enabling the encoder to distinguish between context and input tokens. Additionally, each retrieved image is assigned a unique positional encoding, allowing the tokens to remain distinguishable within the upstream blocks of the ViT. Consequently, if $M$ context samples are retrieved, the context stream utilizes $M$ distinct positional encodings. All positional encoders are trainable to support this differentiation.

The \textit{patch projection} for the input stream follows the original design of the DepthAnything v2 network. For the context stream, we introduce a fourth channel to process the sample's depth, as detailed in Sec.~\ref{sec:depth estimation network}. The procedures for \textit{normalization} and \textit{Q, K, V projection} are identical to those employed in each ViT block of DepthAnything v2. Similarly, the \textit{attention} mechanism remains unchanged, utilizing scaled dot-product attention.

\begin{figure*}[tbh]
    \centering
    \begin{subfigure}[b]{0.33\textwidth}
        \centering
        \includegraphics[width=\textwidth]{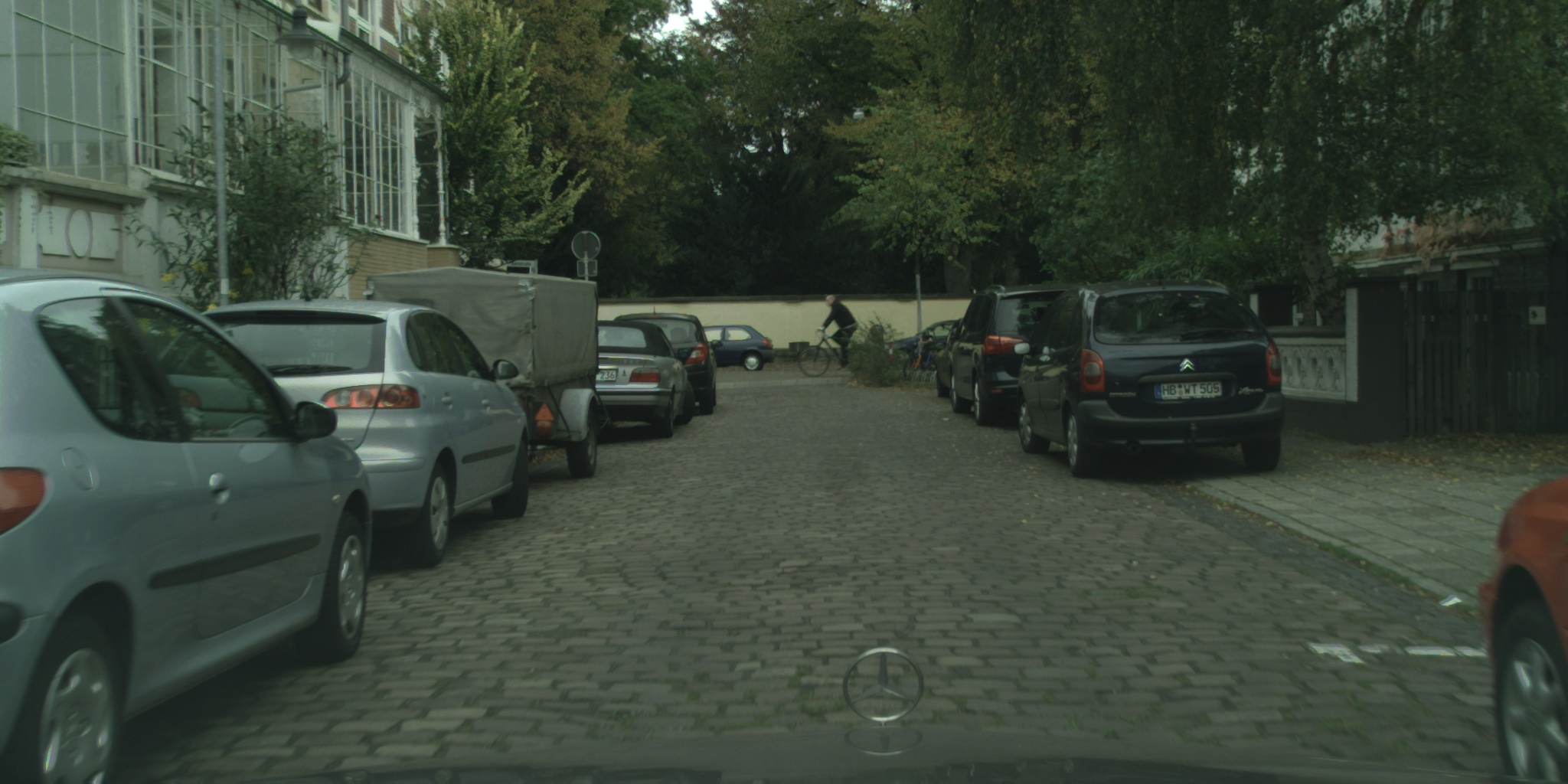}
        \caption*{RGB input} 
    \end{subfigure}
    \begin{subfigure}[b]{0.33\textwidth}
        \centering
        \includegraphics[width=\textwidth]{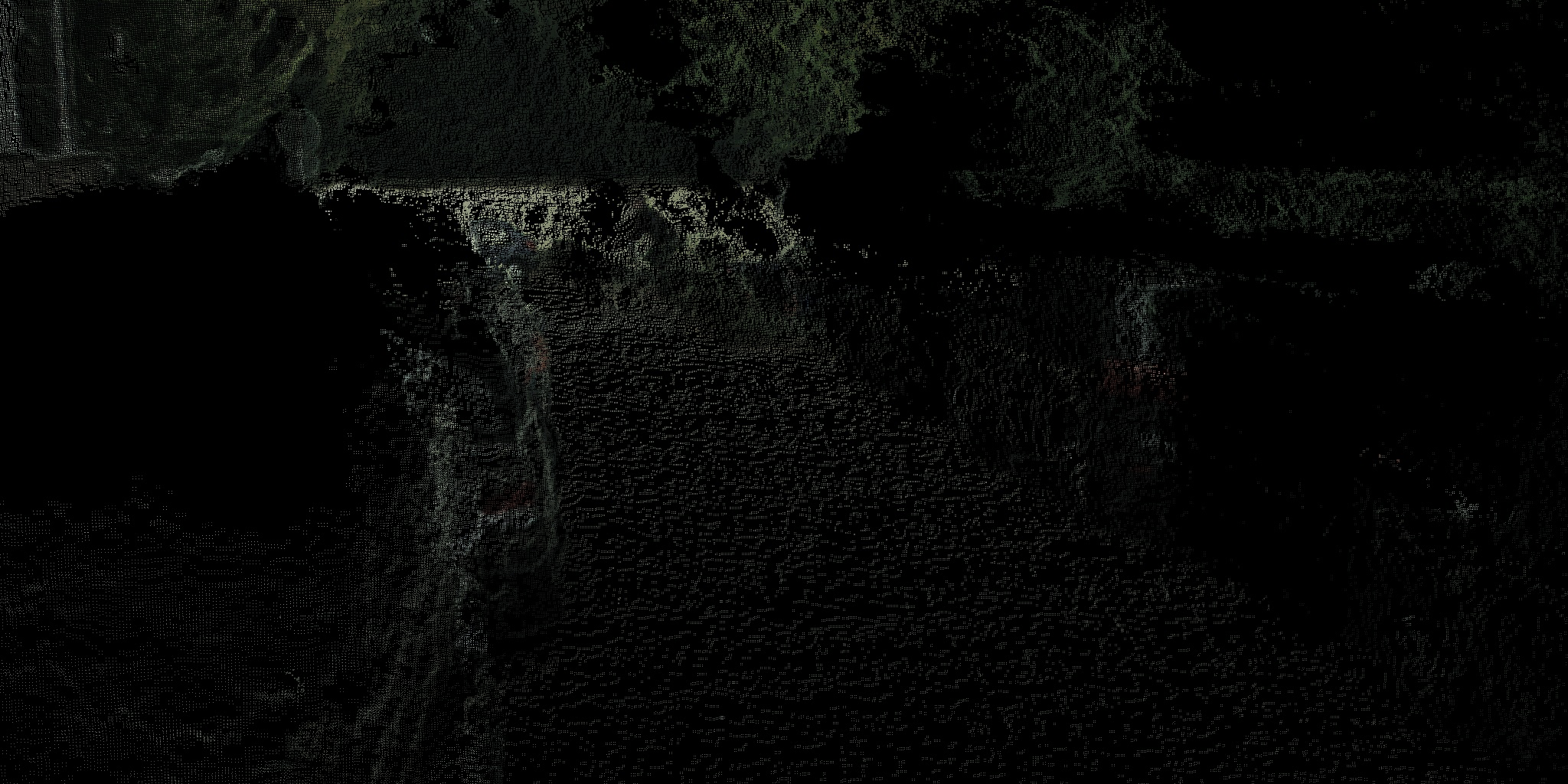}
        \caption*{Point cloud projection} 
    \end{subfigure}
    \begin{subfigure}[b]{0.33\textwidth}
        \centering
        \includegraphics[width=\textwidth]{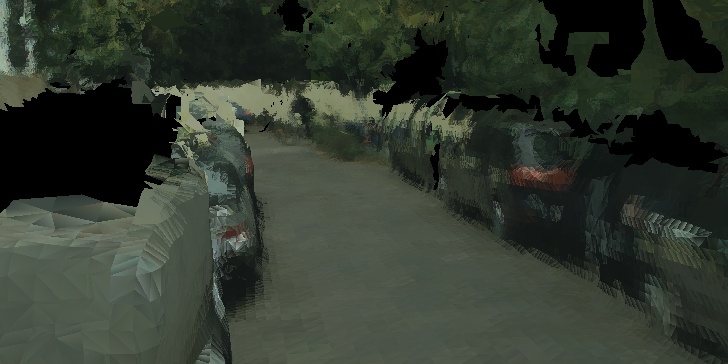}
        \caption*{Mesh projection} 
    \end{subfigure}
    \caption{\textbf{Mesh creation for 3D augmentation.} The input image (left) is projected from a new point of view. When using only the point cloud defined by the image pixels, projection yields large unmeasured regions (middle). When a mesh is used to reconstruct geometry, the scene is densely reconstructed.}
    \label{fig:mesh creation}
\end{figure*}

\begin{figure*}[tbh]
  \centering
  \includegraphics[width=\linewidth]{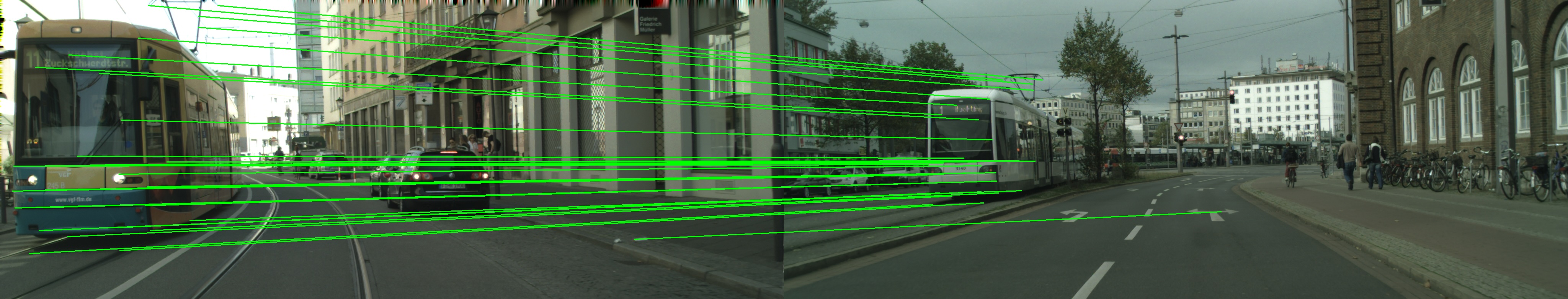}
  \includegraphics[width=\linewidth]{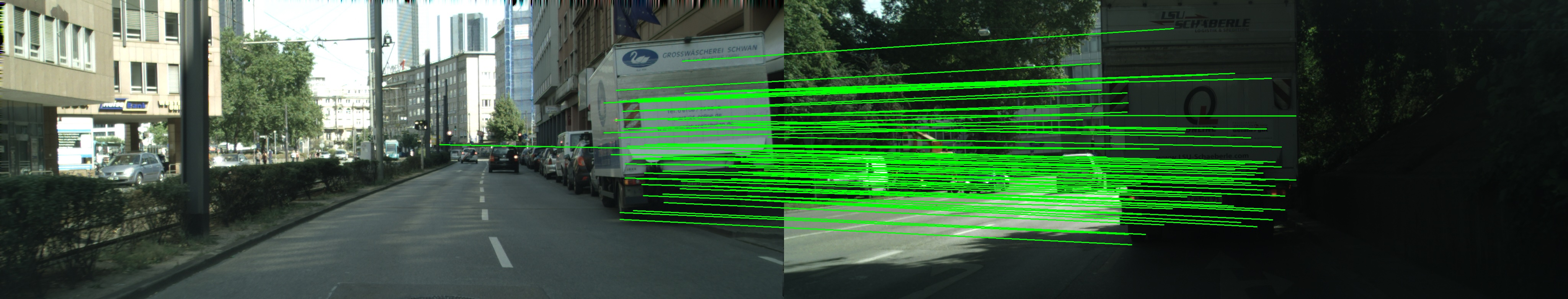}
  \includegraphics[width=\linewidth]{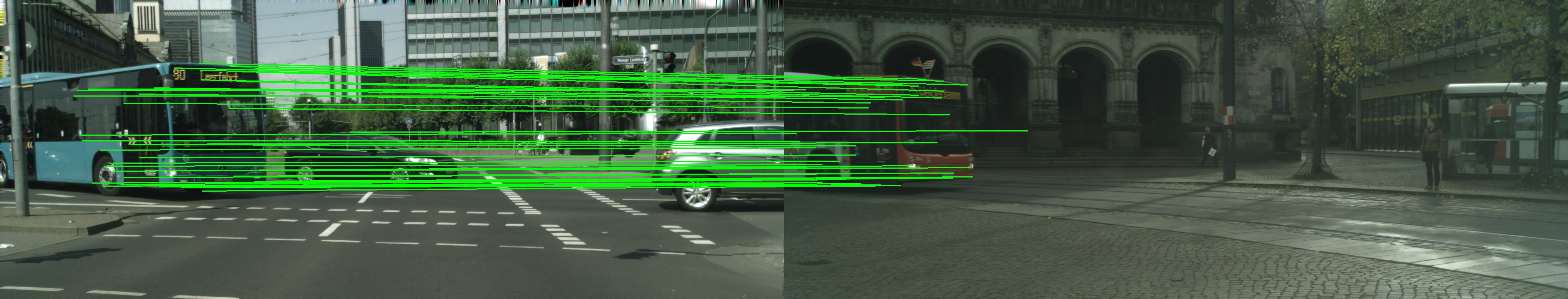}
  \caption{\textbf{Point matches for underrepresented classes} (top: tram, middle: truck, bottom: bus). The left image is sampled from the validation set, while the right image is from the training set. Matches remain consistent despite variations in illumination and scale.}
  \label{fig:point matching}
\end{figure*}

\section{Additional Qualitative Results}

Additional qualitative results are presented in Fig.~\ref{fig:additional qualitative results}.

\section{Mesh Creation for 3D Augmentation}

During training, context samples may be generated through 3D augmentation (Sec.~\ref{sec:data preparation}). In this process, the input image is rendered from a novel viewpoint using ground-truth depth and camera calibration parameters. A straightforward approach is to back-project each pixel into 3D and re-project it onto a new image plane; however, this often produces large regions without valid signal because the new viewpoint is misaligned with the original pixel grid. Such gaps degrade the realism of augmented samples and can negatively impact downstream learning.

\begin{figure*}[tbh]
    \centering
    \begin{subfigure}[b]{0.33\textwidth}
        \includegraphics[width=\textwidth]{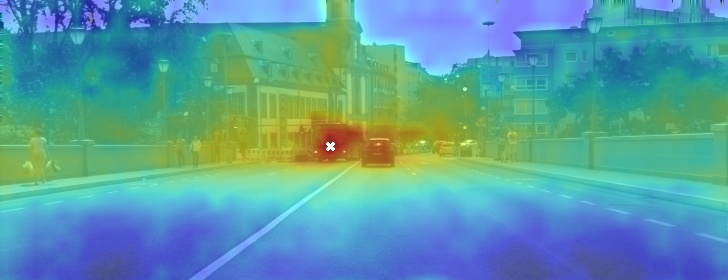}
    \end{subfigure}
    \begin{subfigure}[b]{0.66\textwidth}
        \includegraphics[width=\textwidth]{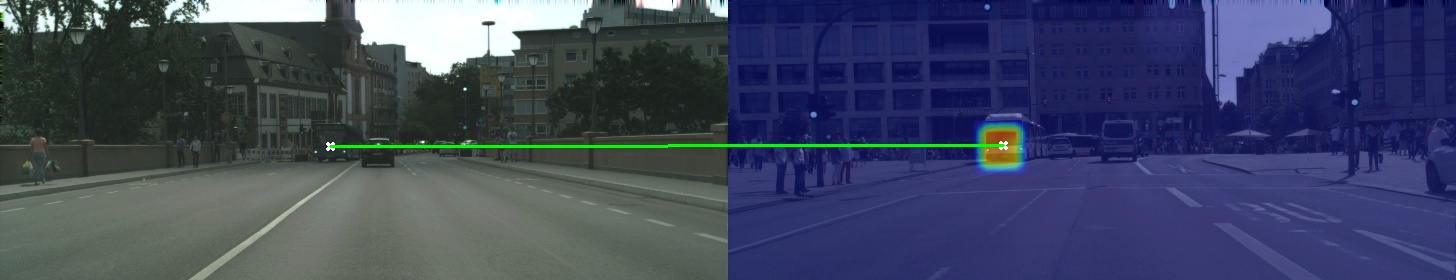}
    \end{subfigure} \\
    \begin{subfigure}[b]{0.33\textwidth}
        \includegraphics[width=\textwidth]{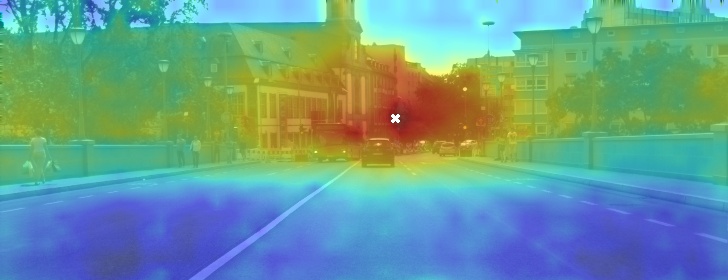}
    \end{subfigure}
    \begin{subfigure}[b]{0.66\textwidth}
        \includegraphics[width=\textwidth]{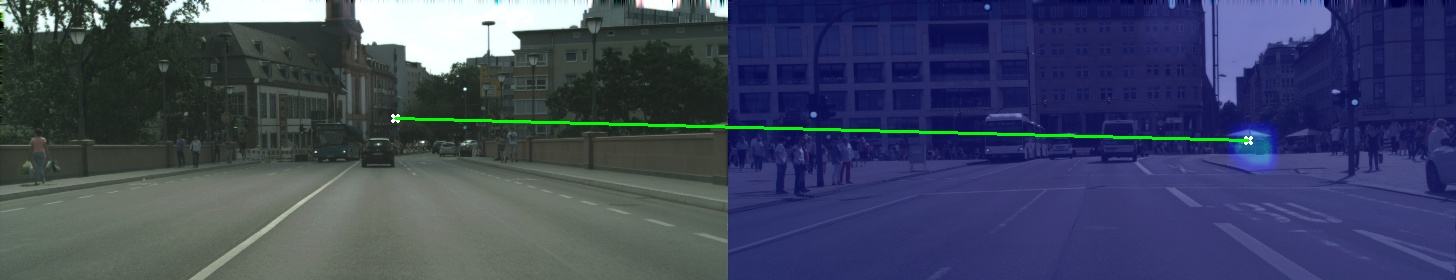}
    \end{subfigure} \\
    \begin{subfigure}[b]{0.33\textwidth}
        \includegraphics[width=\textwidth]{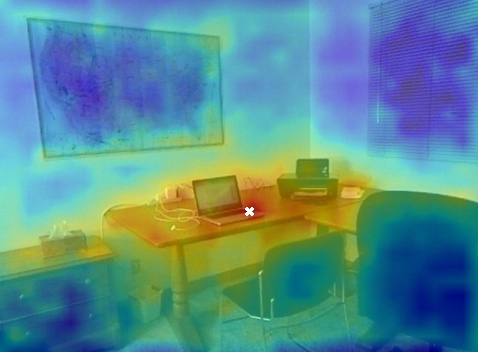}
    \end{subfigure}
    \begin{subfigure}[b]{0.66\textwidth}
        \includegraphics[width=\textwidth]{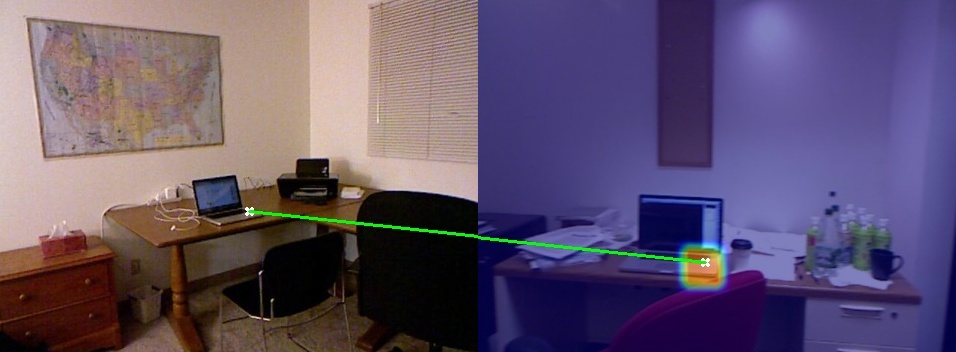}
    \end{subfigure} \\
    \begin{subfigure}[b]{0.33\textwidth}
        \includegraphics[width=\textwidth]{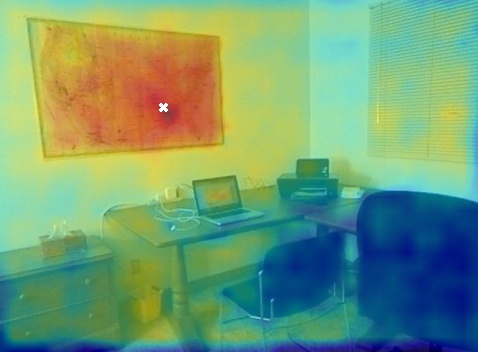}
        \caption{}
    \end{subfigure}
    \begin{subfigure}[b]{0.66\textwidth}
        \includegraphics[width=\textwidth]{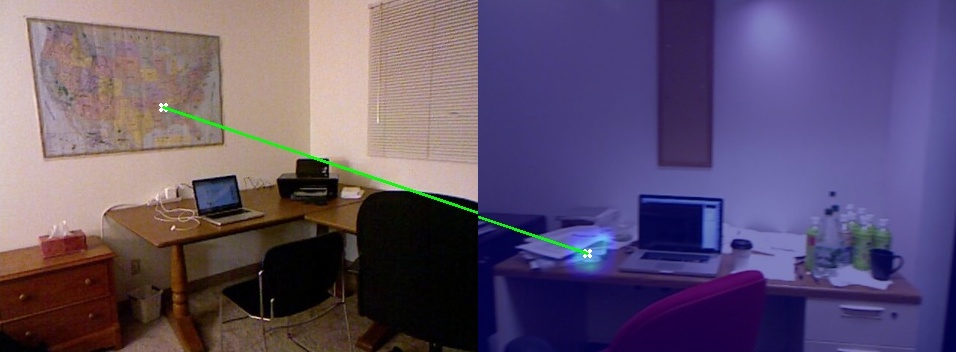}
        \caption{}
    \end{subfigure}
    \caption{\textbf{Visualization of matched cross-attention}. For a selected patch in the input image, indicated by a white marker, and its corresponding matched point in the context image, we separately visualize the attention directed toward the input (a) and the context (b). 
    When the match is correct, strong cross-attention emerges within the local neighborhood of the matched point in the context image. In contrast, incorrect matches yield weak cross-attention responses. Colormaps are kept consistent across images. }
    \label{fig:attention supplementary}
\end{figure*}

To address this limitation, we reconstruct a continuous surface mesh from the back-projected 3D point cloud, enabling dense coverage and improved geometric consistency. For mesh generation, we employ NKSR \cite{nksr}, which has demonstrated strong performance across multiple domains. Figure~\ref{fig:mesh creation} compares image re-projection using only the point cloud versus the full mesh, highlighting the substantial improvement in completeness and fidelity.

\section{Additional Component Visualizations}

\subsection{Point Matching}

Fig.~\ref{fig:point matching} demonstrates the performance of the point-matching algorithm on three representative input images containing underrepresented classes (left) and their corresponding retrieved images (right) from Cityscapes. While point-matching methods are typically trained on multiple views of the same scene, LightGlue \cite{lightglue} exhibits strong generalization to semantic similarities across different scenes, even under variations in scale and illumination. Notably, most matches occur on underrepresented objects, as these elements tend to be the most visually similar due to the uncertainty-aware retrieval mechanism described in Sec.~\ref{sec:data preparation}.

\subsection{Matched cross-attention}

Fig.~\ref{fig:attention supplementary} illustrates additional results for matched cross-attention. Attention to input patches consistently concentrates on relevant geometric structures, whereas attention to context patches is pronounced when point correspondences are correct and markedly weaker when correspondences are incorrect. This behavior underscores the model’s ability to leverage accurate matches for contextual reasoning, while reducing reliance on erroneous correspondences, thereby enhancing robustness.

\section{Training and Implementation Details}

\subsection{Network Optimization}

Our training procedure was carried out in stages, as outlined below:
\begin{enumerate}
\item Start with pre-trained DepthAnything v2.

\item Fine-tune the DepthAnything v2 network on the training dataset to produce metric depth rather than relative depth. 
All weights are trained, with a significantly lower learning rate applied to the encoder, as recommended by the original authors.

\item Construct the context stream encoder by duplicating the fine-tuned encoder from Step 2. Modify the projection operation in the context ViT encoder to accept the depth channel (Sec.~\ref{sec:depth estimation network} in the main paper).

\item Freeze the decoder and fine-tune the dual-stream encoder. Optimization is performed on the training dataset using the complete retrieval-augmented pipeline (Sec.~\ref{sec:data preparation} in the main paper).
\end{enumerate}

In Step 4 we optimized the positional encoding in both streams so that the network learns to differentiate between the two types of inputs.

As objective, we used the scale invariant log loss \cite{eigen2014depth}, also used by DepthAnything v2.
We trained our models using the AdamW optimizer with $\beta_1 = 0.9$, $\beta_2 = 0.999$, and a weight decay of $1 \times 10^{-2}$. The initial learning rate was set to $5 \times 10^{-5}$ and decayed by a factor of 10 after 5 epochs of stagnation. Training was performed on four Nvidia L40S GPUs.

\subsection{Uncertainty-aware Image Retrieval}
To compute uncertainty, we generated 5 noisy variants of a given image image by adding Gaussian noise with a standard deviation  $\sigma = 0.1$ (for normalized image values). The parameters $h$ and $q$ in Section~\ref{sec:data preparation} were set to 0.05 and 20\%, respectively. Unless explicitly stated otherwise, in the experiments (Sec.~\ref{sec:results}) we retrieved 4 images and applied matched cross-attention using a $3 \times 3$ patch neighborhood. To perform efficient search we use the FAISS library \cite{faiss}.


\subsection{Baseline Details}
For each baseline in the experiments section (Sec.~\ref{sec:results}) we use the publicly available code and weights. Specifically, for ZoeDepth \cite{ZoeDepth} we use the horizontal flip augmentation suggested by the authors. All methods are run according to their official instructions.

\section{Runtime and Memory Analysis}

We evaluate the model size, memory footprint, and inference runtime of RAD, and compare these metrics against those of the baseline architecture, DepthAnything v2 \cite{DepthAnything_V2}, as presented in Tab.~\ref{tab:runtime and memory analysis}. All measurements are conducted on Cityscapes images at a resolution of $2048\times 1024$, using a single Nvidia L40S GPU and a RAD configuration that incorporates four context samples.

\begin{table}[tbh]
    \centering
    \caption{\textbf{Parameter count, memory and runtime analysis.} Compared to DepthAnything v2 \cite{DepthAnything_V2}, RAD has approximately twice the parameter count and memory due to the second stream. Runtime is in seconds.}
    \small
    \begin{tabular}{lccc}
        \toprule
        Method & Parameters & Memory & Runtime \\
        \midrule
        DepthAny-Small & 24.8M & 94.55 MB & 0.013 s \\
        DepthAny-Base & 97.5M & 371.82 MB & 0.017 s \\
        DepthAny-Large & 335.3M & 1279.13 MB & 0.019 s \\
        \midrule
        RAD-Small & 48.5M & 184.99 MB & 1.724 s \\
        RAD-Base & 187.4M & 714.72 MB & 1.738 s \\
        RAD-Large & 644.1M & 2457.02 MB & 1.780 s \\
        \bottomrule
    \end{tabular}
    \label{tab:runtime and memory analysis}
\end{table}

\begin{table}[tbh]
    \centering
    \caption{\textbf{Runtime breakdown for RAD-Large.} All values are reported in seconds.}
    \small
    \begin{tabular}{lccc}
        \toprule
        Process & RAD-Small & RAD-Base & RAD-Large \\
        \midrule
        Uncertainty & 0.017 s & 0.020 s &  0.024 s \\
        Segmentation & 1.487 s & 1.487 s & 1.487 s \\
        Point matching & 0.191 s & 0.191 s & 0.191 s \\
        KNN search & 0.015 s & 0.015 s & 0.015 s \\
        Feed-forward & 0.014 s & 0.025 s & 0.063 s \\
        \bottomrule
    \end{tabular}
    \label{tab:runtime breakdown}
\end{table}

A detailed breakdown of RAD’s runtime is provided in Tab.~\ref{tab:runtime breakdown}. To estimate uncertainty, five inference operations are executed in parallel, corresponding to five noisy versions of the input image. Similarly, the four context samples and the input image are processed concurrently during RAD’s feed-forward pass, synchronizing only to exchange information for matched cross-attention.

For RAD, both the parameter count and memory consumption are approximately doubled relative to DepthAnything v2, primarily due to the inclusion of the context stream. The inference time is substantially higher, rendering RAD unsuitable for real-time applications. This increase in runtime is largely attributable to the SAM2 segmentation \cite{sam2} and LightGlue point matching \cite{lightglue} procedures. Consequently, the proposed approach would benefit significantly from future advances in segmentation and point-matching frameworks that combine accuracy with computational efficiency. Furthermore, it is possible to adapt the method to real-time applications by removing the segmentation process, and ussing only uncertainty, yielding a bit more noisy retrieval but improved runtime.

\section{Limitations}
RAD exhibits three primary limitations:

\begin{enumerate}
    \item \textbf{High runtime.} The method’s inference time is prohibitively large for real-time applications, with most of the overhead arising from image segmentation used for uncertainty-aware retrieval. A potential alternative is to employ pixel-wise uncertainty without coupling it to segmentation, enabling faster retrieval at the cost of increased noise, a trade-off that may be acceptable for real-time scenarios.
    
    \item \textbf{Dependence on depth-annotated context data.} RAD requires a context dataset with ground-truth depth, which is not always available. While the context stream could operate on RGB-only inputs and leverage proxy multi-view cues to improve depth estimation, ground-truth depth is a strong signal for accurate performance.

    \item \textbf{Sensitivity to retrieval and matching errors.} The approach relies on accurate image retrieval and point matching, as these components feed the matched cross-attention mechanism. Although attention partially mitigates this issue by down-weighting unreliable patches, errors in these stages can still propagate and degrade overall performance.
\end{enumerate}



\end{document}